\documentclass{article}

\usepackage{microtype}
\usepackage{graphicx}
\usepackage{subcaption}
\usepackage{booktabs}

\usepackage{tocloft}
\usepackage{minitoc}

\usepackage{hyperref}

\makeatletter
\let\orig@addcontentsline\addcontentsline
\makeatother

\usepackage[accepted]{icml2026}

\makeatletter
\let\addcontentsline\orig@addcontentsline
\makeatother

\usepackage{mathtools}
\usepackage{changepage}
\usepackage{amsmath, amssymb, amsthm}
\usepackage{bm}
\usepackage{bbm}
\usepackage{enumitem}

\usepackage{multirow}
\usepackage{makecell}
\usepackage{xcolor}   
\usepackage{colortbl} 
\usepackage{booktabs} 

\usepackage[capitalize,noabbrev]{cleveref}

\newcommand{\pdata}{p_{\text{data}}}
\newcommand{\pMN}{p_{\text{\scriptsize{MN}}}}
\newcommand{\gexpand}{\gamma_{\scriptsize{ex}}}
\newcommand{\scollapse}{\sigma_{\text{collapse}}}
\newcommand{\Ttest}{\mathcal{T}_{\text{test}}}
\newcommand{\Ttrain}{\mathcal{T}_{\text{tr}}}

\newcommand{\sigopt}{\sigma_{\text{opt}}}

\theoremstyle{plain}

\theoremstyle{definition}

\theoremstyle{remark}

\usepackage[textsize=tiny]{todonotes}

\icmltitlerunning{Smoothing the Score Function for Generalization}

\begin{document}

\twocolumn[
  \icmltitle{Smoothing the Score Function to Enhance Generalization in Diffusion Models}



  \icmlsetsymbol{equal}{*}

  \begin{icmlauthorlist}
    \icmlauthor{Xinyu Zhou}{yyy}
    \icmlauthor{Jiawei Zhang}{yyy}
    \icmlauthor{Stephen J. Wright}{yyy}
  \end{icmlauthorlist}

  \icmlaffiliation{yyy}{Department of Computer Sciences, University of Wisconsin Madison, WI, USA}

  \icmlcorrespondingauthor{Equal Correspondence}{Emails: \{xzhou555, jzhang2924, sjwright2\}@wisc.edu}

  \icmlkeywords{Machine Learning, ICML}

  \vskip 0.3in
]



\printAffiliationsAndNotice{}  

\begin{abstract}
Diffusion models achieve remarkable generation quality, yet face a fundamental challenge known as \textbf{memorization}, where generated samples can replicate training samples exactly.  
We develop a theoretical framework to explain this phenomenon by showing that the empirical score function (the score function corresponding to the empirical distribution) is a weighted sum of the score functions of Gaussian distributions, in which the weights are sharp softmax functions. 
This structure causes individual training samples to dominate the score function, resulting in sampling collapse. 
In practice, approximating the empirical score function with a neural network can partially alleviate this issue and improve generalization. 
Our theoretical framework explains why:  In training, the neural network learns a smoother approximation of the weighted sum, allowing the sampling process to be influenced by local manifolds rather than single points. Leveraging this insight, we propose two novel methods to further enhance generalization: (1) \textbf{Noise Unconditioning} enables each training sample to adaptively determine its score function weight to increase the effect of more training samples, thereby preventing single-point dominance and mitigating collapse. (2) \textbf{Temperature Smoothing} introduces an explicit parameter to control the smoothness. By increasing the temperature in the softmax weights, we naturally reduce the dominance of any single training sample and mitigate memorization. Experiments across multiple datasets validate our theoretical analysis and demonstrate the effectiveness of the proposed methods in improving generalization while maintaining high generation quality. \href{https://github.com/XinyuZhou2001/smoothing_the_score}{Code and models} are publicly available at Github.

\end{abstract}

\section{Introduction}
\label{sec:intro}

Diffusion models \cite{ho2020denoising, song2020score} have emerged as a leading framework in generative modeling, achieving state-of-the-art results across various applications, from text-to-image generation \cite{rombach2022high, esser2024scaling} to the discovery of protein structures \cite{watson2023novo, wu2024protein} and solving optimization problems \cite{krishnamoorthy2023diffusion, zhang2025gradient}. 
The core mechanism of diffusion models involves gradually perturbing data with Gaussian noise and learning to reverse this process through score-based denoising. 
But in high dimensions, the data distribution generally occupies a limited region under the Manifold Hypothesis \cite{fefferman2016testing}, which means that the score matching \cite{vincent2011connection} will be ineffective due to inaccurate gradient estimation in low-density regions \cite{song2019generative}. 
Consequently, diffusion modeling has been proposed to estimate a series of marginal distributions (with wider support) via neural network approximations, ensuring that the score function can be accurately learned along the entire sampling path, and enabling reliable generation through iterative denoising. 

However, a growing body of research \cite{carlini2023extracting, somepalli2023understanding, wen2024detecting, ren2024unveiling} reveals that although diffusion models can generate high-quality samples, some of these samples are identical to training examples. 
This issue is called \textbf{\textit{memorization}}. 
Memorization can be explained theoretically from the distribution evolution perspective \cite{li2024good}, as follows.
The Fokker-Planck equation for a diffusion process indicates that if we use the empirical score function to solve the reversed process, each sampled marginal distribution will be the same as its corresponding forward one. 
Thus, we will eventually go back to the training distribution and cannot produce any new samples once the neural network learns the empirical score function perfectly. 
Although this theory suggests that perfect learning of the objective function would limit the approach to replicating, novel samples are obtained in practice. 
This inconsistency motivates us to explore a fundamental question:
\vspace{-0.21cm}
\begin{quote}
\textit{How can neural networks \textbf{partially} resolve the memorization issue?}
\end{quote}
\vspace{-0.21cm}
This question is crucial since it is directly related to how we can further improve the \textbf{\textit{generalization}} properties of the diffusion approach.

The first thing to note is that the score function $s_\theta(x,\sigma)$ learned by a neural network will be slightly different from the empirical score function $s^*(x,\sigma)$. 
So, a study of the difference could be key to answering this question. 
To be more precise, let $\pdata(x)$ be the ground-truth distribution, and $\{\mu_j\}^M_{j=1}$ be a set of i.i.d samples drawn from it. 
The empirical distribution can be represented as a delta distribution $p^*(x) \triangleq \frac{1}{M} \sum_{j=1}^{M} \delta(x - \mu_j)$, where we have $\pdata(x)\approx p^*(x)$ if $M$ is sufficiently large. 
The noise-perturbed distribution then will be a series of Gaussian mixture distributions 
\[
p^*_i(x):= \frac{1}{M} \sum_{j=1}^{M}   \mathcal{N}({x}; \mu_j, \sigma_i^2 \mathbf{I}), \quad i\in[N], 
\]
where $\{\sigma_i\}^N_{i=1}$ is the noise schedule. 
For this setting we can explicitly calculate the empirical score functions 
\[
\nabla_x \log p_i^*(x)= -\sum_{j=1}^{M}  w_{j}(x) \frac{x - \mu_j}{\sigma_i^2}, \quad i \in [N],
\]
where $w_{ij}(x) = \frac{ \mathcal{N}(x; \mu_j, \sigma_i^2 \mathbf{I})}{\sum_{l=1}^{M} \mathcal{N}(x; \mu_l, \sigma_i^2 \mathbf{I})}$ 
is the score function weight, which also represents the probability that $x$ is generated by the Gaussian component centered at $\mu_j$.

 \begin{figure}
    \centering
    \includegraphics[width=0.95\linewidth]{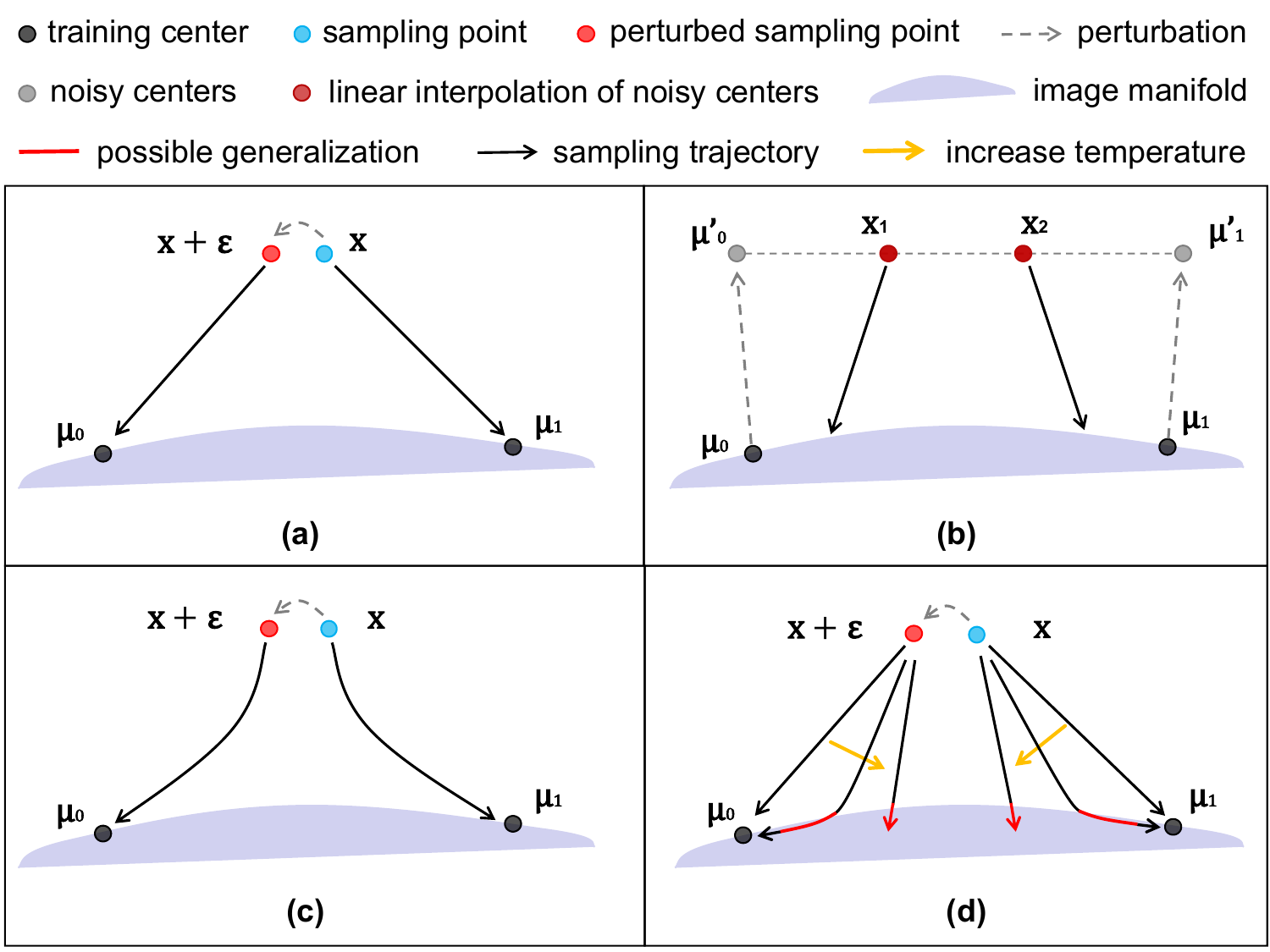}
    \caption{Illustration of sampling behaviors with different score functions. (a) Standard diffusion empirical score function leads to memorization, where trajectories collapse directly to training points. (b) The neural network-learned score function enables generalization. (Interpolation experiment) (c) The noise unconditioning allows smoother score function weights and delayed collapse. (d) The temperature-based score function promotes manifold exploration and prevents premature collapse to training points.}
    \label{fig1}
\end{figure}

We obtain further insights by combining algebraic and geometric methods. In high dimensions ($x \in \mathbb{R}^d$ for large $d$), most of the probability mass of a Gaussian distribution concentrates in a thin shell. Given an empirical marginal distribution $p^*_i(x) = \frac{1}{M} \sum_{j=1}^{M} \mathcal{N}({x}; \mu_j, \sigma_i^2 \mathbf{I})$, we can interpret it geometrically as a collection of $M$ ring shells. Each shell is centered at a training point $\mu_j$ with radius approximately $\sigma_i\sqrt{d}$ and thickness $3\sqrt{2}\sigma_i$ (\cref{A.3}).
Since the sampling point $x$ can be regarded as drawn from one of the marginal distributions $p_i^*$, 
it should be contained in at least one shell with the noise level $\sigma_i$. 
Moreover, at late stages of sampling, the value of $\sigma_i$ is small enough that $x$ is unlikely to be in the overlap of different shells, so the score function weight will be dominated by a single Gaussian component (\cref{prop2}). 
Thus, $\nabla_x \log p_i^*(x) \approx  -\frac{x - \mu_j}{\sigma_i^2}$ for a particular value of $j$, so that subsequent denoising steps collapse to the corresponding center $\mu_j$ --- the memorization issue.

We also find that even a small perturbation to $x$ can significantly alter the empirical score function, as its weight concentration may transfer from one center $\boldsymbol\mu_j$ to another. 
This function is extremely sharp in regions where the shells overlap (\cref{sharp_overlap}), 
as illustrated in  \cref{fig1} (a). 
However, due to capacity and regularizations \cite{yoon2023diffusion, bonnaire2025diffusion}, the neural network tends to learn a smoother function than the spiky empirical score function.
This claim is also supported by the interpolation experiments in diffusion models \cite{ho2020denoising, song2020score}, illustrated in \cref{fig1} (b). 
Specifically, two training points $\mu_0$ and $\mu_1$ are first corrupted with noise to obtain perturbed versions ${\mu'}_0$ and ${\mu'}_1$. 
The interpolated sampling points are then created by taking linear combinations $x_1 = \lambda_1 {\mu'}_0 + (1-\lambda_1) {\mu'}_1$ and $\quad x_2 = \lambda_2 {\mu'}_0 + (1-\lambda_2) {\mu'}_1$, where $\lambda_1,\lambda_2\in(0,1)$ are coefficients sufficiently away from the endpoints. 
After denoising sampling, the interpolated points will converge to the manifold between $\mu_0$ and $\mu_1$ rather than collapsing to either of them.
Based on these observations, we believe that we can resolve the question posed earlier as follows:

\begin{adjustwidth}{-0.1cm}{-0.1cm}
\begin{quote}
\textit{The neural network can resolve the memorization since it smooths the score function weights in the empirical score function. 
Thus, at late sampling stages, the score function is determined by a local manifold approximated by several nearby training samples, rather than by a single training point.}
\end{quote}
\end{adjustwidth}

\noindent
We justify this claim in the remainder of the paper. 

Nevertheless, as the model capacity increases, the neural network is more able to capture the ``spikiness'' of the empirical score function.
Therefore, if a collapse point is isolated or replicated in the dataset, memorization may still occur. 
These are several explanations as to why memorization consistently happens in large models and why memorized samples are typically either far from other training points or repeated in the dataset \cite{yoon2023diffusion, gu2023memorization}.
It follows that by relaxing the objective constraints and smoothing the objective score function weights $w_{ij}$, we may be able to improve generalization (and decrease memorization).
Based on the formula of the empirical score function, we propose two approaches for smoothing.

\textbf{Noise Unconditioning.} In most diffusion models, the noise levels are designed as continuous, which means the shells pack the space. 
Therefore, we know from \cref{prop2.1} that for the fixed center $\mu$ and position $x$, there is an ``optimal shell" corresponding to some noise level $\sigma_*$ that will dominate the score function weight over all noise levels.
However, in standard diffusion models, the noise level is fixed at each sampling stage, which means the sampling point may not lie on the optimal shells around most training points, leading to sub-optimal score function weights (much smaller) for these points.
So, our first smoothing method is to \emph{remove conditioning on noise}, allowing each training point to adaptively find its optimal shell at the given position $x$. 
This ensures that other training points can contribute to the score function, preventing a single point from dominating, and thus mitigating memorization.
In addition, this noise unconditioning  enables us to regard the sampling as a gradient ascent process, where the optimal solutions are the training points. 
In this case, even though the sampling may still collapse to the centers, the ``collapse time'' to a particular center $\mathbf{\mu}_j$ will be delayed, since other centers still make significant contributions during the sampling. 
For illustration, compare \cref{fig1} (c) to \cref{fig1} (a).

\textbf{Temperature Smoothing.} Given that the score function weights ${w}_{ij}(x)$ are defined as softmax functions, another smoothing approach becomes apparent: Modify the definition of these weights by introducing a temperature parameter to smooth them.
The effect of smoothing is to delay the collapse of the denoising process; see \cref{fig1} (d). 
By using higher temperatures at smaller noise levels, the sampling will have chances to continuously explore the local manifold rather than collapsing to a single point. 
This enables the neural network to sample from combinations over local manifold containing nearby samples, making it more likely to generalize rather than memorize. 
If the temperature is not sufficiently high, collapse may still occur eventually, but early stopping may prevent full collapse while maintaining the exploration benefits (see the red regions in \cref{fig1} (d)). 
However, we must also avoid excessively high temperatures as this would introduce influence from training points in an excessively large neighborhood, causing the generated point to deviate significantly from the image manifold.


    

Overall, we make the following contributions.
\vspace{-0.2cm}
\begin{itemize}[leftmargin=10pt]
    \item We theoretically prove that memorization stems from the sharpness of empirical score functions, and experimentally verify that neural networks achieve generalization by implicitly smoothing the score-function weights across different centers, i.e., increasing the influence of samples other than the nearest one.

    \item We propose two novel methods to mitigate memorization: (1) \emph{Noise Unconditioning}, a framework that unifies distributions across noise levels and is equivalent to explicit score matching on a unified Gaussian mixture, so that sampling can be interpreted as gradient ascent on this fixed objective, enabling optimization-based strategies for generation. (2) \emph{Temperature Smoothing}, a plug-and-play modification of the training objective that explicitly controls the smoothness of score weights and improves generalization with only modest extra cost.

    \item Our explanation framework bridges the gap between diffusion theory and practice, providing theoretical foundations for many previously confusing experimental phenomena. This theoretical grounding may facilitate further advances in generative modeling.
\end{itemize}

\section{Preliminaries}
\label{prelim}

\subsection{Forward SDEs and Backward ODEs}

Stochastic differential equations (SDEs) provide a powerful framework for describing diffusion processes \cite{song2020score}. A general forward SDE in \(\mathbb{R}^d\) can be written as:
\begin{equation}
    \mathrm{d} x_t = \mathbf{\mu}(x_t, t) \, \mathrm{d}t + \sigma(t) \, \mathrm{d}\mathbf{w}_t,
    \label{eq:general_sde}
\end{equation}
where \(x_t \in \mathbb{R}^d\) is the state of the system at time \(t \in [0, T]\), \(\mathbf{\mu}(\cdot, \cdot)\) and $\sigma(\cdot)$ are the drift and diffusion coefficients respectively, and \(\mathbf{w}_t\) represents the standard Brownian motion. This forward SDE describes how data is progressively perturbed into a noise-like distribution as \(t\) increases. A remarkable property of this SDE is the existence of an ordinary differential equation (ODE), dubbed the Probability Flow ODE \cite{song2020score}, whose solution trajectories sampled at $t$ are distributed according to:
\begin{equation}
    \mathrm{d} x_t = \left[\mathbf{\mu}(x_t, t) - \frac{1}{2}\sigma(t)^2 \nabla_{x} \log p_t(x)\right] \, \mathrm{d}t,
    \label{eq:reverse_ode}
\end{equation}
where \(\nabla_{x} \log p_t(x)\) is the score function of the perturbed distribution at time \(t\), namely the learning target of a neural network $s_\theta(x,t)$ via \textit{score matching} \cite{vincent2011connection, hyvarinen2005estimation, song2019generative}.

\subsection{Noise Unconditioning: ODE as Gradient Flow}

Note that we let $\mathbf{\mu}(x, t) = 0$ and $\sigma(t)=\sqrt{2t}$ following the variance exploding SDE \cite{song2020score, karras2022elucidating}. This choice ensures that we have $p_t(x)=p^*(x) \otimes \mathcal{N}(0, t^2 \mathbf{I})$, i.e., $p_i^*(x)=p^*(x) \otimes \mathcal{N}(0, \sigma_i^2 \mathbf{I})$, where $\otimes$ denotes the convolution operation. When we remove the noise conditioning, i.e., $s_\theta(x,t)\to s_\theta(x)$, all marginal distributions will be unified into one space. This unified distribution can be approximated by an $M \times N$ Gaussian mixture (\cref{A.1}):
\begin{equation}
\pMN (x) = \frac{1}{ZMN} \sum_{j=1}^{M} \sum_{i=1}^{N} \lambda(\sigma_i) \mathcal{N}(x; \mu_j, \sigma_i^2 \mathbf{I}),
\label{pmn}
\end{equation}
where $Z$ is the normalization constant and $\lambda(\sigma_i)$ represents the weighting function for different noise levels. Following standard practice in diffusion models \cite{song2019generative, song2020score}, we choose $\lambda(\sigma_i) = \sigma_i^2$ during training.

For the unified distribution $\pMN$, the regions close to training points (Gaussian centers) naturally exhibit the highest probability density in high dimensions. So sampling these centers can be regarded as maximizing $\log \pMN(x)$, thereby the sampling process can be reformulated as gradient ascent flow of this fixed objective function $\log \pMN$:
\begin{equation}
    \frac{\mathrm{d} x}{\mathrm{d} t} = \frac{\eta}{2} \nabla_{x} \log \pMN (x),
    \label{gradient_ascent}
\end{equation}
where $dt$ now parameterizes the optimization trajectory (rather than reversed diffusion time, since $\log \pMN (x)$ is time-independent) and $\eta>0$ is the step size. 
This ODE represents continuous gradient flow on $\log \pMN (x)$ to maximize the likelihood.





\section{From Memorization to Generalization}
\label{expan}

We now analyze why memorization occurs and how it is related to the sharpness of the score functions. 
Furthermore, we explain how the neural networks and our methods can generalize through comparing the expansiveness of the gradient ascent operators associated with their score functions.

\subsection{Why Memorize?}
\label{sec:memorize}

For a Gaussian mixture distribution $\pMN$ as in \cref{pmn}, we can explicitly calculate its score function:
\[
\nabla_{x} \log \pMN (x) = -\sum_{j=1}^{M} \sum_{i=1}^{N} w_{ij}(x) \frac{x - \mu_j}{\sigma_i^2}.
\]

\vspace{-0.1cm}
\noindent
where 
\[
w_{ij}(x) = \frac{\sigma_i^2 \mathcal{N}(x; \mu_j, \sigma_i^2 \mathbf{I})}{\sum_{l=1}^{M} \sum_{k=1}^{N} \sigma_k^2 \mathcal{N}(x; \mu_l, \sigma_k^2 \mathbf{I})}.
\]

\noindent
and this score function weight can be written as a function $f(x,\mu_j,\sigma_i) \triangleq   -(d-2)\ln \sigma_i - \frac{||x-\mu_j||^2}{2\sigma_i^2} $ with softmax:
\[
\begin{split}
w_{ij}(x) &= \frac{\exp(f(x,\mu_j,\sigma_i))}{\sum_{l=1}^{M} \sum_{k=1}^{N} \exp(f(x,\mu_l,\sigma_k))} \\
&= \mathrm{Softmax}_{j,i} (f(x,\mu_j,\sigma_i)).
\end{split}
\] 
We have the following conclusion for this score function:

\noindent
\textbf{Proposition 2.} 
In high dimensions $d$, given a position \( x \) that is not far from any Gaussian centers, the score function \( \nabla_{x} \log \pMN (x) \) is predominantly determined by the Gaussian component $\mathcal{N}(x; \mu_*, \sigma_*^2 \mathbf{I})$, where \( \mu_* \) and $\sigma_*$ are defined as follows:
\begin{align*}
\mu_* &= \arg\min_{\mu_j} \|x - \mu_j\|_2, \quad j = 1, \dots, M, \\
\sigma_* &= \arg\min_{\sigma_1,\sigma_2,\dotsc,\sigma_N} \, \left| \frac{\sigma_{opt}^2}{\sigma_i^2} -1 \right|.
\end{align*}
where  $\sigma_{opt} := \frac{\|x-\mu_*\|}{\sqrt{d-2}} \approx  \frac{\|x-\mu_*\|}{\sqrt{d}}$.
\label{prop2}

The proposition can be proved according to two properties of the score function weights:

\noindent
\textbf{Property 1. $\sigma$ domination:} Given a certain position $x$ and any fixed center $\mu_j$, there exists an optimal noise level $\sigma_j^*$ such that its corresponding Gaussian component dominates the score function weight over all other noise levels. Mathematically, $\sigma_j^*$ is the global maximum of the function $f(\sigma)= -(d-2)\ln \sigma - \frac{||x-\mu_j||^2}{2\sigma^2}$, with $f(\sigma)$ demonstrating pronounced sensitivity to deviations from $\sigma_j^*$.

\label{prop2.1}

Under the noise scheduling that the boundaries of Gaussian shells touch (\cref{A.3}), namely a kind of log-uniform distribution as in \cite{song2020improved}, the ratio of the maximum to second-maximum score function weights across noise levels is (derivation in \cref{A.4.1}):
\[
\gamma_\sigma = \frac{w_{ij}(x)}{w_{i+1,j}(x)} \approx e^{18(1-2\alpha)}, \quad \mbox{for some $\alpha \in [0,0.5)$.}
\]

\noindent
We give an example to build intuition about how large the ratio can be even for modest values of $\alpha$. 
For the case of $\alpha=\frac{1}{3}$ (that is, $\sigma_{opt}$ is only slightly closer to $\sigma_i$ than to $\sigma_{i+1}$), we have 
$\gamma_\sigma \approx e^{6} \approx 403$, so $w_{ij}(x)$ dominates its closest neighbor. 
Its dominance over all other weights will be even greater.
Based on this property, we can then simplify the score function, representing the weight of a center $\mu_j$ by only using its optimal noise level $\sigma_j^*$:
\begin{equation} \label{eq:wjs}
w_j^*(x) :=  \frac{\sigma_j^{*2} \mathcal{N}(x; \mu_j, \sigma_j^{*2} \mathbf{I})}{\sum_{l=1}^{M}  \sigma_l^{*2} \mathcal{N}(x; \mu_l, \sigma_l^{*2} \mathbf{I})}.
\end{equation}

\noindent
\textbf{Property 2. $\mu$ domination:} \label{prop2.2} Given a position $x$, the score function weight of a center decreases exponentially as the distance between the center and $x$ increases.
We have the following conclusion for the case that two training points $\mu_j$ and $\mu_l$ are a similar distance from $x$, namely $\sigma_j^*=\sigma_l^*$ but $\|x-\mu_l\|\ge \|x-\mu_j\|$:
\begin{equation}
    \gamma_\mu \approx \frac{w_j^*}{w_l^*} = \exp \left( \frac{\delta||\mu_j-\mu_l||^2}{\sigma_j^{*2}} \right) 
    \label{ratio_mu}
\end{equation}
where $\delta:= \frac{||x-\mu_l||\cos\angle x\mu_l\mu_j}{||\mu_j-\mu_l||}-\frac{1}{2} \ge 0$ represents the relative distance advantage of $\mu_j$ over $\mu_l$ with respect to $x$ (see derivation in \cref{A.4.2}) and $\gamma_\mu$ captures the ratio of the weights corresponding to the two centers $\mu_j$ and $\mu_l$.
Typically, $\delta \lesssim O(\frac{1}{\sqrt{d}})$ maintains the condition $\sigma_j^*=\sigma_l^*$. However, during the sampling, the progressive reduction of $\sigma_j^*$ causes $\gamma_\mu$ to grow. 
When $\sigma_j^*$ is sufficiently small, the single center $\mu_j$ will dominate the score function weight, resulting in collapse and subsequent memorization.

\subsection{Generalization Evaluation}
\label{sharp_overlap}
From Property 2 in \cref{sec:memorize}, we know that memorization occurs when $\sigma$ is small because the closest training point dominates the score function weight, even when it is not much closer than other points. 
However, in practice, we generate samples using \emph{neural networks} to approximate the empirical score function, and this often results in novel samples.
These observations suggest that the score function learned by the neural network must differ from the empirical score function.
A natural explanation is that the neural network fits the score function quite well, but implicitly smooths its weights at small noise levels, preventing the dominance of individual training points and avoiding sampling collapse. 
Following this insight, we analyze the generalization ability of neural networks, standard diffusion, noise unconditioning, and temperature smoothing by comparing the expansiveness of their sampling process at points where the shells from different training points  overlap.
We simplify the analysis by considering only the two training points closest to $x$, which is sufficient for our purposes because Proposition 2 indicates that the closest training samples have the greatest (dominant) influence on the flow.

Denoting the two closest training points by $\mu_0$ and $\mu_1$, assume that the sampling point $x$ is similarly close to both of them, with the same shell radius $\sigma^* \sqrt{d}$ for both.
We assume $\|x-\mu_0\| = C\|x-\mu_1\| = C\sigma^*\sqrt{d}$, where $C \ge 1$. We then slightly perturb $x$ along the direction $(\mu_0-\mu_1)$ to obtain position $\mathbf{y}$, such that $\|\mathbf{y}-\mu_1 \| = C\|\mathbf{y}-\mu_0\| = C\sigma^*\sqrt{d}$. 
After one sampling step for both $x$ and $\mathbf{y}$, we arrive at $\mathbf{x'}$ and $\mathbf{y'}$ respectively. We then evaluate the expansion factor
\[
\gexpand := \frac{\|\mathbf{y'}-\mathbf{x'}\|}{\|\mathbf{y}-x\|}.
\]
This ratio serves as an indicator of generalization because 
1) memorization can induce an arbitrarily large ratio (formally unbounded in the limit); 
2) a bounded ratio (i.e., a locally non-expansive sampling map, in the same sense as non-expansive updates in optimization) preserves the local connectivity of the sampled distribution, and thus serves as a quantitative proxy for generalization.

\noindent
\textbf{Empirical vs. Neural Network:} Under the above assumption, for the empirical score function of standard diffusion, we have:
\begin{equation}
    \gexpand \approx \frac{\| \eta (\mu_0-\mu_1)\|}{\|y-x\|} \left | 1-\frac{2}{a+1} \right |
    \label{ratio_expan}
\end{equation}
where $\eta$ is the sampling stepsize and $a > 1$ represents the ratio of the dominant to subdominant score function weight
(see derivation in \cref{A.5}).
This result still holds for $\|y-x\|\to 0$ if $d\to\infty$, which means that $a \to \infty$ and the ratio could be unbounded, showing poor generalization. 
However, the results of interpolation experiments in diffusion models \cite{song2020score, ho2020denoising} show that $\|y'-x'\| \to 0$ if $\|y-x\| \to 0$. 
Our experiments (\ref{exp_ratio}) also show that the learned score function has a much smaller ratio than the empirical one.

\noindent
\textbf{Conditioning vs. Unconditioning.} Under our assumption, the unconditioning modeling has the same ratio formula as in \cref{ratio_expan} but with a different coefficient $a$. 
From the property of $\sigma$-domination, we know that the unconditioning case has a smoother score function weight, leading to a smaller ratio $\gexpand$ and therefore better generalization.

\noindent
\textbf{Temperature.} We modify the weight calculations by introducing a temperature vector $T \in \mathbb{R}^N$ whose $i$-th component $T_i$ contains the temperature for shell $i$. We define $T^*_j$ to be the value of $T_i$ for which $\sigma_i = \sigma_j^*$.
Generalizing \cref{eq:wjs}, the temperature-based score function weight is defined as:
\[
w_{j}^*(x;T) =  {\frac{\exp \left ( \frac{f(x, \mu_j, \sigma_{j}^*)}{T_j^*}  \right )}
{\sum_{l=1}^{M} \exp \left ( \frac{f(x, \mu_l, \sigma_{l}^*)}{T_l^*}  \right )}}.
\]
(We recover \cref{eq:wjs} by setting $T_i=1$, $i=1,2,\dotsc,N$.) 
As the $T_i$ increase to $\infty$, the temperature smoothing reduces the dominance ratio $a$, resulting in smaller $\gexpand$ and better generalization.

\section{Learning the Smoothed Score Function}

We next describe our methods for training the neural network score functions $\mathbf{s}_\theta$, parametrized by weight vector $\theta$.
Our noise scheduling follows the variance exploding SDE in \cite{song2020score}, where the noise scale \(\sigma\) is sampled from a discrete set of \(N\) points that approximate a log-uniform distribution over the range \([\sigma_{\text{min}}, \sigma_{\text{max}}]\). 
The probability of selecting each \(\sigma_i\) is equally $\frac{1}{N}$. We denote sampling from this noise schedule as $\sigma_i \sim p_{\sigma}$.

\subsection{Unconditioning Score Matching}
The loss function in standard noise conditioning score-based neural networks (NCSN) \cite{song2019generative} is
\begin{equation}
\mathcal{L}_{\text{c}} =  \mathbb{E}_{\mathbf{\sigma}_i\sim p_\sigma} \mathbb{E}_{\mathbf{\mu}\sim p^*} \mathbb{E}_{{x}\sim\mathcal{N}(\mu, \sigma_i^2 \mathbf{I})}
\left[\frac{\sigma_i^2}{2} \left\| \mathbf{s}_\theta({x}, \sigma_i) + \frac{{x} - \mu}{\sigma_i^2} \right\|_2^2 \right].
\end{equation}
We have a similar objective function for the unconditioning modeling:
\begin{equation}
\mathcal{L}_{\text{u}} =  \mathbb{E}_{\mathbf{\sigma}_i\sim p_\sigma} \mathbb{E}_{\mathbf{\mu}\sim p^*} \mathbb{E}_{{x}\sim\mathcal{N}(\mu, \sigma_i^2 \mathbf{I})}
\left[\frac{\sigma_i^2}{2} \left\| \mathbf{s}_\theta({x}) + \frac{{x} - \mu}{\sigma_i^2} \right\|_2^2 \right].
\label{uncon_loss}
\end{equation}
The only difference is that we remove the noise as an input to the neural networks. 
Following the denoising score matching \cite{vincent2011connection}, we show in \cref{A.6} that optimizing this loss function is equivalent to performing explicit score matching for the distribution $\pMN$.

\subsection{Temperature-based Score Matching}
From \cref{ratio_mu} and the analysis in Section \ref{expan}, we know that when the noise level is small, training samples close to $x$ will dominate the score function weights. 
Therefore, we set a threshold $\scollapse$ to determine when we should introduce temperature scaling.
When $\sigma_i \leq \scollapse$, for a noisy point $x$ drawn from $\mathcal{N}(\mu, \sigma_i^2 \mathbf{I})$ where $\mu \sim p^*$, $\mu$ should be the closest training sample to $x$. 
We can then approximate the score function using the top-$K$ nearest training samples to $\mu$, denoted by $\mu_{(j)}$, $j=1,2,\dotsc,K$, as follows:
\begin{equation} \label{eq:topk}
\nabla_{x} \log \pMN(x; T) \approx 
\sum_{j=1}^{K} w_{(j)}^*(x; T) \left ( -\frac{x-\mu_{(j)} }{\sigma_{(j)}^{*2}}  \right )
\end{equation}
where 
\[
\quad w_{(j)}^*(x; T) = \frac{\exp \left ( \frac{f(x, \mu_{(j)}, \sigma_{(j)}^*)}{T_{(j)}^*}  \right )}{\sum_{l=1}^{K} \exp \left ( \frac{f(x, \mu_{(l)}, \sigma_{(l)}^*)}{T_{(l)}^*}  \right )}.
\]
We thus define the score matching loss to be:
\[
\mathcal{L}_{\text{T}} =   \mathbb{E}_{{x}\sim \pMN}
\left[\frac{1}{2} \left\| \mathbf{s}_\theta({x}) - \nabla_x \log \pMN (x;T ) \right\|_2^2 \right].
\]
where practically we use \cref{eq:topk} to approximate $\nabla_x \log \pMN (x;T )$.

To sum up, the complete temperature-based training loss adaptively combines both approaches based on the noise level of each sample:
\begin{equation}
\mathcal{L} = \mathbb{E}_{\sigma_i \sim p_\sigma} \left[
\begin{cases}
  \mathcal{L}_{\text{u}},  & \text{if } \sigma_i > \scollapse  \\
  \mathcal{L}_{\text{T}}, & \text{if } \sigma_i \leq \scollapse 
\end{cases}
\right]
\end{equation}
Experimentally, we set $T_i=\max(\frac{\scollapse}{\sigma_i},1)$. 
Note that temperature-based score matching is not rigorously learning the score function of $\pMN$, but its smoothed proxy version.

\section{Experiments}

\begin{figure}[!htb]
    \centering
    \includegraphics[width=0.95\linewidth]{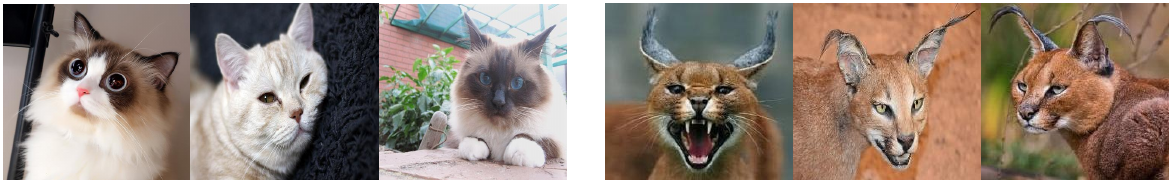}
    \caption{Illustration for the dataset of cat (\textbf{left}) and caracal (\textbf{right}). \textbf{Note} that there are no yellow or long-eared pet cats in the dataset.}
    \label{cat_caracal}
\end{figure}

\textbf{Experimental Setup.} 
To evaluate the effectiveness of our smoothing methods, we adopt VE-SDE \cite{song2020score} as the baseline, which is known as the first time-reverse SDE framework of diffusion models. 
For fair comparison, we reproduce the VE-SDE setup and apply our smoothing methods with the same implementations. For example, our unconditioning approach uses the same NN architecture as the baseline, with only the time embedding layers removed. (see \cref{B.1} for other details) 
In addition to the four commonly used datasets, CIFAR-10 \cite{krizhevsky2009learning}, CelebA 64$\times$64 \cite{liu2015deep}, ImageNet 64$\times$64 \cite{deng2009imagenet} and CelebA-HQ 256$\times$256 \cite{karras2017progressive}, we also collect a small 64$\times$64 dataset for ablation studies, consisting of 1,000 pet cats and 200 images of caracals. 
Caracals differ from pet cats in having long ears and ferocious faces (see \cref{cat_caracal}), so that we can clearly see any possible generalizations.

\noindent
\textbf{KNN Spaces.} The curvature of the image manifold in pixel space is large, so when K is large, or the temperature is high, the learned score function may guide samples deviating from the manifold. Therefore, we also apply KNN in the feature space (with significantly smaller curvature). Specifically, we use a pre-trained ResNet-18 first to map all samples to a 512-dimensional latent space, then calculate the KNN and return their indexes for later explicit score function calculation, still in the pixel space. We find that under the same temperature and K, the feature space-based KNN will lead to obviously better generation quality than the pixel space-based one. This phenomenon further verifies our former explanation that neural networks achieve generalization via learning a smoother score function determined by a local manifold. 
Note that in this section, all illustrated samples are obtained by pixel space-based KNN to support our claim. We show more results and comparison in the \cref{B.4.2}.

\noindent
\textbf{Quality Metric.} Besides calculating the Fr\'echet Inception Distance (FID) \cite{heusel2017gans} between generated samples ${G}$ and training samples $\Ttrain$, we also calculate the FID between generated samples and test samples $\Ttest$. Smaller ratio $\frac{\text{FID}(G, \Ttest)}{\text{FID}(G, \Ttrain)}$ could be roughly regarded as better generalization. Specifically, we randomly select the same number of samples per class from the training set as in the test set to calculate FID. For CelebA, we randomly select 10k images from both the training set and test set.

\noindent
\textbf{Sampling for Unconditioning.} We first try the existing samplers \cite{song2020score} by replacing the conditioning score function $s_\theta(x,t)$ with our unconditional version $s_\theta(x)$. 
We find that SDE samplers can still succeed while ODE samplers may fail catastrophically. 
Under our framework, this occurs because at late sampling stages, the unconditional score function adapts to the true noise level $\sigma_{n*} := \frac{||x_n-\mu_*||}{\sqrt{d}}$, where $x_n$ is current sample and $\mu_*$ is its closest training point.
However, the sampling schedule still uses the predetermined noise level $\sigma_n$. 
In ODE samplers, we empirically observe that $\sigma_n \gg \sigma_{n*}$ as $x_n$ approaches training data, creating a critical mismatch.
We prove in \cref{A.7.1} that this mismatch causes ODE samplers to take excessively large steps $\propto \sigma_n^2/\sigma_{n*}^2$, leading to catastrophic overshoot. 
In contrast, SDE samplers remain stable because their stochastic noise term creates a self-correcting mechanism that keeps $\sigma_{n*}$ reasonably close to $\sigma_n$ (\cref{A.7.2}). 
To address this mismatch, one way is to fix the implicit noise level after every ODE step like the Predictor-Corrector (PC) sampling in \cite{song2020score}. 
Another way is to directly replacing the predetermined $\sigma_n$ by $\sigma_{n*}$:
\[
x_{n+1} = x_n + \alpha \sigma_{n*}^2 s_\theta(x_n)
\]
where $\alpha$ controls the step size. 
Although computing $\sigma_{n*}$ requires finding the nearest training point which may be expensive for large-scale datasets, our unconditional score function enables bigger step sizes than standard diffusion models, resulting in significantly fewer function evaluations (NFE) that often compensate for this overhead. 
This is because the unconditioning score function is well-defined across the entire support of $\pMN$, and so avoids the error accumulation of large time steps in standard diffusion models.

\subsection{Ablation Studies on Cat Caracal Dataset}
\subsubsection{Qualitative Experiments}
\label{5.1.1}
\begin{figure}[!htb]
    \centering
    \includegraphics[width=0.95\linewidth]{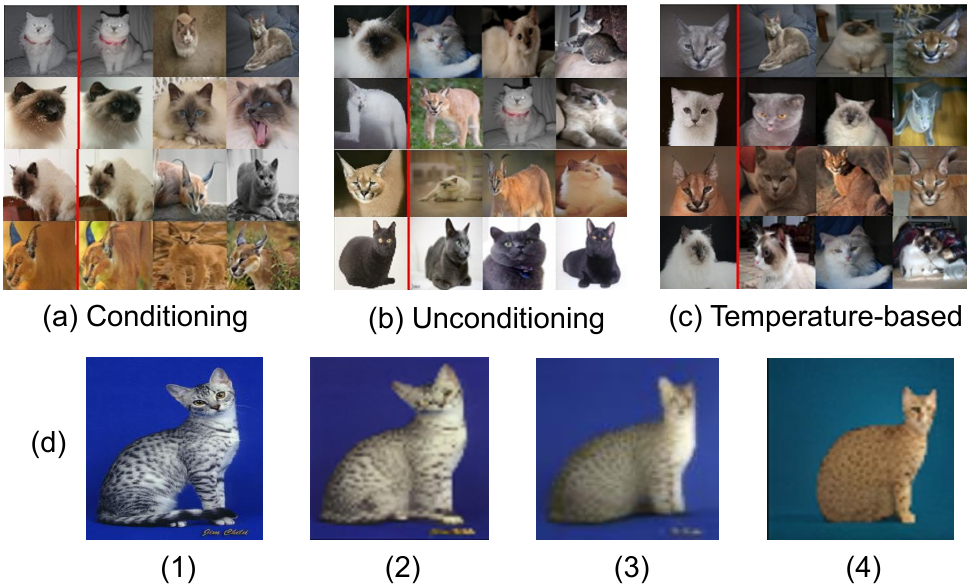}
    \caption{Qualitative comparison with different methods. (a, b, c) illustrates the generated images (left of red line) and their top-3 nearest neighbors of training samples in pixel space. (d) shows the generated images (2, 3, 4) and their corresponding closest  training sample (1). (closest in both pixel and feature spaces)}
    \label{fig3}
\end{figure}
\vspace{-0.2cm}

\noindent
Since the cat-caracal dataset uses only 1,200 images for training, there are relatively few regions where the empirical score function is sharp, i.e., where different training points compete.
Therefore, a neural network with sufficient capacity can capture this sharpness, leading to pure memorization, as we see in \cref{fig3} (a).
However, either unconditioning modeling or temperature smoothing (here $T_i=7/\sigma_i$ for $i=1,2,\dotsc,N$ and K=10) can generalize well, illustrated by \cref{fig3} (b) and (c) respectively. We can see obvious generalization in (c), e.g., the first generated cat, which has a caracal face but short ears and gray color. Meanwhile, in \cref{fig3} (d), we compare the generated images with their same collapse training sample (1). Conditioning modeling (2) can only replicate the source image. In contrast, the unconditioning modeling (3) will not fully memorize, and temperature smoothing (4) shows significant generalization. 

\subsubsection{Quantitative Experiments}
\label{exp_ratio}

To empirically support our argument in \cref{sharp_overlap}, we calculate $\gexpand$ under different noise levels using the learned score functions (NN Conditioning and Unconditioning), the empirical conditioning score function, the unconditioning (T=1), and the temperature-based ones (T=10,100,1000). Specifically, we randomly choose 10k data pairs in the cat-caracal dataset, and obtain $x,y,x',y'$ as the assumption in \cref{sharp_overlap}, where we let $||x-y||=0.2\sigma^*$ and $\eta=0.1$.

\begin{figure}
    \centering
    \includegraphics[width=0.95\linewidth]{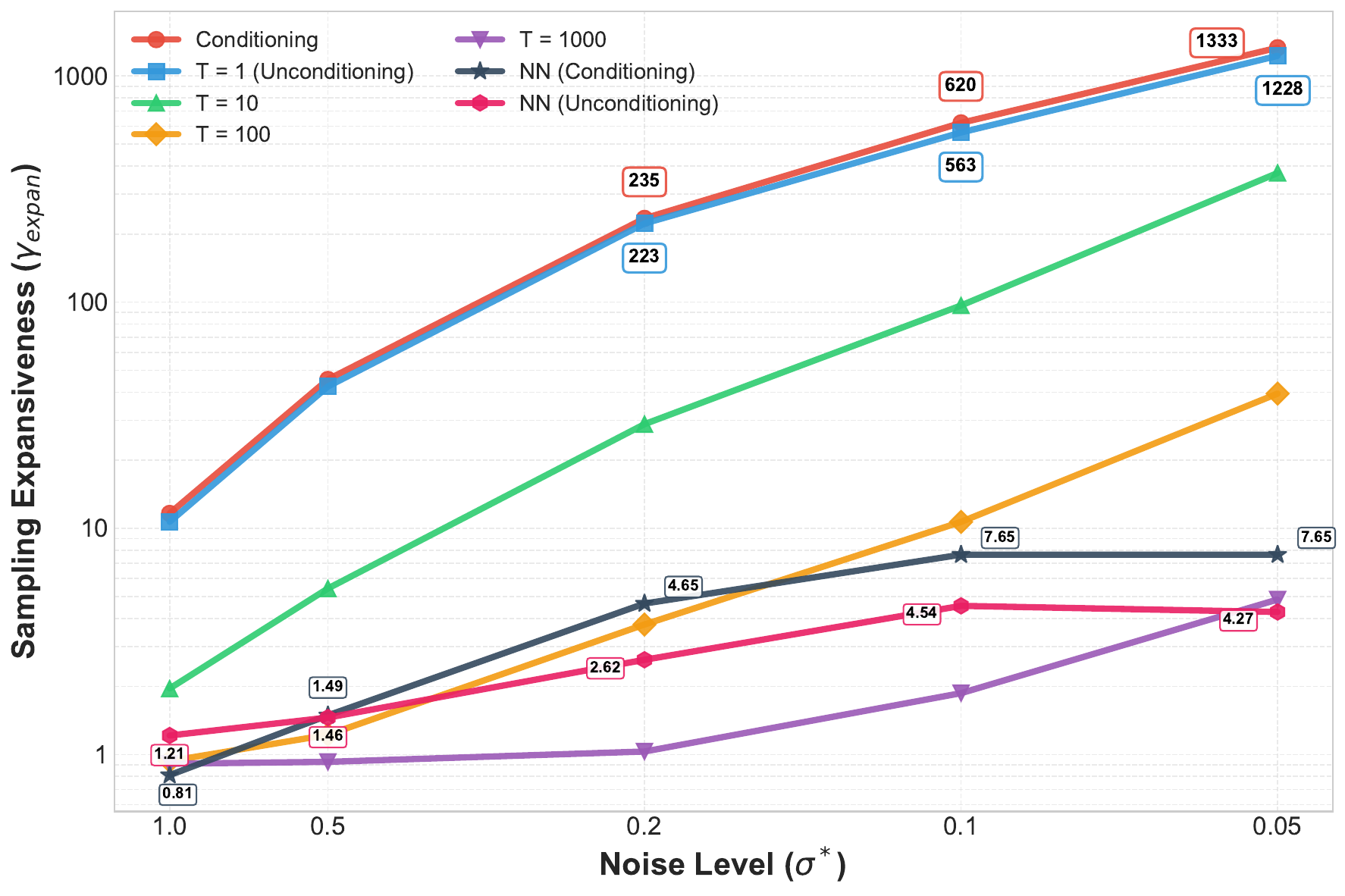}
    \caption{Sampling expansiveness ratio $\gamma_{\text{ex}}$ at different noise levels. Empirical conditioning scores are highly expansive at low noise levels, while unconditioning, temperature smoothing, and NN approximation all yield much smaller $\gamma_{\text{ex}}$.}
    \label{sample_ratio}
\end{figure}

We can see the results in \cref{sample_ratio}. They are consistent with our predictions of the behavior of $\gexpand$ for the various score functions and their NN approximations.
We observe that larger temperature values yield smaller expansion ratios $\gexpand$ for the empirical score function variants, and that the NN behavior is similar to that of the temperature-smoothed empirical score function.
Additionally, as the noise level decreases (toward the right of the chart), $\gexpand$ tends to increase, as the points $x$ and $y$  collapse more rapidly toward their respective (different) training points.
We note too the large difference in behavior between the unconditioned empirical score function and its NN approximation; the latter has much smaller values of $\gexpand$ and therefore better generalization properties.


\subsection{Sample Quality} 

\begin{figure}[!htb]
    \centering
    \includegraphics[width=1.0\linewidth]{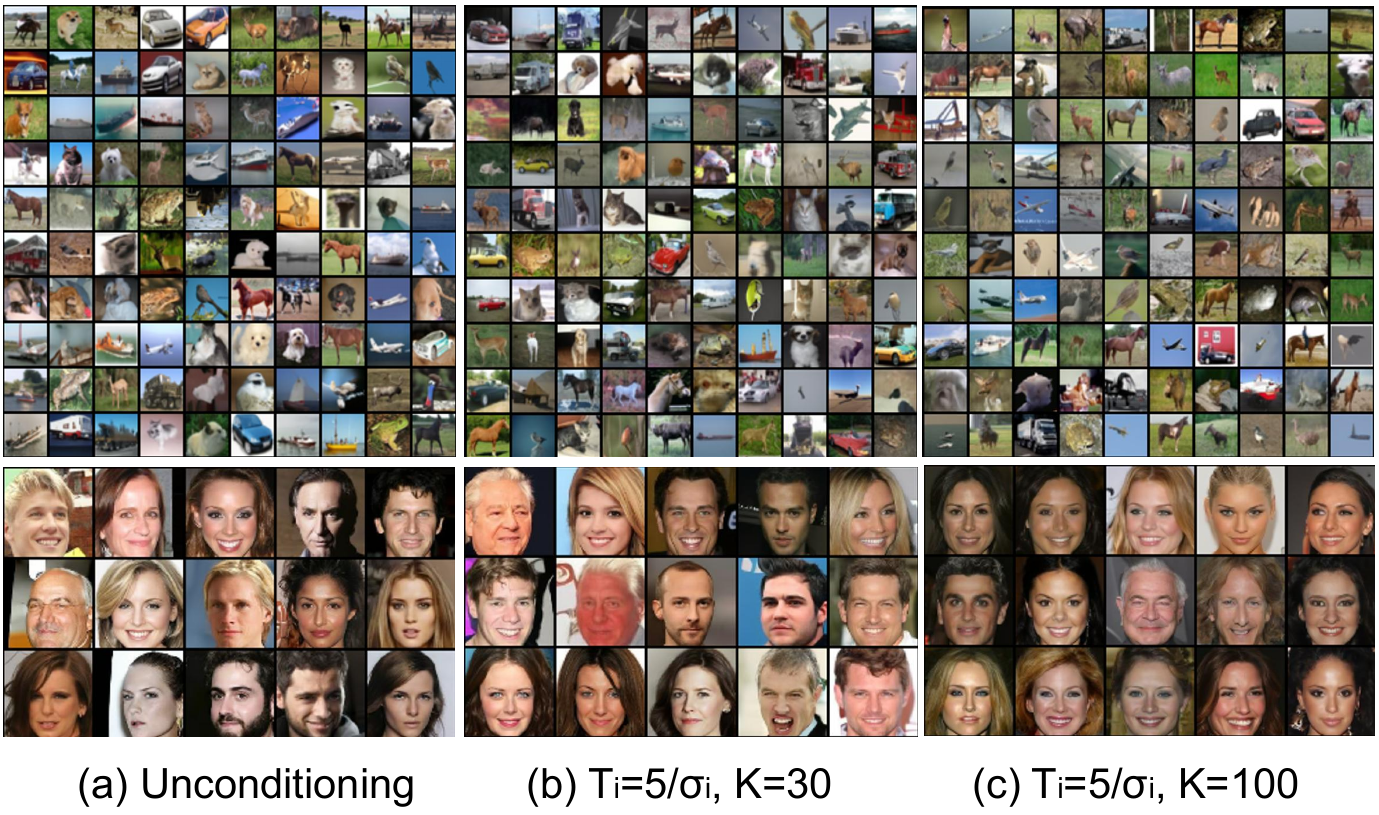}
    \caption{Generated images (top to bottom) for variants of SDE 1K NFE on datasets CIFAR-10 and CelebA. (pixel space KNN)}
    \label{cifar10_celeba}
\end{figure}

To further verify the effectiveness of our methods, we also implement experiments on some commonly used datasets. We first present ablation studies evaluated by FID on the CIFAR-10 and CelebA datasets. As shown in  \cref{tab1}, although the FID score slightly deteriorates at some settings, this is accompanied by observable and meaningful generalization. As illustrated in \cref{cifar10_celeba} (b) and (c), for example, the generated images on CelebA exhibit clear generalization: the red face in (b) and a similar countenance in (c), both of which indicate that the NN has captured the common characteristics of training samples in the local manifold. In practice, we can simply choose an appropriate temperature to achieve strong generalization while maintaining comparable quality.

\begin{table}[htb]
\centering 
\caption{Ablation studies of our smoothing methods on the CIFAR-10 and CelebA datasets. ``$\to$" indicates the performance change from pixel space-based KNN to feature space-based KNN.}
\label{tab1}
\scalebox{0.9}{
\begin{tabular}{lcc}
\toprule 

{\large \textbf{Method}} & { {FID}$(G, \Ttrain)$} & { {FID}$(G, \Ttest)$}\\
\midrule

\multicolumn{3}{l}{\textbf{CIFAR-10 32$\times$32}} \\
\midrule

\textbf{SDE (PC) 1K NFE } & &  \\
Conditioning  & \textbf{6.49} & \textbf{6.56}  \\
Unconditioning & 7.33 & 7.34  \\
$T_i=1/\sigma_i$, K=30 & 8.32 $\to$ 8.08 & 8.25 $\to$ 8.02  \\
$T_i=1/\sigma_i$, K=100 & 8.67 $\to$ 7.97 & 8.61 $\to$ 7.89  \\
$T_i=5/\sigma_i$, K=30 & 8.56 $\to$ 8.12 & 8.60 $\to$ 8.15 \\
$T_i=5/\sigma_i$, K=100 & 13.25 $\to$ 8.35 & 13.41 $\to$ 8.30 \\
$T_i=7/\sigma_i$, K=30 & 21.28 $\to$ 8.26  & 21.34 $\to$ 8.22   \\
$T_i=7/\sigma_i$, K=100 & 50.81 $\to$ 7.96 & 51.08 $\to$ 7.98   \\
\midrule 
\textbf{ODE (PC) 1K NFE} & &  \\
Conditioning  & 8.48  & 8.50  \\
Unconditioning & 8.84 & 8.88   \\
$T_i=1/\sigma_i$, K=30 & 8.75 $\to$ \textbf{8.45}  & 8.67 $\to$ \textbf{8.43} \\

\midrule 

\multicolumn{3}{l}{\textbf{CelebA 64$\times$64}} \\
\midrule

\textbf{SDE (PC) 1K NFE } & &  \\
Conditioning  & 7.25  &  7.81 \\
Unconditioning & \textbf{7.07}& \textbf{7.34}   \\
$T_i=5/\sigma_i$, K=30 &  9.58 $\to$ 8.78 &  9.32 $\to$ 8.54\\
$T_i=5/\sigma_i$, K=100 &  36.63 $\to$ 9.47 & 35.01 $\to$ 9.10 \\
$T_i=10/\sigma_i$, K=30 &  9.88 $\to$ 8.64  &  9.48 $\to$ 8.56 \\
$T_i=10/\sigma_i$, K=100 &   61.97 $\to$ 8.40 &  60.91 $\to$ 8.19  \\

\midrule

\textbf{ODE (PC) 1K NFE } & & \\
Conditioning  &  \textbf{7.65}  & \textbf{7.87}  \\
Unconditioning &  7.71 & 7.88 \\
$T_i=5/\sigma_i$, K=30 & 9.36 $\to$ 8.67 & 9.50 $\to$ 8.80  \\

\bottomrule 
\end{tabular}
}
\end{table}

Moreover, the results in \cref{tab1} show that the feature space-based KNN consistently outperforms the pixel space-based KNN across all settings. Operating in feature space enables more aggressive smoothing (larger temperatures and $K$) while maintaining strong performance. For instance, on CIFAR-10 with $T_i = 7/\sigma_i$ and $K = 100$, pixel-space KNN collapses (FID increases to 50.81), whereas feature-space KNN still attains an FID of 7.96. This observation also supports our hypothesis that the neural network generalizes by smoothing the local manifolds, as lower local curvature (feature space) allows for stronger score smoothing without driving samples off the manifold. The consistent improvements suggest that feature space provides a more appropriate geometric structure for our smoothing methods. These results motivate us to extend our approach to latent diffusion models, which may further enhance generalization while preserving computational efficiency.

In the Appendix, we present complementary results on other high-resolution datasets, including ImageNet 64$\times$64 and CelebA-HQ 256$\times$256. We also provide detailed proofs, training cost analysis, and extensive experiments to validate our theoretical approximations and claims.

\section{Conclusion}
This work establishes a theoretical framework for understanding memorization in diffusion models through the analysis of several variants of the empirical score function and its NN approximations.
It reveals that memorization arises from the dominance of individual training samples in softmax-weighted score functions, while generalization emerges through smoothed weight distributions that enable local manifold exploration. 
By exploring the fundamental properties of the score function, we identify the cause of memorization and propose two smoothing methods --- Noise Unconditioning and Temperature Smoothing --- which elegantly control the concentration of the score function weight and effectively reduce memorization while maintaining high generation quality.

We developed mathematical analysis to explain a theory-practice inconsistency, and made experimental validations, revealing how neural networks achieve remarkable generalization. Our framework rationalizes previously observed memorization phenomena and unifies prior unconditioning approaches from an optimization perspective.

\clearpage

\section{Broader Impact}
\textbf{Generalization Perspective.} Memorization in diffusion models poses significant concerns across multiple domains. In healthcare applications, patient data leakage through generated medical images could violate privacy regulations and compromise sensitive health information. In creative industries, models that memorize copyrighted content risk intellectual property infringement when used for commercial image generation. Our methods address these risks by reducing the likelihood of exact training data replication while preserving the quality and diversity of generated samples, enabling safer deployment of diffusion models in privacy-sensitive and legally regulated contexts.

\textbf{Gradient Ascent Perspective.} Our noise unconditioning reformulates sampling as gradient ascent on a unified distribution, enabling constraints to be integrated via  projected gradient methods. This feature is particularly valuable for applications requiring adherence to physical laws, such as video generation, which  struggles to learn complex physical dynamics from data alone.

\textbf{Intuitive Understanding.} By visualizing the sampling process as moving from large shells (high noise) to small shells (low noise), we offer a natural geometric interpretation that bridges the gap between complex mathematical formulations and intuitive understanding. 
Further, our unconditioning modeling transfers the time-reverse SDE into a familiar optimization process, making diffusion models more accessible to the broader generative community. 
This geometric intuition may also facilitate practical debugging, hyperparameter tuning, and model design decisions.

\section*{Acknowledgement}
The first and third authors wish to acknowledge support from NSF grants 2224213 and 2023239. We thank Qin Li and Sixu Li for valuable discussions.


\bibliography{main}

@String(CVPR   = {IEEE/CVF Conference on Computer Vision and Pattern Recognition (CVPR)})

@String(ICCV   = {IEEE/CVF International Conference on Computer Vision (ICCV)})

@String(ECCV   = {European Conference on Computer Vision (ECCV)})

@String(ICML   = {International Conference on Machine Learning (ICML)})

@String(NIPS   = {Advances in Neural Information Processing Systems (NeurIPS)})

@String(ICLR   = {International Conference on Learning Representations (ICLR)})

@article{ho2020denoising,
  title        = {Denoising diffusion probabilistic models},
  author       = {Ho, Jonathan and Jain, Ajay and Abbeel, Pieter},
  journal      = NIPS,
  year         = {2020}
}

@article{song2020score,
  title        = {Score-based generative modeling through stochastic differential equations},
  author       = {Song, Yang and Sohl-Dickstein, Jascha and Kingma, Diederik P and Kumar, Abhishek and Ermon, Stefano and Poole, Ben},
  journal      = ICLR,
  year         = {2021}
}

@article{song2019generative,
  title        = {Generative modeling by estimating gradients of the data distribution},
  author       = {Song, Yang and Ermon, Stefano},
  journal      = NIPS,
  year         = {2019}
}

@article{li2024good,
  title        = {A good score does not lead to a good generative model},
  author       = {Li, Sixu and Chen, Shi and Li, Qin},
  journal      = {arXiv preprint arXiv:2401.04856},
  year         = {2024}
}

@inproceedings{rombach2022high,
  title        = {High-resolution image synthesis with latent diffusion models},
  author       = {Rombach, Robin and Blattmann, Andreas and Lorenz, Dominik and Esser, Patrick and Ommer, Bj{\"o}rn},
  booktitle    = CVPR,
  year         = {2022}
}

@article{esser2024scaling,
  title        = {Scaling rectified flow transformers for high-resolution image synthesis},
  author       = {Esser, Patrick and Kulal, Sumith and Blattmann, Andreas and Entezari, Rahim and M{\"u}ller, Jonas and Saini, Harry and Levi, Yam and Lorenz, Dominik and Sauer, Axel and Boesel, Frederic and others},
  journal      = ICML,
  year         = {2024}
}

@article{watson2023novo,
  title        = {De novo design of protein structure and function with RFdiffusion},
  author       = {Watson, Joseph L and Juergens, David and Bennett, Nathaniel R and Trippe, Brian L and Yim, Jason and Eisenach, Helen E and Ahern, Woody and Borst, Andrew J and Ragotte, Robert J and Milles, Lukas F and others},
  journal      = {Nature},
  year         = {2023},
  publisher    = {Nature Publishing Group UK London}
}

@article{wu2024protein,
  title        = {Protein structure generation via folding diffusion},
  author       = {Wu, Kevin E and Yang, Kevin K and van den Berg, Rianne and Alamdari, Sarah and Zou, James Y and Lu, Alex X and Amini, Ava P},
  journal      = {Nature Communications},
  year         = {2024},
  publisher    = {Nature Publishing Group UK London}
}

@article{fefferman2016testing,
  title        = {Testing the manifold hypothesis},
  author       = {Fefferman, Charles and Mitter, Sanjoy and Narayanan, Hariharan},
  journal      = {Journal of the American Mathematical Society},
  volume       = {29},
  number       = {4},
  pages        = {983--1049},
  year         = {2016}
}

@article{vincent2011connection,
  title        = {A connection between score matching and denoising autoencoders},
  author       = {Vincent, Pascal},
  journal      = {Neural Computation},
  volume       = {23},
  number       = {7},
  pages        = {1661--1674},
  year         = {2011},
  publisher    = {MIT Press}
}

@article{somepalli2023understanding,
  title        = {Understanding and mitigating copying in diffusion models},
  author       = {Somepalli, Gowthami and Singla, Vasu and Goldblum, Micah and Geiping, Jonas and Goldstein, Tom},
  journal      = NIPS,
  year         = {2023}
}

@inproceedings{wen2024detecting,
  title        = {Detecting, explaining, and mitigating memorization in diffusion models},
  author       = {Wen, Yuxin and Liu, Yuchen and Chen, Chen and Lyu, Lingjuan},
  booktitle    = ICLR,
  year         = {2024}
}

@inproceedings{carlini2023extracting,
  title        = {Extracting training data from diffusion models},
  author       = {Carlini, Nicolas and Hayes, Jamie and Nasr, Milad and Jagielski, Matthew and Sehwag, Vikash and Tramer, Florian and Balle, Borja and Ippolito, Daphne and Wallace, Eric},
  booktitle    = {USENIX Security Symposium (USENIX Security 23)},
  year         = {2023}
}

@article{gu2023memorization,
  title        = {On memorization in diffusion models},
  author       = {Gu, Xiangming and Du, Chao and Pang, Tianyu and Li, Chongxuan and Lin, Min and Wang, Ye},
  journal      = {arXiv preprint arXiv:2310.02664},
  year         = {2023}
}

@inproceedings{ren2024unveiling,
  title        = {Unveiling and mitigating memorization in text-to-image diffusion models through cross attention},
  author       = {Ren, Jie and Li, Yaxin and Zeng, Shenglai and Xu, Han and Lyu, Lingjuan and Xing, Yue and Tang, Jiliang},
  booktitle    = ECCV,
  year         = {2024},
}

@article{bonnaire2025diffusion,
  title        = {Why Diffusion Models Don't Memorize: The Role of Implicit Dynamical Regularization in Training},
  author       = {Bonnaire, Tony and Urfin, Rapha{\"e}l and Biroli, Giulio and M{\'e}zard, Marc},
  journal      = NIPS,
  year         = {2025}
}

@inproceedings{yoon2023diffusion,
  title        = {Diffusion probabilistic models generalize when they fail to memorize},
  author       = {Yoon, TaeHo and Choi, Joo Young and Kwon, Sehyun and Ryu, Ernest K},
  booktitle    = {ICML Workshop on Structured Probabilistic Inference and Generative Modeling},
  year         = {2023}
}

@article{karras2022elucidating,
  title        = {Elucidating the design space of diffusion-based generative models},
  author       = {Karras, Tero and Aittala, Miika and Aila, Timo and Laine, Samuli},
  journal      = NIPS,
  year         = {2022}
}

@article{krizhevsky2009learning,
  title        = {Learning multiple layers of features from tiny images},
  author       = {Krizhevsky, Alex and Hinton, Geoffrey and others},
  year         = {2009},
  publisher    = {University of Toronto}
}

@inproceedings{liu2015deep,
  title        = {Deep learning face attributes in the wild},
  author       = {Liu, Ziwei and Luo, Ping and Wang, Xiaogang and Tang, Xiaoou},
  booktitle    = ICCV,
  year         = {2015}
}

@article{deng2009imagenet,
  title        = {ImageNet: A large-scale hierarchical image database},
  author       = {Deng, Jia and Dong, Wei and Socher, Richard and Li, Li-Jia and Li, Kai and Fei-Fei, Li},
  journal      = CVPR,
  year         = {2009},
}

@article{song2020improved,
  title        = {Improved techniques for training score-based generative models},
  author       = {Song, Yang and Ermon, Stefano},
  journal      = NIPS,
  year         = {2020}
}

@article{sun2025noise,
  title        = {Is Noise Conditioning Necessary for Denoising Generative Models?},
  author       = {Sun, Qiao and Jiang, Zhicheng and Zhao, Hanhong and He, Kaiming},
  journal      = ICML,
  year         = {2025}
}

@article{hyvarinen2005estimation,
  title        = {Estimation of non-normalized statistical models by score matching},
  author       = {Hyv{\"a}rinen, Aapo and Dayan, Peter},
  journal      = {Journal of Machine Learning Research},
  volume       = {6},
  number       = {4},
  year         = {2005}
}

@article{heusel2017gans,
  title        = {GANs trained by a two time-scale update rule converge to a local Nash equilibrium},
  author       = {Heusel, Martin and Ramsauer, Hubert and Unterthiner, Thomas and Nessler, Bernhard and Hochreiter, Sepp},
  journal      = NIPS,
  year         = {2017}
}

@article{zhang2016understanding,
  title        = {Understanding deep learning requires rethinking generalization},
  author       = {Zhang, Chiyuan and Bengio, Samy and Hardt, Moritz and Recht, Benjamin and Vinyals, Oriol},
  journal      = ICLR,
  year         = {2017}
}

@article{sohl2015deep,
  title        = {Deep unsupervised learning using nonequilibrium thermodynamics},
  author       = {Sohl-Dickstein, Jascha and Weiss, Eric and Maheswaranathan, Niru and Ganguli, Surya},
  journal      = ICML,
  year         = {2015}
}

@article{papamakarios2021normalizing,
  title        = {Normalizing flows for probabilistic modeling and inference},
  author       = {Papamakarios, George and Nalisnick, Eric and Rezende, Danilo Jimenez and Mohamed, Shakir and Lakshminarayanan, Balaji},
  journal      = {Journal of Machine Learning Research},
  volume       = {22},
  number       = {57},
  pages        = {1--64},
  year         = {2021}
}

@article{liu2022flow,
  title        = {Flow straight and fast: Learning to generate and transfer data with rectified flow},
  author       = {Liu, Xingchao and Gong, Chengyue and Liu, Qiang},
  journal      = {arXiv preprint arXiv:2209.03003},
  year         = {2022}
}

@article{song2023consistency,
  title        = {Consistency models},
  author       = {Song, Yang and Dhariwal, Prafulla and Chen, Mark and Sutskever, Ilya},
  journal      = ICML,
  year         = {2023}
}

@article{song2023improved,
  title        = {Improved techniques for training consistency models},
  author       = {Song, Yang and Dhariwal, Prafulla},
  journal      = ICLR,
  year         = {2024}
}

@article{lu2024simplifying,
  title        = {Simplifying, stabilizing and scaling continuous-time consistency models},
  author       = {Lu, Cheng and Song, Yang},
  journal      = ICLR,
  year         = {2025}
}

@article{albergo2025stochastic,
  title        = {Stochastic interpolants: A unifying framework for flows and diffusions},
  author       = {Albergo, Michael S and Boffi, Nicholas M and Vanden-Eijnden, Eric},
  journal      = {Journal of Machine Learning Research},
  volume       = {26},
  pages        = {1--80},
  year         = {2025}
}

@article{lipman2022flow,
  title        = {Flow matching for generative modeling},
  author       = {Lipman, Yaron and Chen, Ricky TQ and Ben-Hamu, Heli and Nickel, Maximilian and Le, Matt},
  journal      = {arXiv preprint arXiv:2210.02747},
  year         = {2022}
}

@article{loshchilov2017decoupled,
  title        = {Decoupled weight decay regularization},
  author       = {Loshchilov, Ilya and Hutter, Frank},
  journal      = ICLR,
  year         = {2019}
}

@article{dhariwal2021diffusion,
  title        = {Diffusion models beat GANs on image synthesis},
  author       = {Dhariwal, Prafulla and Nichol, Alexander},
  journal      = NIPS,
  year         = {2021}
}

@article{zhang2025gradient,
  title        = {A Gradient Guided Diffusion Framework for Chance Constrained Programming},
  author       = {Zhang, Boyang and Wang, Zhiguo and Liu, Ya-Feng},
  journal      = NIPS,
  year         = {2025}
}

@article{karras2017progressive,
  title        = {Progressive growing of GANs for improved quality, stability, and variation},
  author       = {Karras, Tero and Aila, Timo and Laine, Samuli and Lehtinen, Jaakko},
  journal      = ICLR,
  year         = {2018}
}

@article{krishnamoorthy2023diffusion,
  title        = {Diffusion models for black-box optimization},
  author       = {Krishnamoorthy, Siddarth and Mashkaria, Satvik Mehul and Grover, Aditya},
  journal      = ICML,
  year         = {2023}
}
\bibliographystyle{icml2026}

\clearpage
\onecolumn

\tableofcontents
\newpage  
\appendix

\section*{Preface}
\addcontentsline{toc}{section}{Preface}

We give some background on the development of our framework and the potential  it holds for better understanding of diffusion models. 

Besides being inspired by the theory–practice inconsistency discussed in the main paper, our framework is also motivated by the empirical difficulty of explicit score matching in practice. 
In high-dimensional spaces, the data distribution generally occupies a limited region under the Manifold Hypothesis \cite{fefferman2016testing}, which means that standard score matching \cite{vincent2011connection} tends to be ineffective due to unstable or highly inaccurate gradient estimation in low-density regions. 
Unlike the solution proposed in NCSN \cite{song2019generative} --- commonly known as diffusion models (VE-SDE \cite{song2020score}) --- which ensures that each marginal score function is learned by samples within its corresponding shell (a high-density region at a given noise level), our unconditioning modeling unifies all noise-perturbed distributions into a single Gaussian mixture and effectively spreads the support of the training objective across all possible sampling regions. 
However, this noise-perturbed Gaussian mixture distribution is not our final target; rather, the distribution of its centers is. 
Regions closer to the centers exhibit higher probability density (same probability mass but much smaller volume), so the score function of this perturbed distribution will eventually guide samples toward the centers --- a gradient ascent process.

Our unconditioning framework is essentially \textit{a parallel solution} to the conditioning diffusion models, both of which emerge from addressing the limitations of high-dimensional score matching. 
Based on the inherent properties of the score function of a Gaussian mixture, we show that unconditioning modeling tends to exhibit smoother score-function weights across different centers and, in this sense, better generalization capabilities in comparison to conditioning modeling. 
Meanwhile, we identify the softmax weighting in the empirical score function as providing a natural mechanism (temperature) to further control this smoothness and enhance generalization.

Our framework rephrases much of the diffusion theory in terms of basic optimization and high-dimensional geometry, \emph{largely} avoiding complex conditional probability or partial differential equation derivations. 
This perspective relies only on elementary optimization concepts and mathematical derivations, making the core ideas accessible to a broad range of practitioners and easier to reason about, which in turn lowers the barrier for developing new intuitions and extensions.

\paragraph{A Metaphor.}
We can imagine the generative process as a mountaineering expedition: A \textit{hiker} (a sample point $\boldsymbol{x}$), starting on a flat, empty plain (of pure noise), must climb to any \textit{peak} (a valid data sample).

In the traditional diffusion models, the \textit{guide} (a time-conditional network ${s_{\theta}(x, t)}$) uses altitude-specific guidebooks, each only valid for a certain elevation (the time step $t$). This creates a narrow view (score function weight dominance) that leads the hiker more or less directly to a  nearby peak, missing other possible destinations.

Our unconditioning modeling trains a guide (a time-independent network ${s_{\theta}(x)}$) with a single, complete map of the entire mountain range. This guide can see the complex network of valleys, ridges, and passes connecting all peaks, understanding diverse pathways to multiple destinations (via a smoother score-function weight in which several nearby peaks contribute).

We enhance exploration with temperature smoothing, which works like terrain engineering that builds \textit{bridges} between isolated peaks. Appropriately higher temperatures create denser bridge networks, which enables the hiker to move smoothly between and around peaks, discovering new highlands that were not on the original map: generalization!

\clearpage
\section*{Literature Review}
\addcontentsline{toc}{section}{Literature Review}

\textbf{Memorization and Generalization.}
Recent work has revealed memorization behaviors in diffusion models, where generated samples can closely replicate training data \cite{carlini2023extracting, somepalli2023understanding, wen2024detecting}. \cite{carlini2023extracting} demonstrated systematic extraction of training images from large-scale models like Stable Diffusion \cite{rombach2022high}, raising privacy and copyright concerns. \cite{somepalli2023understanding} analyzed the connection between data duplication and memorization, proposing mitigation strategies through caption diversification. To investigate the causes of memorization, studies like \cite{gu2023memorization} provided empirical evidence showing that the optimal solution of denoising score matching can only generate training data copies; \cite{yoon2023diffusion} proposed the memorization–generalization dichotomy, arguing that generalization occurs when models fail to memorize, contrasting with overfitting in supervised learning \cite{zhang2016understanding}. Recent work \cite{bonnaire2025diffusion} further elaborates on how neural networks learn the empirical score function: the model first fits an approximation to the population score and later converges to the highly irregular empirical score that leads to exact memorization.

On the one hand, these works, from the model-learning perspective, show that memorization is highly related to model capacity, dataset size, and training time. On the other hand, our framework approaches the problem from a complementary angle by explicitly characterizing the properties of the empirical score function itself: we show that its complexity grows with the number and arrangement of training samples, because the overlap regions of Gaussian shells increase, which not only offers a more complete understanding of memorization in diffusion models but also naturally suggests concrete score-smoothing interventions to improve generalization.

\noindent
\textbf{Noise Conditioning and Unconditioning.}
Noise conditioning has been a core component of modern generative models since early denoising-based methods \cite{sohl2015deep}, then popularized in DDPMs \cite{ho2020denoising} and Noise Conditional Score Networks (NCSNs) \cite{song2019generative}. It has become standard practice, adopted by flow matching \cite{papamakarios2021normalizing, lipman2022flow, liu2022flow, albergo2025stochastic}, consistency models \cite{song2023consistency}, and distillation techniques \cite{song2023improved, lu2024simplifying}. A few recent works have also explored \emph{noise unconditioning}---removing the explicit noise-level input. \cite{song2020improved} first showed that a score network with a preconditioned objective can be trained without conditioning and still generate reasonable samples using Annealing Langevin Dynamics. \cite{sun2025noise} further demonstrated that such unconditioning works across DDPMs and flow models. These works, however, mainly establish that unconditioning \emph{can work} in practice, without clarifying its theory.

\vspace{-0.05cm}
Our paper goes beyond demonstrating that noise unconditioning can work by providing a theoretical framework for understanding what it essentially learns and how it should be correctly used. Specifically, we prove that the standard unconditioning loss \cref{uncon_loss} corresponds to explicit score matching on a unified Gaussian mixture $\pMN$ (\cref{A.6}), and use this perspective to characterize how unconditioning smooths score weights and improves generalization. At the same time, this framework explains why diffusion ODE and SDE samplers behave very differently after removing the noise condition (\cref{A.7}).

This unified-view also clarifies an important point compared to \cite{sun2025noise}: the claim that “given the same input $x$, the outputs of the NN trained by noise conditioning and unconditioning are almost the same” does not generally hold, since in unconditioning the score field couples \emph{all} noise levels rather than a fixed one, and the prescribed noise level is not always aligned with the optimal noise scale determined by the closest training point to $x$. Consistently with our theory, all our ablations that directly compare conditioning vs. unconditioning modelings (e.g., qualitative samples in \cref{fig3}, expansiveness measurements in \cref{fig4}, and memorization statistics in \cref{B.3.2}) show clear and systematic differences between the two, especially at low noise levels. This conclusion also reveals why the diffusion ODE sampler cannot work with direct replacement of the unconditioned model, which is also observed in \cite{sun2025noise} but unexplainable there.

\clearpage

\section*{Notation Table}
\addcontentsline{toc}{section}{Notation Table}
We summarize here the main symbols used throughout the paper.

\begin{table}[!htb]
\centering
\resizebox{1\linewidth}{!}{
\begin{tabular}{ll}
\toprule
Symbol & Meaning \\
\midrule
$x \in \mathbb{R}^d$ & Data / sample in the ambient (pixel) space. \\
$\mu_j \in \mathbb{R}^d$ & The $j$-th training sample (Gaussian center), $j=1,\dots,M$. \\
$M$ & Number of training samples used to form the empirical distribution. \\
$N$ & Number of noise levels. \\
$t \in [0,T]$ & Diffusion time. \\
$x_t$ & State at time $t$ in the forward / backward process. \\
$x_n$ & Sample at discrete step $n$ of the sampler. \\
\midrule
$p_{\mathrm{data}}(x)$ & True data distribution. \\
$\hat p(x)$ & Empirical data distribution: $\hat p(x) = \frac{1}{M}\sum_{j=1}^M \delta(x-\mu_j)$. \\
$\hat p_i(x)$ & Empirical marginal at level $\sigma_i$: $\hat p_i(x)=\frac{1}{M}\sum_{j=1}^M N(x;\mu_j,\sigma_i^2 I)$ (i.e., $\hat p_{\sigma_i}(x)$). \\
$ p_\sigma(x)$ & The distribution of the noise levels. \\
$N(x;\mu,\sigma^2 I)$ & Isotropic Gaussian with mean $\mu$, covariance $\sigma^2 I$. \\
$\sigma \in [\sigma_{\min},\sigma_{\max}]$ & Noise scale (standard deviation) in VE-SDE. \\
$\{\sigma_i\}_{i=1}^N$ & Discrete noise schedule, typically log-uniform. \\
$\pMN(x)$ & Unified Gaussian mixture: $\frac{1}{Z_{MN}} \sum_{j=1}^M \sum_{i=1}^N \lambda(\sigma_i) N(x;\mu_j,\sigma_i^2 I)$. \\
$Z_{MN}$ & Normalization constant of $\pMN(x)$. \\
$\lambda(\sigma_i)$ & Noise level weight; in our experiments, $\lambda(\sigma_i)=\sigma_i^2$. \\
\midrule
$s_\theta(x,\sigma)$ & NN score in the noise conditioning modeling, approx.\ $\nabla_x \log \hat p_\sigma(x)$. \\
$s_\theta(x)$ & NN score in the unconditioning modeling (no noise input). \\
$\nabla_x \log p(x)$ & Score function of distribution $p(x)$. \\
$w_{ij}(x)$ & Weight of component $(\mu_j,\sigma_i)$ at $x$ (posterior responsibility). \\
$f(x,\mu_j,\sigma_i)$ & Pre-softmax log-weight: $-(d-2)\ln\sigma_i - \frac{\|x-\mu_j\|^2}{2\sigma_i^2}$. \\
\midrule
$\mu^\star$ & Nearest training point to $x$: $\mu^\star = \arg\min_j \|x-\mu_j\|^2$. \\
$\sigma_{\mathrm{opt}}$ & ``Optimal'' noise for $(x,\mu^\star)$; $\sigma_{\mathrm{opt}}\approx \|x-\mu^\star\|/\sqrt{d}$. \\
$\sigma_j^\star$ & Discrete optimal shell for $\mu_j$: $\sigma_j^\star = \arg\min_{\sigma_i}\big|\tfrac{\sigma_{\mathrm{opt}}^2}{\sigma_i^2}-1\big|$. \\
$w_j^\star(x)$ & Dominant weight for center $\mu_j$ under its optimal shell $\sigma_j^\star$. \\
\midrule
$T>0$ & Temperature for smoothing softmax weights (scalar or per-shell). \\
$T_i$ & Temperature on shell $i$ in the temperature vector $T\in\mathbb{R}^N$. \\
$T_j^\star$ & Temperature at the optimal shell of center $\mu_j$ ($T_j^\star = T_i$ with $\sigma_i=\sigma_j^\star$). \\
$w_j^\star(x;T)$ & Temperature-smoothed dominant weight for center $\mu_j$ at $x$. \\
\midrule
$\mu(x,t)$ & Drift of the forward SDE (zero for VE-SDE). \\
$\sigma(t)$ & Diffusion scale in VE-SDE. \\
$dx_t = \mu(x_t,t)\,dt + \sigma(t)\,dw_t$ & Forward SDE. \\
$dx_t = \big(\mu(x_t,t) - \tfrac12 \sigma(t)^2 \nabla_x \log p_t(x_t)\big)\,dt$ & Probability flow ODE. \\
$\eta$ & Step size in gradient flow / ODE sampling. \\
$\sigma_n$ & Pre-scheduled noise at step $n$. \\
$\sigma_n^\star$ & ``Actual'' noise at step $n$: $(\sigma_n^\star)^2 := \min_j \|x_n-\mu_j\|^2/d$. \\
\midrule
$L_c$ & Noise-conditioned denoising score matching loss. \\
$L_u$ & Unconditioning score matching loss (explicit score matching on $\pMN$). \\
$L_T$ & Temperature-based score matching loss using a KNN-based approximation. \\
$\sigma_{\mathrm{collapse}}$ & Noise threshold below which temperature smoothing is applied. \\
\midrule
$\gamma_{\mathrm{ex}}(x,y)$ & Local sampling expansiveness: $\|y'-x'\| / \|y-x\|$ after one update. \\
$H(x; T)$ & Hessian of the log distribution: $H(x;T):=\nabla^2_{x,x} \log \pMN(x,T)$. \\
$\lambda_{\max}(x;T)$ & Largest eigenvalue of $H(x; T)$ (controls worst-case local expansion). \\
$\mathrm{FID}(G,T_{\mathrm{train}})$ & Fr\'echet Inception Distance between generated and training samples. \\
$\mathrm{FID}(G,T_{\mathrm{test}})$ & Fr\'echet Inception Distance between generated and test samples. \\
\bottomrule
\end{tabular}
}
\end{table}

\clearpage

\section{Proofs}

In this section, we provide details of proofs of the results in the paper.

\subsection{Unified Distribution as a Gaussian Mixture}
\label{A.1}

\subsubsection*{Step 1: Defining the Target Distribution}

We define a unified distribution $p(\tilde{x})$ by perturbing the data distribution with multiple noise levels $\sigma$, weighted by a function $\lambda(\sigma)$:
\[
p(\tilde{x}) = \int_{\sigma_{\min}}^{\sigma_{\max}} \frac{\lambda(\sigma)}{Z} \int_{x} p(\tilde{x} \mid x, \sigma) p_{\text{data}}(x) p(\sigma) \, dx \, d\sigma,
\]
where:
\begin{itemize}
    \item $\lambda(\sigma)$ is a weighting function (e.g., $\lambda(\sigma) := \sigma^2$);
    \item $Z = \int_{\sigma_{\min}}^{\sigma_{\max}} \lambda(\sigma) p(\sigma) d\sigma$ is the normalization constant;
    \item $p(\tilde{x} \mid x, \sigma) = \mathcal{N}(\tilde{x}; x, \sigma^2\mathbf{I})$ is the perturbed Gaussian component;
    \item $p_{\text{data}}(x)$ is the data distribution;
    \item $p(\sigma) = \frac{1}{\sigma \ln(\sigma_{\text{max}} / \sigma_{\text{min}})}$ for $\sigma \in [\sigma_{\text{min}}, \sigma_{\text{max}}]$ is the log-uniform prior.
\end{itemize}

\subsubsection*{Step 2: Discretizing the Data Distribution}

We approximate the data distribution using $M$ discrete training samples $\{\mu_j\}_{j=1}^M$:
\[
p_{\text{data}}(x) \approx \frac{1}{M} \sum_{j=1}^M \delta(x - \mu_j).
\]

The inner integral over $x$ becomes:
\[
\int_{x} p(\tilde{x} \mid x, \sigma) p_{\text{data}}(x) \, dx \approx \frac{1}{M} \sum_{j=1}^M \mathcal{N}(\tilde{x}; \mu_j, \sigma^2\mathbf{I}).
\]

Thus,
\[
p(\tilde{x}) \approx \int_{\sigma_{\min}}^{\sigma_{\max}} \frac{\lambda(\sigma)}{ZM} \left( \sum_{j=1}^M \mathcal{N}(\tilde{x}; \mu_j, \sigma^2\mathbf{I}) \right) p(\sigma) \, d\sigma.
\]

\subsubsection*{Step 3: Discretizing the Noise Levels}

Using our approximation to the uniform distribution on $[\sigma_{\min}, \sigma_{\max}]$ based on the log-uniformly distributed points $\sigma_i$, $i=1,2,\dotsc,N$, we can write 
\[
\int_{\sigma_{\text{min}}}^{\sigma_{\text{max}}} \frac{\lambda(\sigma)}{Z} p(\sigma) f(\sigma) \, d\sigma \approx \frac{1}{N} \sum_{i=1}^N \frac{\lambda(\sigma_i)}{Z} f(\sigma_i),
\]
where $f$ is a general function and \(\frac{1}{N}\) reflects the uniform weighting of the \(N\) sampled noise levels.
By applying this to our distribution, we obtain
\[
\pMN (\tilde{x})  = p(\tilde{x}) \approx \frac{1}{ZMN} \sum_{i=1}^N \sum_{j=1}^M \lambda(\sigma_i) \mathcal{N}(\tilde{x}; \mu_j, \sigma_i^2 \mathbf{I}).
\]

Therefore, the unified distribution $\pMN$ is a weighted Gaussian mixture with $M \times N$ components, where each component has weight $\frac{\lambda(\sigma_i)}{ZMN}$, mean $\mu_j$, and covariance $\sigma_i^2 \mathbf{I}$.

\medskip
\subsection{Score Function Calculation}
\label{A.2}

First take the logarithm:
\[
\log \pMN(x) = - \log(ZMN) + \log \left(\sum_{j=1}^{M} \sum_{i=1}^{N} \lambda(\sigma_i) \mathcal{N}(x; \mu_j, \sigma_i^2 \mathbf{I})\right),
\]
then the gradient:
\[
\nabla_{x} \log \pMN(x) = \nabla_{x} \log \left(\sum_{j=1}^{M} \sum_{i=1}^{N} \lambda(\sigma_i) \mathcal{N}(x; \mu_j, \sigma_i^2 \mathbf{I})\right).
\]

Then using the chain rule for the gradient of \(\log(\cdot)\):
\begin{equation} \label{eq:ks1}
    \nabla_{x} \log \left(\sum_{j=1}^{M} \sum_{i=1}^{N} \lambda(\sigma_i) \mathcal{N}(x; \mu_j, \sigma_i^2 \mathbf{I})\right) = \frac{\nabla_{x} \sum_{j=1}^{M} \sum_{i=1}^{N} \lambda(\sigma_i) \mathcal{N}(x; \mu_j, \sigma_i^2 \mathbf{I})}{\sum_{j=1}^{M} \sum_{i=1}^{N} \lambda(\sigma_i) \mathcal{N}(x; \mu_j, \sigma_i^2 \mathbf{I})}.
\end{equation}
The multivariate ($d$ dimensional) Gaussian PDF is
\[
\mathcal{N}(x; \mu_j, \sigma_i^2 \mathbf{I}) = \frac{1}{(2\pi \sigma_i^2)^{d/2}} \exp\left(-\frac{\|x - \mu_j\|^2}{2\sigma_i^2}\right),
\]
whose gradient with respect to \(x\) is
\[
\nabla_{x} \mathcal{N}(x; \mu_j, \sigma_i^2 \mathbf{I}) = \mathcal{N}(x; \mu_j, \sigma_i^2 \mathbf{I}) \cdot \left(-\frac{x - \mu_j}{\sigma_i^2}\right).
\]
Substituting the gradient of the Gaussian into the numerator of \eqref{eq:ks1}, we have:
\[
\nabla_{x} \sum_{j=1}^{M} \sum_{i=1}^{N} \lambda(\sigma_i) \mathcal{N}(x; \mu_j, \sigma_i^2 \mathbf{I}) = \sum_{j=1}^{M} \sum_{i=1}^{N} \lambda(\sigma_i) \mathcal{N}(x; \mu_j, \sigma_i^2 \mathbf{I}) \cdot \left(-\frac{x - \mu_j}{\sigma_i^2}\right).
\]
We can thus rewrite the score function \eqref{eq:ks1} as follows:
\[
\nabla_{x} \log \pMN(x) = \frac{\sum_{j=1}^{M} \sum_{i=1}^{N} \lambda(\sigma_i) \mathcal{N}(x; \mu_j, \sigma_i^2 \mathbf{I}) \cdot \left(-\frac{x - \mu_j}{\sigma_i^2}\right)}{\sum_{j=1}^{M} \sum_{i=1}^{N} \lambda(\sigma_i) \mathcal{N}(x; \mu_j, \sigma_i^2 \mathbf{I})}.
\]
Define the normalized weight for each Gaussian component as follows:
\[
w_{ij}(x) := \frac{\lambda(\sigma_i) \mathcal{N}(x; \mu_j, \sigma_i^2 \mathbf{I})}{\sum_{l=1}^{M} \sum_{k=1}^{N} \lambda(\sigma_k) \mathcal{N}(x; \mu_l, \sigma_k^2 \mathbf{I})}.
\]
This represents the posterior probability \(p(\mu_j, \sigma_i \mid x)\), i.e., the probability that the sample \(x\) was generated by the \(j\)-th center \(\mu_j\) and the \(i\)-th variance level \(\sigma_i^2\).
Using \(w_{ij}(x)\), the score function can be written more simply as follows:
\[
\nabla_{x} \log \pMN(x) = -\sum_{j=1}^{M} \sum_{i=1}^{N} w_{ij}(x) \frac{x - \mu_j}{\sigma_i^2}.
\]
This form highlights that the score function of a Gaussian mixture is a weighted average of the score functions of its components, with weights given by the posterior probabilities \(w_{ij}(x)\).

\subsection{Proposition 1 (Gaussian Shell Distribution)}
\label{A.3}

\textbf{Proposition 1.} Let $X \sim \mathcal{N}(0, \sigma^2\mathbf{I})$ and $R=||X||_2$. When \( d \to \infty \), we have \( R \xrightarrow{P} \sigma \sqrt{d} \) and \( R  \xrightarrow{d} \mathcal{N}(\sigma\sqrt{d}, \sigma^2/2) \). Namely in high dimensions, most of the sampling points of a Gaussian will concentrate on a thin shell with radius approximately \( \sigma \sqrt{d} \) and thickness \( 3\sqrt{2}\sigma \), namely $R \in \left [\sigma \sqrt{d}-\frac{3\sigma}{\sqrt{2}} , \sigma \sqrt{d}+\frac{3\sigma}{\sqrt{2}}   \right ] $.

We examine the distribution of the squared radius $R^2 = \|X\|^2$. When $X \sim \mathcal{N}(0, \sigma^2\mathbf{I})$ in $d$ dimensions, the squared radius follows a scaled chi-squared distribution:

$$R^2 = \|X\|^2 \sim \sigma^2\chi^2_d$$

\noindent
where $\chi^2_d$ denotes a chi-squared distribution with $d$ degrees of freedom.
The chi-squared distribution $\chi^2_d$ has mean $d$ and variance $2d$. Therefore, $R^2$ has mean $\mu_{R^2} = d\sigma^2$ and variance $\text{Var}(R^2) = 2d\sigma^4$.

Applying Chebyshev's inequality to the normalized squared radius, we obtain
$$P\left(\left|\frac{R^2 - d\sigma^2}{\sigma^2\sqrt{2d}}\right| \geq \lambda\right) \leq \frac{1}{\lambda^2}.$$
For any $\epsilon > 0$, let $\lambda = \epsilon\sqrt{\frac{d}{2}}$. Then we have
$$P\left(\left|\frac{R^2}{d\sigma^2} - 1\right| \geq \epsilon\right) \leq \frac{2}{\epsilon^2 d}.$$
This bound approaches zero as $d \to \infty$, which means that for large $d$:
$$\frac{R^2}{d\sigma^2} \xrightarrow{P} 1,$$
where $\xrightarrow{P}$ denotes convergence in probability. This implies:
$$R^2 \approx d\sigma^2 \implies R \approx \sigma\sqrt{d}.$$

\paragraph{Distribution of $R$.} 
We can derive a more precise result showing that not only does the norm $R$  concentrate around $\sigma\sqrt{d}$, but its deviation also follows a specific distribution.

For a chi-squared distribution with $d$ degrees of freedom, we have the following asymptotic normality result when $d$ is large:
$$\frac{\chi^2_d - d}{\sqrt{2d}} \xrightarrow{d} \mathcal{N}(0, 1).$$
This implies
$$\frac{R^2/\sigma^2 - d}{\sqrt{2d}} \xrightarrow{d} \mathcal{N}(0, 1),$$
or equivalently
$$\frac{R^2 - d\sigma^2}{\sigma^2\sqrt{2d}} \xrightarrow{d} \mathcal{N}(0, 1).$$
To obtain the distribution of $R$, we use the Delta method. Consider the function $g(x) = \sqrt{x}$ at the point $x_0 = d\sigma^2$ with first-order Taylor expansion:
$$g(x) \approx g(x_0) + g'(x_0)(x - x_0).$$
That is,
$$R = \sqrt{R^2} \approx \sigma\sqrt{d} + \frac{1}{2\sigma\sqrt{d}}(R^2 - d\sigma^2).$$
This gives us
$$R - \sigma\sqrt{d} \approx \frac{R^2 - d\sigma^2}{2\sigma\sqrt{d}}.$$
By combining with our previous result, we obtain
$$R - \sigma\sqrt{d} \xrightarrow{d} \frac{\sigma^2\sqrt{2d}}{2\sigma\sqrt{d}} \cdot \mathcal{N}(0, 1) = \frac{\sigma}{\sqrt{2}} \cdot \mathcal{N}(0, 1) = \mathcal{N}(0, \sigma^2/2).$$
Therefore, we obtain the more precise result
$$R - \sigma\sqrt{d} \xrightarrow{d} \mathcal{N}(0, \sigma^2/2) \text{ as } d \to \infty.$$

Through the ``three sigma rule of thumb'', more than $99.7\%$ of the samples drawn from the Gaussian perturbation concentrate in the ``shell'' centered at its mean with
\[
  R \in \Bigl[\sigma \sqrt{d} - \tfrac{3\sigma}{\sqrt{2}},\;
               \sigma \sqrt{d} + \tfrac{3\sigma}{\sqrt{2}}\Bigr].
\]
In particular, if we choose the noise levels $\{\sigma_i\}^N_{i=1}$ so that the outer $3\sigma$-boundary of shell $i$ coincides with the inner $3\sigma$-boundary of shell $i+1$,
\[
  \sigma_i\sqrt{d} + \frac{3\sigma_i}{\sqrt{2}}
  = \sigma_{i+1}\sqrt{d} - \frac{3\sigma_{i+1}}{\sqrt{2}},
\]
we obtain a natural ``shell-touch'' schedule that packs the space, which is used in \cref{A.4} and \cref{B.4.1} and is numerically verified in \cref{B.2.1}.

\subsection{Proposition 2 (Score Function Weight Domination)} 
\label{A.4}

\textbf{Proposition 2.} In high dimensions $d$, given a position \( x \) that is not far away from any Gaussian centers, the score function \( \nabla_{x} \log \pMN (x) \) is predominantly determined by the Gaussian component $\mathcal{N}(x; \mu_*, \sigma_*^2 \mathbf{I})$, where \( \mu_* \) and $\sigma_*$ are defined as follows:
\begin{align*}
\mu_* &= \arg\min_{\mu_j, \, j=1,2,\dotsc,M} \, \|x - \mu_j\|_2, \\
\sigma_* &= \arg\min_{\sigma_i, \, i=1,2,\dotsc,N} \, \left| \frac{\sigma_{opt}^2}{\sigma_i^2} -1 \right|.
\end{align*}
where  $\sigma_{opt} := \frac{\|x-\mu_*\|}{\sqrt{d-2}} \approx  \frac{\|x-\mu_*\|}{\sqrt{d}}$.

We divide the proof into two sub-propositions, one concerning $\sigma$ domination, and the other, $\mu$ domination.

\subsubsection{Proposition 2.1 ($\sigma$ domination)} 
\label{A.4.1}

\textbf{Proposition 2.1.} Given a certain position $x$ and any fixed center $\mu_j$, there exists an optimal noise level $\sigma_j^*$ such that the corresponding Gaussian component dominates the score function weight over all other noise levels.

Substituting the $\lambda(\sigma)=\sigma^2$, each weight $w_{i}(x)$ is a constant multiple of the ``effective weight" $w_{i}^E$ defined by 
$$w^E_{i} \triangleq \frac{1}{\sigma _i^{d-2}}\exp \left ( -\frac{\left \| x-\mu_* \right \|^2 }{2\sigma _i^2}  \right ). $$
Maximizing $w_{i}(x)$ is equivalent to minimizing $- \ln w^E_{i}$, so we identify $\sigma_*$ by solving the following discrete minimization problem:
\[
   \min_i \, -\ln w_{i}^E
 \iff  \min_{\sigma_i} \,  (d-2)\ln \sigma _i+\frac{\left \| x-\mu_* \right \|^2 }{2\sigma_i^2}.
\]
That is, $\mu_*$ is well defined, independently of $\sigma_i$.
Defining\footnote{Note that the $f$ here is the \emph{negative} of the function used inside the softmax score function weight in the main paper \cref{sec:memorize}; this choice is only for convenience.}
\begin{equation} \label{eq:def.f}
f(\sigma) := (d-2)\ln \sigma+ \frac{\|x-\mu_*\|^2}{2\sigma^2},
\end{equation}

\noindent
we have $f'(\sigma) = \tfrac{1}{\sigma } (d-2) - \frac{||x-\mu_*||^2}{\sigma^3}$. 
We have $f'(\sigopt)=0$ as former definition.
Moreover, we have $f''(\sigma) = -\frac{1}{\sigma^2} (d-2) + \frac{3}{\sigma^4} \|x-\mu_*\|^2$, so that $f''(\sigopt)=\frac{2(d-2)}{\sigopt^2}$, so indeed, $\sigopt = \frac{||x-\mu_*||}{\sqrt{d-2}}\approx \frac{||x-\mu_*||}{\sqrt{d}}$ minimizes $f$ on the interval $\sigma \in (0,\infty)$.

\paragraph{Case of continuous $\sigma$.}

For the continuous case, we can exactly choose $\sigma_*=\sigopt$, so we can just analyze the behavior of $f(\sigma)$ around the minimizer $\sigma_* = \frac{\|x-\mu_*\|}{\sqrt{d-2}}$.
Using the Taylor expansion around $\sigma_*$, for $\sigma = \sigma_* + \delta$ with small $|\delta|$, we have
\[
f(\sigma_* + \delta) = f(\sigma_*) + f'(\sigma_*)\delta + \frac{1}{2}f''(\sigma_*)\delta^2 + O(\delta^3).
\]
Since $f'(\sigma_*) = 0$ by definition and $f''(\sigma_*) = \frac{2(d-2)}{\sigma_*^2} > 0$ for $d > 2$, we have:
\[
f(\sigma_* + \delta) - f(\sigma_*) = \frac{d-2}{\sigma_*^2}\delta^2 + O(\delta^3).
\]
As $d$ is large, the coefficient $\frac{d-2}{\sigma_*^2}$ grows linearly with $d$, making the function $f(\sigma)$ increasingly peaked around $\sigma_*$. Therefore, any perturbation away from $\sigma_*$ results in a significantly larger function value, confirming that $\sigma_*$ dominates the score function weight over all noise levels.

\paragraph{Case of discrete $\sigma$.}

Since $\sigma_*$ must take on one of the values $\sigma_i$, $i=1,2,\dotsc,N$, it will generally not be equal to $\sigopt$. 
We examine next how the coefficient $w_{ij}^E$ dominates the other coefficients, when $\sigma_*=\sigma_i$ for some $i$ and $\mu_* = \mu_j$ for some $j$.
For these purposes, we consider the following change of variables:
\begin{equation}
    \label{eq:siggam}
    \sigma_i^2 = \frac{\sigopt^2}{1+\gamma_i} \; \iff \; \gamma_i = \frac{\sigopt^2}{\sigma_i^2}-1,
\end{equation}
so that $\gamma_1 >\gamma_2 > \cdots > \gamma_N$, with $\gamma_1 >0 $ when $\sigopt>\sigma_{\min}$ and $\gamma_N \in (-1,0)$ when $\sigopt<\sigma_{\max}$.
We define $\gamma_*$ in the corresponding way, that is
\[
\sigma_*^2 = \frac{\sigopt^2}{1+\gamma_*} \; \iff \; \gamma_* = \frac{\sigopt^2}{\sigma_*^2}-1.
\]
Note from $\sigma_* = \arg\min_{\sigma_1,\sigma_2,\dotsc,\sigma_N} \, \left| \frac{\sigopt^2}{\sigma_i^2} -1 \right|$ that $\gamma_*$ is the quantity $\gamma_i$ for which $|\gamma_i|$ is minimized over $i=1,2,\dotsc,N$.
We now have
\begin{align*}
    & f(\sigma_i)-f(\sigma_*) \\
    &= (f(\sigma_i)-f(\sigopt)) - (f(\sigma_*) -f(\sigopt)) \\
    &= (d-2) \ln \frac{\sigma_i}{\sigopt} + \frac{d-2}{2} \left( \frac{\sigopt^2}{\sigma_i^2} -1 \right) -
    (d-2) \ln \frac{\sigma_*}{\sigopt} - \frac{d-2}{2} \left( \frac{\sigopt^2}{\sigma_*^2} -1 \right) \\
    &= \frac{d-2}{2} (\ln (1+\gamma_*) - \ln(1+\gamma_i)) + \frac{d-2}{2} (\gamma_i - \gamma_*),
\end{align*}
so that 
\begin{equation}
    \label{eq:sh1}
    \frac{2}{d-2} (f(\sigma_i)-f(\sigma_*)) = \ln (1+\gamma_*) - \ln(1+\gamma_i) + (\gamma_i-\gamma_*) \approx \frac12 \gamma_i^2 - \frac12 \gamma_*^2,
\end{equation}
so that $f(\sigma_i)> f(\sigma_*)$ unless $\sigma_i = \sigma_*$. Note that when $\sigopt$ is near the midpoint between $\sigma_i$ and $\sigma_{i+1}$, we have $|\gamma_*| \approx \gamma_i \approx -\gamma_{i+1}$ and $f(\sigma_i) \approx f(\sigma_{i+1})$.

We use \eqref{eq:sh1} to examine what happens as $\sigopt$ moves away from the near-midpoint of the interval $[\sigma_i,\sigma_{i+1}]$, or more precisely, when the index $i$ for which $\gamma_i>0$ and $\gamma_{i+1}<0$ deviates away from $\gamma_i \approx -\gamma_{i+1}$. Note first that for any $i$, we have from (boundaries of these shells touch, as illustrated in \cref{A.3})

\begin{equation} \label{eq:sh2}
    \sigma_i \sqrt{d} + \frac{3 \sigma_i}{\sqrt{2}} = \sigma_{i+1} - \frac{3 \sigma_{i+1}}{\sqrt{2}},
\end{equation} 
that 
\[
\frac{\sigma_{i+1}^2}{\sigma_i^2} \approx 1+\frac{6 \sqrt{2}}{\sqrt{d}}.
\]
Meanwhile, from \eqref{eq:siggam}, we have
\[
\frac{\sigma_{i+1}^2}{\sigma_i^2} = \frac{1+\gamma_i}{1+\gamma_{i+1}} \approx 1+\gamma_i -\gamma_{i+1},
\]
so that
\begin{equation} \label{eq:sh3}
    \gamma_i-\gamma_{i+1} \approx \frac{6 \sqrt{2}}{\sqrt{d}}.
\end{equation}
Consider now the case in which $\gamma_* = \gamma_i>0$, that is, $\gamma_i$ is slightly smaller  than $|\gamma_{i+1}|$. Because of \eqref{eq:sh3}, we can write
\[
\gamma_i = \alpha \frac{6 \sqrt{2}}{\sqrt{d}}, \quad \gamma_{i+1} = -(1-\alpha) \frac{6 \sqrt{2}}{\sqrt{d}}, \quad \mbox{for some $\alpha \in [0,.5)$.}
\]
By subsituting into \eqref{eq:sh1}, we obtain
\[
\frac{2}{d-2} (f(\sigma_{i+1}) - f(\sigma_*)) \approx \frac{36}{d} ((1-\alpha)^2-\alpha^2) = \frac{36}{d} (1-2\alpha),
\]
so that $f(\sigma_{i+1}) \approx f(\sigma_*) + 18 (1-2\alpha)$.
Returning to the definition of $f$, $w_{ij}^E$, and $w_{ij}(x)$, we have for $j$ such that $\mu_j = \mu_*$ that
\[
\frac{w_{ij}(x)}{w_{i+1,j}(x)} \approx e^{18(1-2\alpha)}.
\]
For the case of $\alpha=\frac{1}{3}$ (that is, $\sigopt$ is only slightly closer to $\sigma_i$ than to $\sigma_{i+1}$), we have 
$\frac{w_{ij}(x)}{w_{i+1,j}(x)} \approx e^{6} \approx 403$, so $w_{ij}(x)$ dominates its closest neighbor. Its dominance over all other weights will be even greater.

\subsubsection{Proposition 2.2 ($\mu$ domination)} 
\label{A.4.2}

\textbf{Proposition 2.2.} Given a position $x$, the score function weight of a center decreases exponentially as the distance between the center and $x$ increases.

We recall the definition of the score function:
\[
\nabla_{x} \log \pMN(x) = -\sum_{j=1}^{M} \sum_{i=1}^{N} w_{ij}(x) \frac{x - \mu_j}{\sigma_i^2},
\]
where the weights $w_{ij}(x)$  are
\[
w_{ij}(x) = \frac{\lambda(\sigma_i) \mathcal{N}(x; \mu_j, \sigma_i^2 \mathbf{I})}{\sum_{l=1}^{M} \sum_{k=1}^{N} \lambda(\sigma_k) \mathcal{N}(x; \mu_l, \sigma_k^2 \mathbf{I})}.
\]

We have proved that for a given center $\mu_j$, the term involving $\sigma_j^*$ will have the dominant weight. 
The dominant weight corresponding to $\mu_j$ will thus be (approximately)
\[
w_{j}^*(x) = \frac{\lambda(\sigma_j^*) \mathcal{N}(x; \mu_j, \sigma_j^{*2} \mathbf{I})}{\sum_{l=1}^{M}  \lambda(\sigma_l^*) \mathcal{N}(x; \mu_l, \sigma_l^{*2} \mathbf{I})}.
\]
We can compare the contributions of different centers $\mu_j$ by comparing these quantities $w_j^*$. 
For example, the relative contributions of centers $\mu_j$ and $\mu_l$ are captured by the ratio
\[
 \ln \frac{w_j^*}{w_l^*} = \ln\frac{ \lambda(\sigma_j^*) \mathcal{N}(x; \mu_j, \sigma_j^{*2} \mathbf{I})  }{ \lambda(\sigma_l^*) \mathcal{N}(x; \mu_l, \sigma_l^{*2} \mathbf{I})} = \underbrace{(d-2) \ln \frac{\sigma_l^*}{\sigma_j^*}}_{\text{term1}} + \underbrace{\frac{||x-\mu_l||^2}{2\sigma_l^{*2}} - \frac{||x-\mu_j||^2}{2\sigma_j^{*2}}}_{\text{term2}}
\]

We assume $\mu_j$ is the closest center to $x$, so its corresponding $\sigma_j^*$ is also the smallest, which means that term1$\ge 0$. Moreover, we know $x$ should be in both shell $(\mu_j,\sigma_j^*)$ and shell $(\mu_l,\sigma_l^*)$, so obviously in the \textbf{case of continuous $\sigma$},
 term2 will be equal to 0.
 We thus have $\ln \frac{w_j^*}{w_l^*} = (d-2)\ln\frac{\sigma_l^*}{\sigma_j^*}$. 
 Similar to the analysis in $\sigma$ domination, this function can be regarded as a delta function when $d$ is large, so the closest center $\mu_j=\mu_*$ will dominate.  

\paragraph{Case of discrete $\sigma$.}

term2 can be bounded when $x$ is exactly on the boundary of the shells. 
Therefore, if we want to minimize the ratio to see how much the contribution of $\mu_j$ would be bigger that the $\mu_l$ at least, we could only consider the case that $\sigma_l^*=\sigma_j^*$ or $\sigma_l^*$ is the neighbor of the $\sigma_j^*$.

\noindent
We can assume $\frac{\sigma_j^*}{\sigma_l^*}= (\frac{\sigma_{i+1}}{\sigma_i} )^k$, which means $\sigma_l^*$ is the $k$-th bigger noise level of $\sigma_j^*$, and then we have the ratio
\begin{align} 
 \notag \ln \frac{w_j^*}{w_l^*} &=  {(d-2) \ln \frac{\sigma_l^*}{\sigma_j^*}} + {\frac{||x-\mu_l||^2}{2\sigma_l^{*2}} - \frac{||x-\mu_j||^2}{2\sigma_j^{*2}}}\\ \notag
    &= \frac{k(d-2)}{2} \ln (1+\frac{6\sqrt{2} }{\sqrt{d} } )  + {\frac{||x-\mu_l||^2}{2\sigma_l^{*2}} - \frac{||x-\mu_j||^2}{2\sigma_j^{*2}}}\\
    \notag
\end{align}
Since we know the $x$ is in the shells of component $(\mu_j,\sigma_j^*)$ and $(\mu_l,\sigma_l^*)$, the last two term should be bounded in the interval $[\frac{1}{2}(\sqrt{d}-\frac{3}{\sqrt{2}})^2,  \frac{1}{2}(\sqrt{d} + \frac{3}{\sqrt{2}})^2]$. Therefore, we can  lower-bound  the ratio as follows:
\begin{align} 
    \notag \ln \frac{w_j^*}{w_l^*} &= \frac{k(d-2)}{2} \ln (1+\frac{6\sqrt{2} }{\sqrt{d} } )  + {\frac{||x-\mu_l||^2}{2\sigma_l^{*2}} - \frac{||x-\mu_j||^2}{2\sigma_j^{*2}}}\\
    & \ge \frac{k(d-2)}{2} \ln (1+\frac{6\sqrt{2} }{\sqrt{d} } ) - 3\sqrt{2d} \notag \\
    & \approx (k-1)3\sqrt{2d} \notag
\end{align}
Obviously, the ratio increases significantly as the $k$ increases. And for this case (worst case for the sub of the squre terms), $k=1$ also means $||x-\mu_j||=||x-\mu_l||$,  because
\[
\frac{||x-\mu_j||^2}{2\sigma_j^{*2}}=\frac{1}{2}(\sqrt{d}+\frac{3}{\sqrt{2}})^2 \qquad \text{and} \qquad \frac{||x-\mu_l||^2}{2\sigma_l^{*2}}=\frac{1}{2}(\sqrt{d}-\frac{3}{\sqrt{2}})^2.
\]
After introducing $\frac{\sigma_l^{*2}}{\sigma_j^{*2}} = (1+\frac{3\sqrt{2}}{\sqrt{d}})^2$, we can easily get the conclusion that $||x-\mu_j||=||x-\mu_l||$.

\begin{figure}[!htb]
    \centering
    \includegraphics[width=0.45\linewidth]{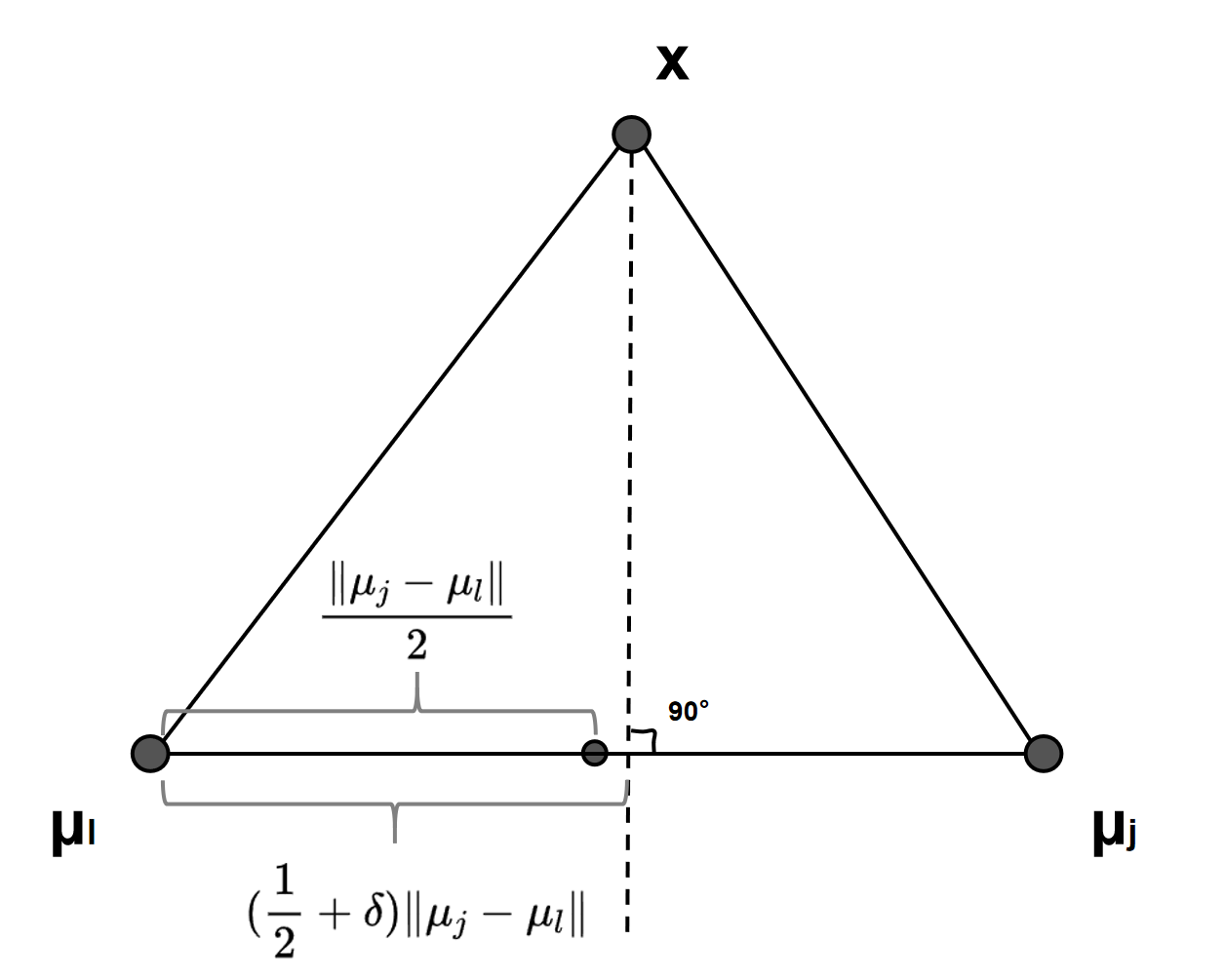}
    \caption{Illustration for our computation of the score function weight ratio between $\mu_j$ and $\mu_l$.}
    \label{fig:placeholder}
\end{figure}

Now we consider the case in which $\|x-\mu_j\|$ is only slightly smaller than $\|x-\mu_l\|$.
To analyze it more specifically, we can just assume $\sigma_l^*=\sigma_j^*$, and we get
\[
\ln \frac{w_j^*}{w_l^*} = \frac{||x-\mu_l||^2-||x-\mu_j||^2}{2\sigma_j^{*2}}
\]
Considering the triangle $\triangle x\mu_j\mu_l$ in \cref{fig:placeholder},

\noindent
we can get the following equations through the cosine theorem:
\[
\cos \angle x\mu_l\mu_j = \frac{||x-\mu_l||^2 + ||\mu_l-\mu_j||^2-||x-\mu_j||^2}{2||\mu_l-\mu_j||||x-\mu_l||} .
\]
So
\[
||x-\mu_l||^2 - ||x-\mu_j||^2 = 2||\mu_l-\mu_j||||x-\mu_l||\cos \angle x\mu_l\mu_j - ||\mu_l-\mu_j||^2.
\]
Since $||x-\mu_j|| \le ||x-\mu_l||$, we can let $||x-\mu_l||\cos\angle x\mu_l\mu_j = (\frac{1}{2}+\delta)||\mu_j-\mu_l||$, where $\delta \ge 0$ and indicates how much the $||x-\mu_l||$ is bigger than the $||x-\mu_j||$. Besides, to ensures $\sigma_j^*=\sigma_l^*$, obviously from the shell's radius ($R=\sigma(\sqrt{d}\pm \frac{3}{\sqrt{2}})$) and the properties of similar triangles, we should have
$$\frac{\delta||\mu_j-\mu_l||}{\frac{1}{2}||\mu_j-\mu_l||} \lesssim \frac{\frac{3}{\sqrt{2}}}{\sqrt{d}} \Rightarrow  \delta \lesssim O(\frac{1}{\sqrt{d}})$$

After substituting the above equations into the weight ratio, we get
\[
\ln \frac{w_j^*}{w_l^*} = \frac{||x-\mu_l||^2 - ||x-\mu_j||^2}{2\sigma_j^{*2}} = \frac{\delta||\mu_j-\mu_l||^2}{\sigma_j^{*2}}.
\]
Therefore,
\[
\frac{w_j^*}{w_l^*} = \exp \left( \frac{\delta||\mu_j-\mu_l||^2}{\sigma_j^{*2}} \right).
\]
When $\sigma^* \to 0$, this ratio grows exponentially, explaining why memorization occurs when the noise level is small.

Finally, we \textbf{emphasize} that the “shell-touch’’ noise schedule is used only as a convenient discretization device. Our core dominance results for both $\sigma$ and $\mu$ follow from the continuous analysis of $f(\sigma)$, and therefore do not depend on any particular discrete noise schedule. In other words, $\sigma$- and $\mu$-domination already hold in the continuous-$\sigma$ setting for general noise schedules; the shell-touch construction simply picks one specific discretization so that (i) every sampling position $x$ lies in at least one shell and (ii) adjacent shells have analytically tractable spacing. The only part that truly exploits this particular schedule is the specific closed-form expression for the discrete weight ratio in Proposition 2.1, which is used to quantify and illustrate how strong the domination can be.

\subsection{Sampling Expansiveness Ratio $\gamma_{ex}$}
\label{A.5}

Under the geometric setup in \cref{sharp_overlap} of the main paper, the two closest training points are denoted by $\mu_0$ and $\mu_1$, and we choose $x$ and $y$ such that
\[
\|x-\mu_0\| = C\|x-\mu_1\|,\qquad
\|y-\mu_1\| = C\|y-\mu_0\|,\qquad C \ge 1,
\]
with the same shell radius $\sigma^*\sqrt{d}$ for both centers, and $y$ is obtained from $x$ by a small perturbation along the direction $(\mu_0-\mu_1)$. Under this symmetric configuration, we can draw a figure as the following:

\begin{figure}[!htb]
    \centering
    \includegraphics[width=0.35\linewidth]{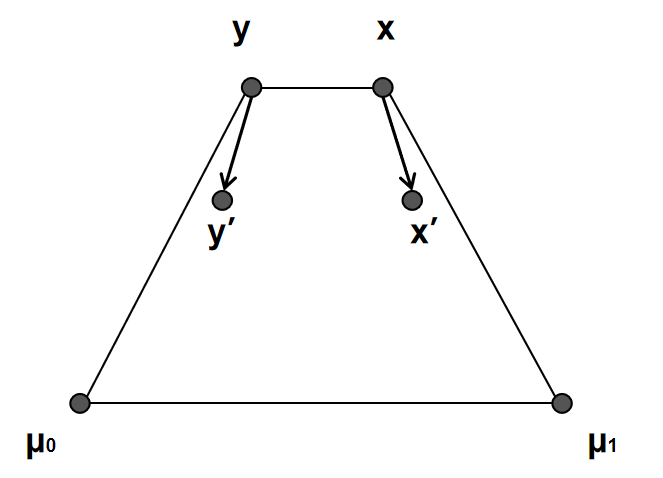}
    \caption{Illustration for our computation of the sampling expansiveness ratio $\gamma_{ex}$.}
    \label{fig_trapezoids}
\end{figure}

\noindent
where both of the $xy\mu_0\mu_1$ and $xyy'x'$ are isosceles trapezoids. 
Since $x$ and $y$ are symmetric in this sense, the score function weight ratios (dominant to subdominant score function weight) are the same for positions $x$ and $y$, which is denoted by $a$. Therefore, the conditional score function at $x$ and $y$ can be represented as:
\[
s(x, \sigma^*)= \frac{1}{a+1}\Bigl(-\frac{x-\mu_0}{\sigma^{*2}}\Bigr)+\frac{a}{a+1}\Bigl(-\frac{x-\mu_1}{\sigma^{*2}}\Bigr), \qquad
s(y, \sigma^*)= \frac{a}{a+1}\Bigl(-\frac{y-\mu_0}{\sigma^{*2}}\Bigr)+\frac{1}{a+1}\Bigl(-\frac{y-\mu_1}{\sigma^{*2}}\Bigr).
\]
After one-step ODE sampling, we have:
\[
x'=x + \eta \sigma^{*2}s(x, \sigma^*), \qquad y'=y + \eta \sigma^{*2}s(y, \sigma^*),
\]
so
\[
\|y'-x'\| = \bigl\|y-x+\eta\sigma^{*2}\bigl(s(y, \sigma^*)-s(x, \sigma^*)\bigr)\bigr\|
= \bigl\|(y-x)(1-\eta) + \eta\frac{a-1}{a+1}(\mu_0-\mu_1)\bigr\|,
\]
where the last equality follows from a direct algebraic expansion under the above symmetry assumptions. Therefore, the expansion factor $\gamma_{ex}$ can be represented as:
\[
 \gamma_{ex} = \frac{\bigl\|(y-x)(1-\eta) + \eta\frac{a-1}{a+1}(\mu_0-\mu_1)\bigr\|}{\|y-x\|}.
\]
Since $x$ is close to $y$, which means $\|y-x\|\ll\|\mu_0-\mu_1\|$, and $(y-x)$ is a small perturbation along $(\mu_0-\mu_1)$, the term proportional to $(\mu_0-\mu_1)$ dominates. Thus this ratio can be further approximated as
\[
 \gamma_{ex} \approx \frac{\bigl\|\eta\frac{a-1}{a+1}(\mu_0-\mu_1)\bigr\|}{\|y-x\|}
 = \frac{\| \eta (\mu_0-\mu_1)\|}{\|y-x\|} \left | 1-\frac{2}{a+1} \right |.
\]

Note that if the noise levels are sparse (that is, for a fixed center, $x$ and $y$ have the same optimal noise level), the conditioning and unconditioning modeling will have the same expansion ratio. If the noise level is more continuous, we have:
\[
s(x)= \frac{1}{a+1}\Bigl(-\frac{x-\mu_0}{C^2\sigma^{*2}}\Bigr)+\frac{a}{a+1}\Bigl(-\frac{x-\mu_1}{\sigma^{*2}}\Bigr), \qquad
s(y)= \frac{a}{a+1}\Bigl(-\frac{y-\mu_0}{\sigma^{*2}}\Bigr)+\frac{1}{a+1}\Bigl(-\frac{y-\mu_1}{C^2\sigma^{*2}}\Bigr),
\]
\[
x'=x + \eta \sigma^{*2}s(x), \qquad y'=y + \eta \sigma^{*2}s(y),
\]
where $C=\frac{\|x-\mu_0\|}{\|x-\mu_1\|}=\frac{\|y-\mu_1\|}{\|y-\mu_0\|} \ge 1$ encodes that the farther center uses a proportionally larger optimal noise level (smaller precision) under the optimal-shell approximation. 
The ratio is
\[
\gamma_{ex} \approx \frac{\bigl\|\eta\frac{a-\frac{1}{C^2}}{a+1}(\mu_0-\mu_1)\bigr\|}{\|y-x\|}
 \approx \frac{\| \eta (\mu_0-\mu_1)\|}{\|y-x\|} \left | 1-\frac{2}{a+1} \right |,
\]
which is because, as shown in Appendix~A.4.2, $a\approx C^{(d-2)}\gg C$. 
We know the unconditioning case has a smaller $a$, leading to smaller $\gamma_{ex}$. For temperature-based smoothing, increasing $T$ similarly reduces the effective weight contrast $a$ between centers, and thus also decreases $\gamma_{ex}$ in the same way.

\subsection*{General View of the Expansiveness: Hessian Analysis of the Unified Log-Density}
\addcontentsline{toc}{subsection}{General View of the Expansiveness: Hessian Analysis of the Unified Log-Density}

We now move from the symmetric two-center configuration of Appendix~\ref{A.5} to the general case. Our goal is to
characterize the one-step sampling expansiveness ratio $\gamma_{ex}$ via the Hessian of the unified, temperature-smoothed
log-density under the optimal-shell approximation, and to understand how the spectrum of this Hessian depends on the
temperature $T$ and on conditioning versus unconditioning. The construction in Appendix~\ref{A.5} can then be recovered
as a special case of this Hessian-based analysis in a particular two-center configuration.

Recall from Section~2.2 that under noise unconditioning the sampling dynamics can be viewed as gradient ascent on a
fixed objective:
\[
  \frac{dx}{dt} = \frac{\eta}{2} \,\nabla_x \log \pMN(x),
\]
and in the temperature-smoothed setting we work with the temperature-modified log-density $\log \pMN(x,T)$, whose
gradient is the temperature-smoothed score $s(x;T)$.
By definition,
\[
  s(x;T) \;\coloneqq\; \nabla_x \log \pMN(x,T).
\]
Therefore, the Jacobian of the score coincides with the Hessian of the unified log-density:
\[
  J(x;T)
  \;\coloneqq\;
  \nabla_x s(x;T)
  \;=\;
  \nabla^2_{x,x} \log \pMN(x,T)
  \;\eqqcolon\; H(x;T).
\]
In what follows we denote this matrix by $J(x;T)$ (to be consistent with the rest of the paper), but we interpret it
throughout as the Hessian of $\log \pMN(x,T)$.

\paragraph{Hessian decomposition.}
Under the optimal-shell approximation, the empirical score of $\pMN$ with a uniform temperature $T$ can be written as
\begin{equation*}
    s(x; T)
    \;\approx\;
    \sum_{j=1}^M w_j^*(x;T)\,a_j(x),
    \qquad
    a_j(x) = -\,\frac{x-\mu_j}{\sigma_j^{*2}}
    =
    \nabla_x \log \mathcal{N}(x;\mu_j,\sigma_j^{*2}\mathbf{I}).
\end{equation*}
The Hessian of the unified log-density is
\begin{equation*}
    H(x; T)
    \;\coloneqq\;
    \nabla^2_{x,x} \log \pMN(x,T)
    \;=\;
    \nabla_x s(x; T)
    \;\in\; \mathbb{R}^{d\times d}.
\end{equation*}
We derive a closed form for $H(x;T)$ when
\[
    f_j(x):= -(d-2)\ln \sigma_j^* - \frac{\|x-\mu_j\|^2}{2\sigma_j^{*2}}
\]
is the Gaussian log-likelihood and all optimal temperatures are equal, $T_j^* \equiv T$.
By definition,
\begin{equation*}
    \nabla_x f_j(x)
    \;=\;
    -\frac{x-\mu_j}{\sigma_j^{*2}}
    \;=\;
    a_j(x),
    \qquad
    \nabla_x^2 f_j(x)
    \;=\;
    -\frac{1}{\sigma_j^{*2}} \mathbf{I}.
\end{equation*}
The temperature-smoothed weights are
\begin{equation*}
    w_j^*(x;T)
    =
    \frac{\exp(f_j(x)/T)}{\sum_{k=1}^M \exp(f_k(x)/T)}
    \in (0,1), \qquad \sum_j w_j^*(x;T) = 1.
\end{equation*}
Differentiating $w_j^*(x;T)$ with respect to $x$ gives the standard softmax gradient:
\begin{align*}
    \nabla_x w_j^*(x;T)
    &=
    \frac{1}{T}\,w_j^*(x;T)
    \Bigl(
        \nabla_x f_j(x)
        - \sum_{k=1}^M w_k^*(x;T)\,\nabla_x f_k(x)
    \Bigr) \\
    &=
    \frac{1}{T}\,w_j^*(x;T)
    \Bigl(
        a_j(x)
        - \sum_{k=1}^M w_k^*(x;T)\,a_k(x)
    \Bigr).
\end{align*}
Using $s(x; T)=\sum_j w_j^*(x;T)a_j(x)$ and the product rule,
\begin{align*}
    H(x; T)
    &=
    \nabla^2_{x,x} \log \pMN(x,T)
    =
    \nabla_x s(x; T) \\
    &=
    \sum_{j=1}^M
    \Bigl[
        \nabla_x w_j^*(x;T)\,a_j(x)^\top
        + w_j^*(x;T)\,\nabla_x a_j(x)
    \Bigr].
\end{align*}
Since $a_j(x)$ is affine in $x$,
\begin{equation*}
    \nabla_x a_j(x)
    \;=\;
    -\frac{1}{\sigma_j^{*2}} \mathbf{I}.
\end{equation*}
Substituting $\nabla_x w_j^*$ and $\nabla_x a_j$ yields
\begin{align*}
    H(x; T)
    &=
    \sum_{j=1}^M
    \left[
        \frac{1}{T}\,w_j^*(x;T)
        \Bigl(
            a_j(x)
            - \sum_{k=1}^M w_k^*(x;T)\,a_k(x)
        \Bigr)
        a_j(x)^\top
        - w_j^*(x;T)\,\frac{1}{\sigma_j^{*2}} \mathbf{I}
    \right] \\
    &=
    \frac{1}{T}\sum_{j=1}^M w_j^*(x;T)
    \Bigl(
        a_j(x)a_j(x)^\top
        - a_j(x) \Bigl[\sum_{k=1}^M w_k^*(x;T)\,a_k(x)\Bigr]^\top
    \Bigr)
    \;-\;
    \sum_{j=1}^M \frac{w_j^*(x;T)}{\sigma_j^{*2}} \mathbf{I}.
\end{align*}
Define the (temperature-smoothed) mean component score
\begin{equation*}
    \bar{a}(x;T)
    \;\coloneqq\;
    \sum_{k=1}^M w_k^*(x;T)\,a_k(x),
\end{equation*}
so that $\sum_j w_j^*(x;T)a_j(x) = \bar a(x;T)$. The first term in $H(x; T)$ can be rewritten as
\begin{align*}
    &\frac{1}{T}
    \left[
        \sum_{j=1}^M w_j^*(x;T)\,a_j(x)a_j(x)^\top
        - \sum_{j=1}^M w_j^*(x;T)\,a_j(x)\bar{a}(x;T)^\top
    \right] \\
    &\qquad=
    \frac{1}{T}
    \left[
        \sum_{j=1}^M w_j^*(x;T)\,a_j(x)a_j(x)^\top
        - \bar a(x;T)\bar a(x;T)^\top
    \right] \\
    &\qquad=
    \frac{1}{T}\,\mathrm{Cov}_{w^*(x;T)}\bigl(a_j(x)\bigr),
\end{align*}
where the weighted covariance is
\begin{equation*}
    \mathrm{Cov}_{w^*(x;T)}(a_j)
    \;\coloneqq\;
    \sum_{j=1}^M w_j^*(x;T)\,\bigl(a_j(x)-\bar a(x;T)\bigr)\bigl(a_j(x)-\bar a(x;T)\bigr)^\top.
\end{equation*}
Writing
\begin{equation*}
    c_T(x)
    \;\coloneqq\;
    \sum_{j=1}^M \frac{w_j^*(x;T)}{\sigma_j^{*2}},
\end{equation*}
we obtain the compact form
\begin{equation}
    H(x; T)
    \;=\;
    \frac{1}{T}\,
    \mathrm{Cov}_{w^*(x;T)}\bigl(a_j(x)\bigr)
    \;-\;
    c_T(x)\,\mathbf{I}.
\end{equation}
Since covariance matrices are symmetric and positive semidefinite, $H(x; T)$ is symmetric: its anisotropic part is given by
the covariance term and its isotropic part by the scalar $-c_T(x)$.

Let the eigenvalues of $\mathrm{Cov}_{w^*(x;T)}(a_j)$ be $0 \le \lambda_1(x;T) \le \dots \le \lambda_d(x;T)$. Then the
eigenvalues of the Hessian $H(x;T)$ are
\begin{equation*}
    \mu_k(x;T)
    \;=\;
    \frac{1}{T}\,\lambda_k(x;T)
    \;-\;
    c_T(x),
    \qquad k=1,\dots,d,
\end{equation*}
and in particular
\begin{align}
    \mu_{\max}(x;T)
    &\;\coloneqq\;
    \max_{k} \mu_k(x;T)
    \;=\;
    \frac{1}{T}\,\lambda_{\max}(x;T)
    \;-\;
    c_T(x),
    \label{eq:mu-max-def-hess}
    \\
    \mu_{\min}(x;T)
    &\;\coloneqq\;
    \min_{k} \mu_k(x;T)
    \;=\;
    \frac{1}{T}\,\lambda_{\min}(x;T)
    \;-\;
    c_T(x)
    \;\ge\;
    -\,c_T(x),
\end{align}
because $\lambda_{\min}(x;T)\ge 0$ and $c_T(x)>0$. In the regime of interest, $\mu_{\min}(x;T)$ is negative. Consequently,
\begin{equation}
    |\mu_{\min}(x;T)|
    = -\mu_{\min}(x;T)
    \;\le\; c_T(x).
    \label{eq:mu-min-upper-hess}
\end{equation}

\paragraph{From Hessian eigenvalues to $\gamma_{ex}$.}
Consider one-step ODE sampling with step size $\eta$:
\begin{equation*}
    x' = x + \eta\,s(x; T)
    \;=\; x + \eta\,\nabla_x \log \pMN(x,T).
\end{equation*}
For two nearby points $x,y$, a first-order Taylor expansion gives
\begin{equation*}
    y' - x'
    \;\approx\;
    \bigl(\mathbf{I} + \eta\,H(x; T)\bigr)\,(y-x).
\end{equation*}
Therefore the one-step expansiveness ratio
\begin{equation*}
    \gamma_{ex}(x,y)
    \;\coloneqq\;
    \frac{\|y' - x'\|}{\|y-x\|}
\end{equation*}
admits the linearized bound
\begin{equation*}
    \gamma_{ex}(x,y)
    \;\lesssim\;
    \max_{\|v\|=1}
    \bigl\|
        (\mathbf{I} + \eta\,H(x; T))v
    \bigr\|
    \;=\;
    \max_k \bigl|1 + \eta\,\mu_k(x;T)\bigr|.
\end{equation*}

The symmetric two-center example in Appendix~\ref{A.5} can be viewed as an explicit computation of $\gamma_{ex}$ in the
2D subspace spanned by the center difference $(\mu_0-\mu_1)$ and the displacement $(y-x)$. There, the factor
\begin{equation*}
    \left|1 - \frac{2}{a+1}\right|
\end{equation*}
(where $a$ is the dominant-to-subdominant weight ratio) plays the role of an effective Hessian eigenvalue in that subspace:
a larger $a$ leads to a larger expansion factor $\gamma_{ex}$, and reducing $a$ (via unconditioning or temperature
smoothing) reduces this effective eigenvalue and hence $\gamma_{ex}$, exactly as shown in the trapezoid calculation.

\paragraph{Effect of unconditioning.}
The Hessian decomposition
\begin{equation*}
    H(x; T)
    \;=\;
    \frac{1}{T}\,\mathrm{Cov}_{w^*(x;T)}(a_j)
    \;-\;
    c_T(x)\,\mathbf{I}
\end{equation*}
also clarifies the effect of unconditioning. Removing conditioning changes how the optimal noise levels $\{\sigma_j^*\}$
and weights $\{w_j^*(x;T)\}$ are assigned across centers. In particular, it reduces extreme dominance of a single center
(smaller weight ratio $a$ in the toy example), spreads the weights over more nearby centers, and thereby reduces the largest
covariance eigenvalue $\lambda_{\max}(x;T)$ along directions corresponding to large center separations $(\mu_j-\mu_{j'})$.
Under the score function weight domination, if the fixed noise level of conditioning is equal to the smallest optimal noise
level of unconditioning, we can reasonably assume the $c_T(x)$ will not change a lot. Then through
\eqref{eq:mu-max-def-hess}, a smaller $\lambda_{\max}(x;T)$ again leads to a smaller $\mu_{\max}(x;T)$ and thus a smaller
local expansion factor $\gamma_{ex}(x,y)$.

\paragraph{Monotonicity of $c_T(x)$ in $T$.}
We now show that the upper bound in~\eqref{eq:mu-min-upper-hess}, $|\mu_{\min}(x;T)|\le c_T(x)$, becomes tighter as $T$
increases, in the sense that $c_T(x)$ is non-increasing in $T$.

Differentiating the softmax weights yields
\[
    \frac{\partial w_j^*(x;T)}{\partial T}
    =
    \frac{w_j^*(x;T)}{T^2}
    \Bigl(
        \mathbb{E}_{w^*(x;T)}[f(\cdot)]
        - f_j(x)
    \Bigr),
\]
and hence
\begin{align*}
    \frac{\partial c_T(x)}{\partial T}
    &=
    \sum_{j=1}^M \frac{1}{\sigma_j^{*2}}
    \frac{\partial w_j^*(x;T)}{\partial T} \\
    &=
    \frac{1}{T^2}
    \Bigl(
        \mathbb{E}_{w^*(x;T)}[f(\cdot)]\,\mathbb{E}_{w^*(x;T)}[1/\sigma^{*2}]
        - \mathbb{E}_{w^*(x;T)}[f(\cdot)\,1/\sigma^{*2}]
    \Bigr) \\
    &=
    -\frac{1}{T^2}\,
    \mathrm{Cov}_{w^*(x;T)}\bigl(f_j(x),1/\sigma_j^{*2}\bigr).
\end{align*}
By Proposition~2, components with larger log-likelihood $f_j(x)$ tend to have smaller $\sigma_j^*$ and hence larger
precision $1/\sigma_j^{*2}$. Thus, under the posterior weights $w_j^*(x;T)$, $f_j(x)$ and $1/\sigma_j^{*2}$ are positively
correlated, implying
\[
    \mathrm{Cov}_{w^*(x;T)}\bigl(f_j(x),1/\sigma_j^{*2}\bigr) \;\ge\; 0
\]
in this regime. Combining with the derivative above gives
\begin{equation*}
    \frac{\partial c_T(x)}{\partial T}
    \;=\;
    -\frac{1}{T^2}\,
    \mathrm{Cov}_{w^*(x;T)}\bigl(f_j(x),1/\sigma_j^{*2}\bigr)
    \;\le\; 0,
\end{equation*}
with strict inequality whenever the covariance is strictly positive. Intuitively, starting from small $T$ where the weight
concentrates on the closest, smallest-noise component, increasing $T$ redistributes mass toward components with larger
$\sigma_j^*$, thereby decreasing the weighted average $\sum_j w_j^*(x;T)/\sigma_j^{*2}$. The limit as $T \to \infty$ is as follows:
\[
c_{\infty}(x) = \frac{1}{M} \sum_{j=1}^M \frac{1}{{\sigma_j^*}^2},
\]
which is strictly positive in general.

Therefore, the simple bound
\begin{equation*}
    |\mu_{\min}(x;T)|
    \;\le\;
    c_T(x)
\end{equation*}
is itself a non-increasing function of $T$: the magnitude of the most negative eigenvalue of $H(x;T)$ cannot grow faster
than $c_T(x)$ and typically shrinks as $T$ increases.

\paragraph{An upper bound on $\mu_{\max}(x;T)$ that decreases with $T$.}
To obtain an explicit dependence on $T$ for the positive side of the spectrum, we use the simple bound
\begin{equation*}
    \lambda_{\max}(x;T)
    \;\le\;
    \Lambda(x)
    \;\coloneqq\;
    \max_{j} \, \|a_j(x)\|^2,
\end{equation*}
because $\mathrm{Cov}_{w^*(x;T)}(a_j)$ is a convex combination of outer products $a_j(x)a_j(x)^\top$. Hence
\begin{equation}
    \mu_{\max}(x;T)
    \;\le\;
    \frac{\Lambda(x)}{T} - c_T(x).
    \label{eq:mu-max-upper-bound-hess}
\end{equation}
The right-hand side is now explicit in $T$. Differentiating it and using $\partial_T c_T(x) \le 0$ yields
\begin{equation*}
    \frac{\partial}{\partial T}
    \Bigl(
        \frac{\Lambda(x)}{T} - c_T(x)
    \Bigr)
    =
    -\frac{\Lambda(x)}{T^2}
    \;-\;
    \frac{\partial c_T(x)}{\partial T}
    =
    -\frac{\Lambda(x)}{T^2}
    +
    \frac{1}{T^2}\,
    \mathrm{Cov}_{w^*(x;T)}\bigl(f_j(x),1/\sigma_j^{*2}\bigr).
\end{equation*}
In our study, we focus on the top-$k$ centers (used for temperature smoothing), for which the optimal noise levels do not
differ by orders of magnitude. Hence the covariance satisfies
\begin{equation*}
    \mathrm{Cov}_{w^*(x;T)}\bigl(f_j(x),1/\sigma_j^{*2}\bigr)
    \;\le\;
    \frac{1}{4}
    \Bigl(
        \max_j f_j(x) - \min_j f_j(x)
    \Bigr)
    \Biggl(
        \max_j \frac{1}{\sigma_j^{*2}}
        - \min_j \frac{1}{\sigma_j^{*2}}
    \Biggr)
    \;<\;
    \frac{1}{4}\,\frac{d}{\min_j \sigma_j^{*2}}\,
    \ln\frac{\max_j \sigma_j^*}{\min_j \sigma_j^*},
\end{equation*}
where the first inequality is Gr\"uss's inequality and the second inequality follows from Appendix~\ref{A.4.2}. This upper
bound is typically smaller than $\Lambda(x)\approx \frac{d}{\min_j \sigma_j^{*2}}$ (since
$\frac{1}{4}\ln\frac{\max_j \sigma_j^*}{\min_j \sigma_j^*}<1$), so the derivative of the upper bound
in~\eqref{eq:mu-max-upper-bound-hess} is negative in the regime of interest:
\begin{equation*}
    \frac{\partial}{\partial T}
    \Bigl(
        \frac{\Lambda(x)}{T} - c_T(x)
    \Bigr)
    \;<\; 0.
\end{equation*}
Thus, $\mu_{\max}(x;T)$ is controlled by a quantity that shrinks as $T$ increases.

\paragraph{Effect of temperature on the Hessian spectrum and expansiveness.}
Combining the bounds on both sides of the spectrum, we have
\begin{equation*}
    \max_k |\mu_k(x;T)|
    \;\le\;
    \max\Bigl\{
        |\mu_{\max}(x;T)|,\;|\mu_{\min}(x;T)|
    \Bigr\}
    \;\le\;
    \max\Bigl\{
        \tfrac{\Lambda(x)}{T} - c_T(x),\;c_T(x)
    \Bigr\},
\end{equation*}
where the final bound uses the fact that $\Lambda(x) \ge 0$.
As $T$ increases, the upper bound on $\mu_{\max}(x;T)$ decreases to zero (by~\eqref{eq:mu-max-upper-bound-hess} and its
derivative), and $c_T(x)$ is non-increasing in $T$, with a positive lower bound $c_{\infty}(x)$. Therefore all eigenvalues of the Hessian $H(x;T)$ approach $-c_{\infty}(x)$ as $T \to \infty$.

In summary, the Hessian perspective provides an explicit spectral characterization of the sampling expansiveness ratio: in the small-step limit $\eta\to0$, the worst-case local $\gamma_{ex}$ is governed by the $\rho(x;T) := \max_{k=1,2,\dots,d} |1+\eta \mu_k(x;T)|$,
\[
    H(x;T) = \nabla^2_{x,x} \log \pMN(x,T),
    \qquad
    \mu_k(x;T) = \frac{1}{T}\lambda_k(x;T) - c_T(x).
\]
As $T\to \infty$, we have $\rho(x;T) \to (1- \eta c_{\infty}(x))$ provided that $\eta < c_{\infty}(x)^{-1}$. Note that for $\mu_{\max} (x;T)> 0$, then $\rho(x;T)>1$. Therefore, to avoid the memorization while maintaining the sampling quality, we hope the $\rho(x;T)$ is bigger than 1 by some modest amount, that is, $\rho(x;T)= \max_{k=1,2,\dots,d} |1+\eta \mu_k(x;T)| = 1+\tau$, where $\tau>0$. 
We need $T$ to be small enough that $\mu_{\max}(x;T)>0$; then the latter requirement reduces to $\eta\mu_{\max}(x;T)=\tau$.
This formula is helpful in designing the sampling stepsize and temperature $T$ for a given $\tau$.  

Both unconditioning and temperature smoothing act primarily by reducing the dominant covariance eigenvalue $\lambda_{\max}(x;T)$ and by reallocating weight away from the closest component/center, while $c_T(x)$ is non-increasing in $T$. As a result, the $\rho(x;T)$ shrinks with temperature, the log-density landscape becomes locally less curved and more Lipschitz-smooth, and the induced sampling dynamics are less expansive and thus generalize better. We numerically verify our analysis in Appendix~\ref{B.2.4}.

\subsection{Equivalent Optimization between Score Matchings}
\label{A.6}
(Here we let $\lambda(\sigma_i)=1$ to simplify writing, which can be easily extended to other cases.)

The explicit score matching loss is
\begin{equation*}
\mathcal{L}_{ESM}(\theta) = \mathbb{E}_{\pMN(x)} \left[ \frac{1}{2} \left\| s_\theta(x) - \frac{\partial \log \pMN(x)}{\partial x} \right\|^2 \right],
\end{equation*}
which we can develop as:
\begin{equation*}
\mathcal{L}_{ESM}(\theta) = \mathbb{E}_{\pMN(x)} \left[ \frac{1}{2} \|s_\theta(x)\|^2 \right] - \mathcal{T}(\theta) + C_1,
\end{equation*}
where $C_1 = \mathbb{E}_{\pMN(x)} \left[ \frac{1}{2} \left\| \frac{\partial \log \pMN(x)}{\partial x} \right\|^2 \right]$ is a constant that does not depend on $\theta$, and
\begin{align*}
\mathcal{T}(\theta) &= \mathbb{E}_{\pMN(x)} \left[ \left\langle s_\theta(x), \frac{\partial \log \pMN(x)}{\partial x} \right\rangle \right] \\
&= \int_x \pMN(x) \left\langle s_\theta(x), \frac{\partial \log \pMN(x)}{\partial x} \right\rangle dx \\
&= \int_x \left\langle s_\theta(x), \frac{\partial \pMN(x)}{\partial x} \right\rangle dx.
\end{align*}
Now, note that the Gaussian mixture is:
\begin{equation*}
\pMN(x) = \sum_{i=1}^N w_i \mathcal{N}(x; \mu_i, \sigma_i^2 \mathbf{I}),
\end{equation*}
which can be written as:
\begin{equation*}
\pMN(x) = \mathbb{E}_{\boldsymbol{\sigma}_i \sim p_\sigma} \mathbb{E}_{\mu \sim p^*} \left[ \mathcal{N}(x; \mu, \sigma_i^2 \mathbf{I}) \right],
\end{equation*}
where $p^*(\mu)$ and $p_\sigma(\sigma_i)$ are the discrete distributions over the M centers and N noise levels respectively. 
By using the linearity of expectation and the log-derivative trick, we have
\begin{align*}
\mathcal{T}(\theta) &= \int_x \left\langle s_\theta(x), \frac{\partial}{\partial x} \mathbb{E}_{\boldsymbol{\sigma}_i \sim p_\sigma} \mathbb{E}_{\mu \sim p^*} \left[ \mathcal{N}(x; \mu, \sigma_i^2 \mathbf{I}) \right] \right\rangle dx \\
&= \int_x \left\langle s_\theta(x), \mathbb{E}_{\boldsymbol{\sigma}_i \sim p_\sigma} \mathbb{E}_{\mu \sim p^*} \left[ \frac{\partial \mathcal{N}(x; \mu, \sigma_i^2 \mathbf{I})}{\partial x} \right] \right\rangle dx \\
&= \mathbb{E}_{\boldsymbol{\sigma}_i \sim p_\sigma} \mathbb{E}_{\mu \sim p^*} \int_x \left\langle s_\theta(x), \frac{\partial \mathcal{N}(x; \mu, \sigma_i^2 \mathbf{I})}{\partial x} \right\rangle dx \\
&= \mathbb{E}_{\boldsymbol{\sigma}_i \sim p_\sigma} \mathbb{E}_{\mu \sim p^*} \mathbb{E}_{x \sim \mathcal{N}(\mu, \sigma_i^2 \mathbf{I})} \left[ \left\langle s_\theta(x), \frac{\partial \log \mathcal{N}(x; \mu, \sigma_i^2 \mathbf{I})}{\partial x} \right\rangle \right].
\end{align*}
Therefore,
\begin{equation*}
\mathcal{L}_{ESM}(\theta) = \mathbb{E}_{\pMN(x)} \left[ \frac{1}{2} \|s_\theta(x)\|^2 \right] - \mathbb{E}_{\boldsymbol{\sigma}_i \sim p_\sigma} \mathbb{E}_{\mu \sim p^*} \mathbb{E}_{x \sim \mathcal{N}(\mu, \sigma_i^2 \mathbf{I})} \left[ \left\langle s_\theta(x), \frac{\partial \log \mathcal{N}(x; \mu, \sigma_i^2 \mathbf{I})}{\partial x} \right\rangle \right] + C_1.
\end{equation*}
The denoising score matching loss is defined as
\begin{equation*}
\mathcal{L}_{DSM}(\theta) = \mathbb{E}_{\boldsymbol{\sigma}_i \sim p_\sigma} \mathbb{E}_{\mu \sim p^*} \mathbb{E}_{x \sim \mathcal{N}(\mu, \sigma_i^2 \mathbf{I})} \left[ \frac{1}{2} \left\| s_\theta(x) - \frac{\partial \log \mathcal{N}(x; \mu, \sigma_i^2 \mathbf{I})}{\partial x} \right\|^2 \right]
\end{equation*}
By expanding we obtain
\begin{align*}
\mathcal{L}_{DSM}(\theta) &= \mathbb{E}_{\boldsymbol{\sigma}_i \sim p_\sigma} \mathbb{E}_{\mu \sim p^*} \mathbb{E}_{x \sim \mathcal{N}(\mu, \sigma_i^2 \mathbf{I})} \left[ \frac{1}{2} \|s_\theta(x)\|^2 \right] \\
&\quad - \mathbb{E}_{\boldsymbol{\sigma}_i \sim p_\sigma} \mathbb{E}_{\mu \sim p^*} \mathbb{E}_{x \sim \mathcal{N}(\mu, \sigma_i^2 \mathbf{I})} \left[ \left\langle s_\theta(x), \frac{\partial \log \mathcal{N}(x; \mu, \sigma_i^2 \mathbf{I})}{\partial x} \right\rangle \right] + C_2,
\end{align*}
where $C_2 = \mathbb{E}_{\boldsymbol{\sigma}_i \sim p_\sigma} \mathbb{E}_{\mu \sim p^*} \mathbb{E}_{x \sim \mathcal{N}(\mu, \sigma_i^2 \mathbf{I})} \left[ \frac{1}{2} \left\| \frac{\partial \log \mathcal{N}(x; \mu, \sigma_i^2 \mathbf{I})}{\partial x} \right\|^2 \right]$ is a constant.
Note that:
\begin{equation*}
\mathbb{E}_{\pMN(x)} \left[ \frac{1}{2} \|s_\theta(x)\|^2 \right] = \mathbb{E}_{\boldsymbol{\sigma}_i \sim p_\sigma} \mathbb{E}_{\mu \sim p^*} \mathbb{E}_{x \sim \mathcal{N}(\mu, \sigma_i^2 \mathbf{I})} \left[ \frac{1}{2} \|s_\theta(x)\|^2 \right].
\end{equation*}
Thus, $\mathcal{L}_{ESM}(\theta) = \mathcal{L}_{DSM}(\theta) + C_1 - C_2$, proving the optimization equivalence.

\medskip
\subsection{Analysis of Noise Level Mismatch}
\label{A.7}


Consider sampling from a single data point $\mu \in \mathbb{R}^d$ with a fixed noise schedule. At sampling step $n$, we define the following.

\noindent
\textbf{1. Scheduled noise ($\sigma_n$):} The noise level prescribed by the sampling schedule.

\noindent
\textbf{2. Actual noise ($\sigma_n^*$):} The true noise level that is defined as $\sigma_n^{*2} := \|x_n - \mu\|^2/d$.

\noindent
The score function for this setting is $\nabla_{x} \log p(x_n) = -(x_n - \mu)/\sigma_n^{*2}$. In practice, samples often arrive near the target faster than expected, leading to $\sigma_n^* \ll \sigma_n$. However, solvers compute step sizes based on the schedule $\sigma_n$, not the actual state $\sigma_n^*$.

\subsubsection{Catastrophic Failure of ODE Samplers}
\label{A.7.1}

The ODE update rule for the residual $\mathbf{r}_n = x_n - \mu$ is:
\begin{equation*}
    \mathbf{r}_{n+1} = \mathbf{r}_n \left(1 - \frac{\sigma_n^2 - \sigma_{n+1}^2}{2\sigma_n^{*2}}\right).
\end{equation*}

\noindent
When $\sigma_n^* \ll \sigma_n$, the step coefficient becomes huge:
\begin{equation*}
    \lambda_n := \frac{\sigma_n^2 - \sigma_{n+1}^2}{2\sigma_n^{*2}} \gg 1.
\end{equation*}

\noindent
When $\lambda_n > 2$, the coefficient $(1 - \lambda_n)$ becomes negative with large magnitude, causing catastrophic overshoot:
\begin{equation*}
    \|\mathbf{r}_{n+1}\| = |\lambda_n - 1| \|\mathbf{r}_n\| \approx \lambda_n \|\mathbf{r}_n\| = \frac{\sigma_n^2 - \sigma_{n+1}^2}{2\sigma_n^{*2}} \cdot \sigma_n^* \sqrt{d} \propto \frac{1}{\sigma_n^*}.
\end{equation*}

\noindent
As $\sigma_n^* \to 0$, the overshoot grows without bound, causing sampling to break down completely.

\subsubsection{Self-correction of SDE Samplers}
\label{A.7.2}

The SDE adds a stochastic term to the update:
\begin{equation*}
    \mathbf{r}_{n+1} = \mathbf{r}_n \left(1 - \beta_n\right) + \sqrt{\sigma_n^2 - \sigma_{n+1}^2} \boldsymbol{\epsilon}_n,
\end{equation*}
where $\beta_n := (\sigma_n^2 - \sigma_{n+1}^2)/\sigma_n^{*2}$ and $\boldsymbol{\epsilon}_n \sim \mathcal{N}(0, \mathbf{I})$.

\noindent
\textbf{Self-Correcting Mechanism:} The expected squared distance after the update is:
\begin{align*}
    \mathbb{E}[\|\mathbf{r}_{n+1}\|^2] &= \|\mathbf{r}_n\|^2(1-\beta_n)^2 + d(\sigma_n^2 - \sigma_{n+1}^2) \\
    &= \sigma_n^{*2} d(1-\beta_n)^2 + d \beta_n \sigma_n^{*2} \\
    &= \sigma_n^{*2} d(1 - \beta_n + \beta_n^2).
\end{align*}
The new actual noise level becomes $\sigma_{n+1}^{*2} = \sigma_n^{*2}(1 - \beta_n + \beta_n^2)$.

\noindent
\textbf{The Correction Condition:} The sample maintains appropriate distance when:
\begin{equation*}
    \sigma_{n+1}^{*2} > \sigma_n^{*2} \quad \Leftrightarrow \quad 1 - \beta_n + \beta_n^2 > 1 \quad \Leftrightarrow \quad \beta_n > 1.
\end{equation*}

\noindent
\textbf{The Feedback Loop and Why It Prevents Large Deviations:} 
\begin{enumerate}
    \item When early arrival occurs ($\sigma_n^* \ll \sigma_n$), we get $\beta_n = (\sigma_n^2 - \sigma_{n+1}^2)/\sigma_n^{*2} \gg 1$.
    
    \item The condition $\beta_n > 1$ triggers the correction mechanism: the stochastic term dominates the deterministic drift, and the noise injection effectively controls the distance. Specifically:
    \begin{equation*}
        \sigma_{n+1}^{*2} = \sigma_n^{*2}(1 - \beta_n + \beta_n^2) \approx \sigma_n^{*2} \beta_n^2 = \frac{(\sigma_n^2 - \sigma_{n+1}^2)^2}{\sigma_n^{*2}}
    \end{equation*}
    Since $\beta_n \gg 1$, this gives $\sigma_{n+1}^* \approx \frac{\sigma_n^2 - \sigma_{n+1}^2}{\sigma_n^*} \gg \sigma_n^*$.
    
    \item \textbf{Why this prevents large deviations:} The noise injection creates a lower bound for how small $\sigma_n^*$ can become. Even if $\sigma_n^*$ tries to collapse to zero, the stochastic term ensures that:
    \begin{equation*}
        \sigma_{n+1}^* \ge \sqrt{\sigma_n^2 - \sigma_{n+1}^2}
    \end{equation*}
    This lower bound is determined by the schedule difference $\sqrt{\sigma_n^2 - \sigma_{n+1}^2}$, which is typically on the same order as the scheduled noise levels.
    
    \item \textbf{Automatic regulation:} If $\sigma_n^*$ becomes much smaller than $\sigma_n$, the correction pushes it back up. If $\sigma_n^*$ is already close to $\sigma_n$, then $\beta_n \approx 1$ and the system evolves smoothly without dramatic corrections. This creates a self-correction mechanism that keeps $\sigma_n^*$ within a reasonable range of $\sigma_n$.
\end{enumerate}

\clearpage

\section{Experiments}
In this section, we provide training details and additional experimental results.

\subsection{Implementation Details}
\label{B.1}
\textbf{Model Architecture.} As we mentioned in the main paper, we use the NCSN++ architecture and VE-SDE (our baseline) \cite{song2020score} on all datasets. The only difference is that, for the unconditioning modeling, we remove the noise embedding blocks. All models are unconditional (i.e., without class labels) across different datasets and trained on 4$\times$ NVIDIA-L40S GPUs (45G) and 2$\times$ NVIDIA-H200 GPUs (140G). 

\medskip
\noindent
\textbf{Training Setup.} We train all models with the AdamW \cite{loshchilov2017decoupled} optimizer using learning rate 0.0002. Using a batch size of 128, we train 1M iterations on CIFAR-10; Using a batch size of 64, we train 40,000 iterations on Cat-Caracal, 700,000 iterations on CelebA 64$\times$64, and 700,000 iterations on ImageNet 64$\times$64; Using a batch size of 36, we train 500,000 iterations on CelebA-hq 256$\times$256. Although the reported FIDs are based on the best checkpoints of each setting, for fair comparison, we do not modify other hyperparameters in the baseline paper, e.g., we use EMA decay rate of 0.9999, and all experiments are trained with the PyTorch random seed 42.

\noindent
\textbf{Training Cost and Scalability.}
For the standard setting with a single temperature level ($T=1$), conditioning vs.\ unconditioning only differs in whether the noise is fed into the network.
Consequently, noise unconditioning introduces \emph{no additional asymptotic cost}; in fact, removing the noise embedding slightly reduces both GPU memory and wall-clock time per batch.
When enabling temperature-based score matching, the only extra component is the KNN-based approximation of the explicit score, which naively scales with both dataset size and data dimension and can be prohibitive on very large, high-resolution datasets.
However, this additional cost is avoidable. We provide two simple but effective strategies:

\begin{enumerate}
\item \textbf{Class-restricted neighborhoods for labeled data.}
For datasets with class labels (e.g., ImageNet), we can restrict KNN search to samples within the same class.
In practice, this means we only search over about $1{,}300$ images per query on ImageNet, so the nearest-neighbor step adds almost no overhead.

\item \textbf{KNN in the feature space.}
We perform KNN in a low-dimensional feature space produced by a fixed encoder rather than in pixel space.
This converts a high-dimensional search (e.g., $64\times64\times3$) into a low-dimensional one (hundreds of dimensions), and is compatible with approximate nearest neighbor libraries that scale to tens or hundreds of millions of points.
As a result, the additional cost of temperature smoothing is dominated by the encoder forward pass and becomes essentially independent of image resolution.
\end{enumerate}

\noindent
With these design choices, the extra computation from temperature-based KNN becomes a \emph{small, bounded overhead} rather than a bottleneck, even as the dataset and resolution grow.
Table~\ref{tab:train-cost} reports the empirical per-batch runtime and peak GPU memory under different variants, averaged over the first 1000 iterations on two NVIDIA L40S GPUs.

From Table~\ref{tab:train-cost}, we observe:

\begin{enumerate}
\item \textbf{Unconditioning is essentially free.}
Removing the noise embedding consistently \emph{reduces} runtime and memory relative to the conditioning baseline.
This empirically confirms that noise-unconditioned models do not introduce any extra training overhead and remain compatible with existing large-scale training budgets.

\item \textbf{Feature-space KNN scales like standard diffusion.}
When KNN is computed in feature space, the per-batch runtime and memory are virtually identical to the conditioning baseline (e.g., CIFAR-10: 0.3551\,s vs.\ 0.3540\,s and 16.22\,GB vs.\ 16.86\,GB; CelebA: 0.4115\,s vs.\ 0.4168\,s and 45.02\,GB vs.\ 45.02\,GB).
This indicates that temperature smoothing with feature-space KNN behaves, in practice, as a constant-factor modification to standard diffusion training, and does not change the overall scaling with model size or data size.

\item \textbf{Naive pixel-space KNN is the only non-scalable variant—and is unnecessary.}
Direct KNN in pixel space leads to noticeable overhead on higher resolutions (CelebA: 0.4115\,s $\to$ 0.5066\,s and 45.02\,GB $\to$ 52.50\,GB), which is expected because the cost grows with the ambient dimension.
Since class-restricted and feature-space KNN both avoid this issue and empirically perform better, there is no need to rely on the pixel-space variant in large-scale settings.
\end{enumerate}

\begin{table}[!htb]
\centering
\caption{Per-batch training cost (time and peak GPU memory) under different model variants. Runtimes are measured on two NVIDIA L40S GPUs and averaged over the first 1000 batches. ``KNN-pixel'' and ``KNN-feature'' denote temperature-based score matching with KNN computed in pixel space and feature space, respectively (all with $T_i = 5/\sigma_i$, $K=30$).}
\label{tab:train-cost}
\begin{tabular}{lcccc}
\toprule
Dataset \& Resolution & Method & Time / batch (s) & Mem (GB) & Batch size \\
\midrule
\multirow{4}{*}{CIFAR-10 $32\times32$} 
  & Conditioning      & 0.3551 & 16.22 & 128 \\
  & Unconditioning    & 0.3371 & 16.13 & 128 \\
  & KNN-pixel         & 0.3466 & 16.40 & 128 \\
  & KNN-feature       & 0.3540 & 16.86 & 128 \\
\midrule
\multirow{4}{*}{Cat-Caracal $64\times64$}
  & Conditioning      & 0.4075 & 24.74 & 64 \\
  & Unconditioning    & 0.3957 & 24.72 & 64 \\
  & KNN-pixel         & 0.3976 & 24.92 & 64 \\
  & KNN-feature       & 0.4031 & 24.98 & 64 \\
\midrule
\multirow{4}{*}{CelebA $64\times64$}
  & Conditioning      & 0.4115 & 45.02 & 64 \\
  & Unconditioning    & 0.3946 & 45.01 & 64 \\
  & KNN-pixel         & 0.5066 & 52.50 & 64 \\
  & KNN-feature       & 0.4168 & 45.02 & 64 \\
\bottomrule
\end{tabular}
\end{table}

Overall, Noise Unconditioning and Temperature Smoothing can be trained with virtually the \emph{same} computational budget as standard diffusion models.
The only non-negligible overhead arises from a deliberately naive KNN implementation in pixel space, which is avoidable in practice.
By restricting neighborhoods (for labeled datasets) and operating in a compact feature space, our method maintains the same asymptotic scaling as conventional diffusion training and can be plugged into existing large diffusion pipelines without increasing the number of GPUs or the total training time.

\medskip
\noindent

\clearpage
\subsection{Numerical Verifications}
In this section, we provide numerical verifications for our key claims and check that the asymptotic, high-dimensional arguments used in the analysis are already accurate at the finite dimensions relevant for modern diffusion models. We design experiments that mirror our theoretical setup as closely as possible (same shell-touch noise schedules, same effective weights, same temperature smoothing) and then compare empirical measurements with the corresponding closed-form predictions. The first experiment validates the Gaussian shell phenomenon and quantifies how tightly Gaussian mass concentrates around its typical radius in moderate to high dimensions. The second experiment tests Proposition 2.1 (\cref{A.4.1}) by directly measuring the weight ratios between adjacent noise levels under the shell-touch schedule, demonstrating that a single ``optimal'' noise level overwhelmingly dominates its neighbors even at realistic image resolutions. The third experiment probes the relationship between the learned score and the empirical mixture score on CIFAR-10 and Cat–Caracal via cosine similarity, providing concrete evidence for our claim that the neural network learns a smooth score field aligned with local data manifolds rather than reproducing the highly irregular empirical score. Finally, we compute Jacobians of the empirical score near overlap regions to validate the expansiveness-based analysis in \cref{A.5}, confirming that unconditioning and temperature smoothing substantially reduce local expansion and thereby enable generalization.

\subsubsection{Gaussian shell}
\label{B.2.1}

In this section, we numerically verify Proposition~1 about the ``Gaussian shell''
phenomenon and check how accurate the asymptotic approximation is in finite
dimensions.

Recall Proposition~1: Let $X \sim \mathcal{N}(0, \sigma^2 \mathbf{I})$ and
$R = \|X\|$. As $d \to \infty$,
\[
R \xrightarrow{P} \sigma\sqrt{d}, 
\qquad
R \xrightarrow{d} \mathcal{N}\!\left(\sigma\sqrt{d},\, \frac{\sigma^2}{2}\right),
\]
so that about $99.7\%$ of the probability mass concentrates in a thin spherical
shell of radius $\sigma\sqrt{d}$ and thickness $3\sqrt{2}\,\sigma$:
\[
\Pr\!\left(
\sigma\sqrt{d} - \frac{3}{\sqrt{2}}\sigma
\;\le R \le\;
\sigma\sqrt{d} + \frac{3}{\sqrt{2}}\sigma
\right)
\approx 99.73\%.
\]

\paragraph{Experimental setup.}
We perform Monte Carlo experiments for several dimensions
$d \in \{50, 100, 500, 1000\}$.
For each $d$:

\begin{enumerate}
  \item We choose $N=10$ noise scales $\{\sigma_i\}_{i=1}^{10}$ that follow the
  ``shell-touch'' schedule from \cref{A.3}:
  \begin{equation*}
    \sigma_i \sqrt{d} + \frac{3}{\sqrt{2}}\sigma_i
    \;=\;
    \sigma_{i+1}\sqrt{d} - \frac{3}{\sqrt{2}}\sigma_{i+1},
    \label{eq:shell-touch-numerical}
  \end{equation*}
  so that the outer $3\sigma$ boundary of shell $i$ coincides with the inner
  $3\sigma$ boundary of shell $i+1$, packing the radius axis without gaps.

  \item For each pair $(d,\sigma_i)$, we draw $10^6$ i.i.d.\ samples
  \[
    X \sim \mathcal{N}(0, \sigma_i^2 \mathbf{I}),\qquad R = \|X\|,
  \]
  and compute the empirical probability
  \[
  \hat{p}_{d,i}
  \;=\;
  \Pr_{\text{emp}}\!\Big(
    \sigma_i\sqrt{d} - 3\sigma_i/\sqrt{2}
    \;\le R \le\;
    \sigma_i\sqrt{d} + 3\sigma_i/\sqrt{2}
  \Big),
  \]
  i.e., the fraction of samples whose radius falls inside the theoretical shell.

  \item For each dimension $d$, we also keep the per-shell statistics
  $\{\hat{p}_{d,i}\}_{i=1}^{10}$, and compute
  \(
    \bar{p}_d = \frac{1}{N}\sum_{i=1}^N \hat{p}_{d,i}
  \)
  together with the standard deviation over $i$.
\end{enumerate}

\begin{figure}[!htb]
    \centering
    \includegraphics[width=\linewidth]{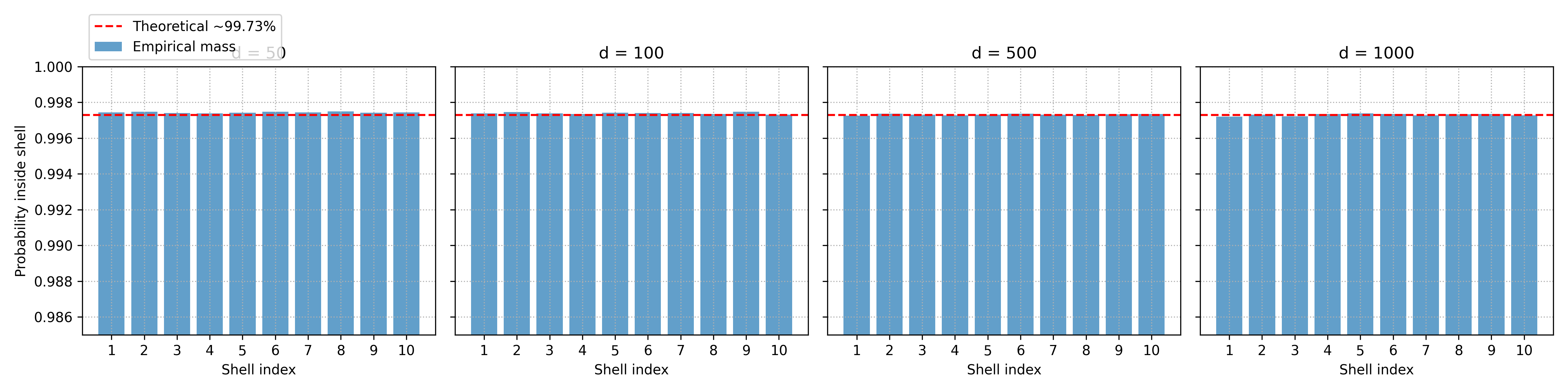}
    \caption{
    Empirical probability mass inside the theoretical Gaussian shell
    $\left[\sigma_i\sqrt{d} - 3\sigma_i/\sqrt{2},\, \sigma_i\sqrt{d} + 3\sigma_i/\sqrt{2}\right]$
    for each of the $N=10$ shells (x-axis) and for
    $d \in \{50, 100, 500, 1000\}$.
    The dashed red line marks the theoretical value $99.73\%$.
    For all dimensions and all shells, the empirical mass is tightly
    concentrated around this value, supporting the Gaussian shell
    approximation in finite dimensions.
    }
    \label{fig:shell-masses-per-dim}
\end{figure}

\paragraph{Results.}
Figure~\ref{fig:shell-masses-per-dim} shows the empirical shell mass
$\hat{p}_{d,i}$ for all shells $i=1,\dots,10$ and all dimensions
$d \in \{50,100,500,1000\}$.
Across all settings, each individual shell contains very close to
$99.73\%$ of the mass: the bars cluster tightly around the theoretical
value with deviations of order $10^{-4}$.

\begin{figure}[!htb]
    \centering
    \includegraphics[width=0.6\linewidth]{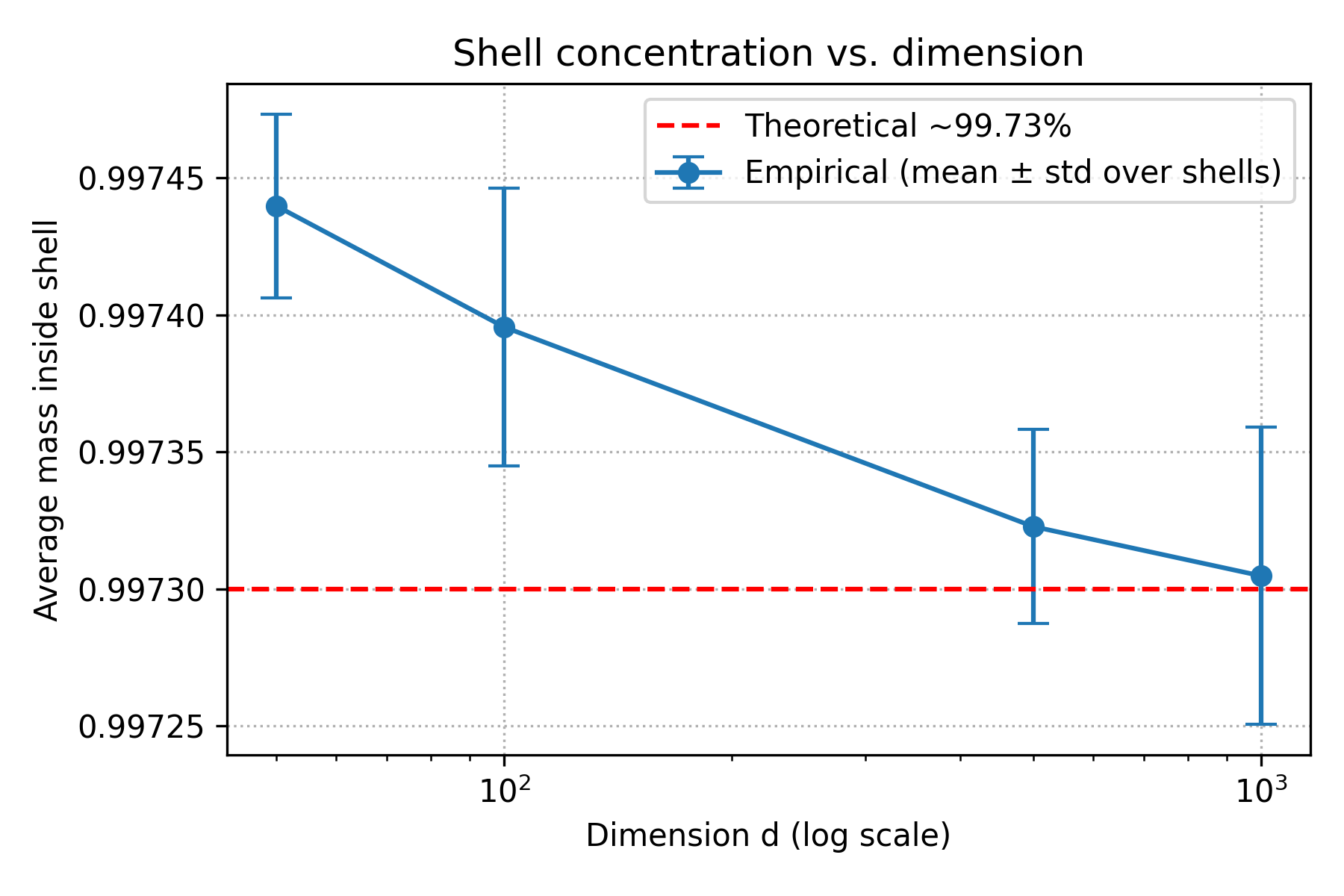}
    \caption{
    Average empirical probability mass inside the theoretical Gaussian shells
    as a function of dimension $d$.
    For each $d$, we average over $N=10$ noise levels $\{\sigma_i\}$ chosen by
    the shell-touch schedule and show mean $\bar{p}_d$ with error bars
    (standard deviation over shells).
    The values concentrate tightly around the theoretical $99.73\%$ for all
    $d$ and move slightly closer as $d$ increases, confirming that the Gaussian
    shell approximation remains accurate in moderate to high dimensions.
    }
    \label{fig:mass-vs-d}
\end{figure}

Aggregating over shells, the raw logs and Figure~\ref{fig:mass-vs-d} show that,
for all tested dimensions, the empirical shell masses remain extremely close to
the theoretical $99.73\%$:
\begin{itemize}
  \item For $d=50$ and $d=100$, most $\hat{p}_{d,i}$ values lie in the range
  $[0.99740, 0.99755]$.
  \item For $d=500$ and $d=1000$, the averages $\bar{p}_d$ are even closer to
  $0.9973$, with deviations on the order of $10^{-4}$.
  \item The standard deviation across shells is also about $10^{-4}$, matching
  the theoretical standard deviation of a Bernoulli ($p\approx0.9973$) mean
  with $10^6$ samples, confirming that the remaining fluctuations are due to
  finite-sample noise rather than systematic bias.
\end{itemize}

\begin{figure}[!htb]
    \centering
    \includegraphics[width=0.7\linewidth]{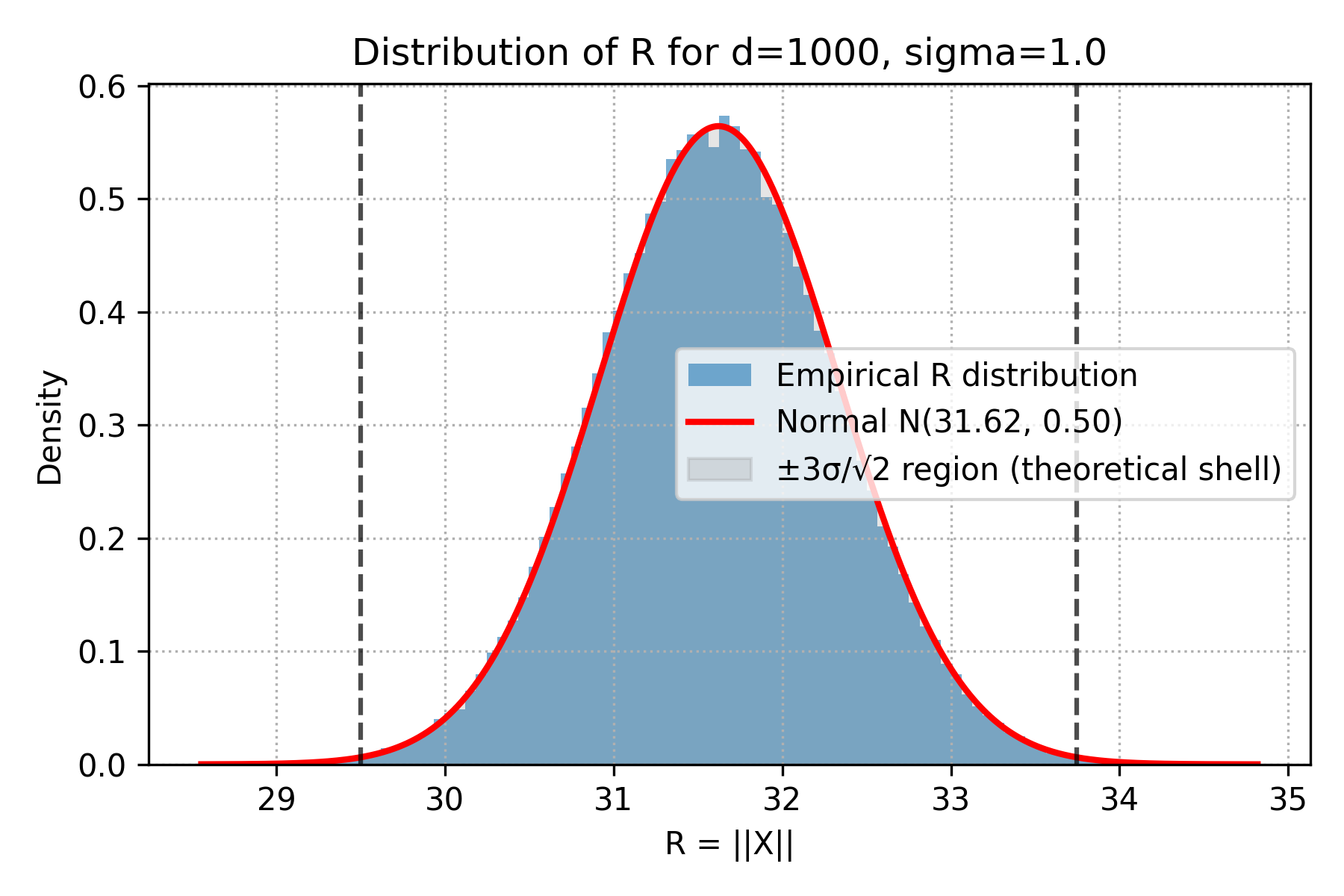}
    \caption{
    Empirical distribution of the radius $R = \|X\|$ for
    $X \sim \mathcal{N}(0, I_{1000})$ (histogram) together with the Gaussian
    approximation $\mathcal{N}(\sqrt{1000},\, 0.5)$ from Proposition~1
    (red curve).
    The fitted normal density closely matches the empirical histogram, and
    the interval $\big[\sqrt{1000} - 3/\sqrt{2},\, \sqrt{1000} + 3/\sqrt{2}\big]$
    (vertical dashed lines) captures almost all of the mass, consistent with
    the $3\sigma$ rule used in the Gaussian shell approximation.
    }
    \label{fig:r-hist-d1000}
\end{figure}

Figure~\ref{fig:r-hist-d1000} visualizes the empirical distribution of $R$
for $d=1000$ and $\sigma=1$ together with the one-dimensional Gaussian
approximation from Proposition~1, namely
\[
R \approx \mathcal{N}\!\big(\sigma\sqrt{d},\, \sigma^2/2\big)
= \mathcal{N}\!\big(\sqrt{1000},\, 1/2\big).
\]
The fitted normal curve almost perfectly matches the histogram, and the
interval
\[
\big[\sigma\sqrt{d} - 3\sigma/\sqrt{2},\; \sigma\sqrt{d} + 3\sigma/\sqrt{2}\big]
= \big[\sqrt{1000} - 3/\sqrt{2},\; \sqrt{1000} + 3/\sqrt{2}\big]
\]
(corresponding to the theoretical ``$3\sigma$ shell'') covers the overwhelming
majority of the mass. This confirms that, even in relatively high but finite
dimension, the one-dimensional normal approximation of $R$ and the associated
shell description remain highly accurate.

Overall, this experiment numerically validates the geometric picture used throughout our analysis: in high dimensions, each Gaussian component of the empirical mixture can be regarded as a thin shell centered at a training sample.

\subsubsection{Noise level domination}
\label{B.2.2}

In Proposition~2.1, we show that the empirical score weights across different noise levels are extremely sharp: For a fixed training point $\mu_*$ and sampling position $x$, the weight associated with the ``optimal'' noise level $\sigma_i$ dominates that of its neighboring noise level $\sigma_{i+1}$ by an exponential factor. In the notation of  \cref{A.4.1}, this domination is captured by the closed-form approximation
\[
\frac{w_{ij}(x)}{w_{i+1,j}(x)}
\;\approx\;
e^{18(1-2\alpha)},
\]
where $\alpha \in [0,0.5)$ quantifies how much the optimal noise scale $\sigma_{\mathrm{opt}}$ is closer to
shell $i$ than to shell $i+1$. Crucially, this ratio does not depend on the dimension $d$, and is expected
to be extremely large whenever $\alpha$ is noticeably smaller than $1/2$, thus justifying the
``$\sigma$-domination'' step in our analysis.

\paragraph{Experimental setup.}
We numerically validate this domination effect in dimensions corresponding to common image resolutions,
as well as in a very high-dimensional setting. For each spatial resolution
$H \times W \in \{32 \times 32, 64 \times 64, 128 \times 128, 256 \times 256\}$,
we set $d = 3HW$ and construct a pair of adjacent noise scales $(\sigma_i, \sigma_{i+1})$ that satisfy the
shell-touch condition from \cref{A.3}:
\[
\sigma_i \sqrt{d} + \frac{3}{\sqrt{2}} \sigma_i
\;=\;
\sigma_{i+1} \sqrt{d} - \frac{3}{\sqrt{2}} \sigma_{i+1}.
\]
We additionally consider a $d = 10^{12}$ experiment to validate the infinite-dimensional limit predicted by our theory.

We fix the center at $\mu_* = 0$ (by translation invariance). For a given $\alpha \in (0,1/2)$, we place
the sampling point $x_\alpha$ on the line segment connecting the typical radii of shells $i$ and $i+1$ by
setting
\[
r_\alpha
\;:=\;
\|x_\alpha\|
\;=\;
(1-\alpha)\,\sigma_i \sqrt{d}
\;+\;
\alpha\,\sigma_{i+1} \sqrt{d}.
\]
For such $x_\alpha$, we compute the empirical ratio between the two noise-level weights for the same center
$\mu_*$ using the effective weight $w^E$ from \cref{A.4}:
\[
\log w^E_{ij}(x)
= -(d-2)\ln\sigma_i
- \frac{\|x-\mu_*\|^2}{2\sigma_i^2},
\qquad
\frac{w^E_{ij}(x_\alpha)}{w^E_{i+1,j}(x_\alpha)}
=
\exp\!\Big(\log w^E_{ij}(x_\alpha) - \log w^E_{i+1,j}(x_\alpha)\Big).
\]
We then compare this empirical ratio with the theoretical prediction
$e^{18(1-2\alpha)}$ for three values of $\alpha$:
\[
\alpha \in \Bigl\{\tfrac{1}{10},\ \tfrac{1}{6},\ \tfrac{1}{3}\Bigr\}.
\]

\paragraph{Results.}
Table~\ref{tab:sigma-domination-ratios} summarizes the empirical and theoretical ratios
for different dimensions and $\alpha$ values. 

\begin{table}[!htb]
  \centering
  \caption{
  Empirical vs.\ theoretical dominance ratios
  $\frac{w^E_{ij}(x_\alpha)}{w^E_{i+1,j}(x_\alpha)}$
  for different image resolutions and interpolation parameters
  $\alpha \in \{\tfrac{1}{10}, \tfrac{1}{6}, \tfrac{1}{3}\}$.
  We report the theoretical prediction
  $e^{18(1-2\alpha)}$ (top block) and the empirical ratios
  computed from the effective weights $w^E$ under the
  shell-touch schedule (bottom block).}
  \label{tab:sigma-domination-ratios}
  \vspace{0.5em}
  \small
  \begin{tabular}{c|ccc}
    \toprule
    Resolution / $d$ & $\alpha = \tfrac{1}{10}$ & $\alpha = \tfrac{1}{6}$ & $\alpha = \tfrac{1}{3}$ \\
    \midrule
    \multicolumn{4}{c}{\textbf{Theoretical ratio} $e^{18(1-2\alpha)}$} \\
    \midrule
    $32 \times 32$            & $1.794\times 10^{6}$ & $1.628\times 10^{5}$ & $4.034\times 10^{2}$ \\
    $64 \times 64$            & $1.794\times 10^{6}$ & $1.628\times 10^{5}$ & $4.034\times 10^{2}$ \\
    $128 \times 128$          & $1.794\times 10^{6}$ & $1.628\times 10^{5}$ & $4.034\times 10^{2}$ \\
    $256 \times 256$          & $1.794\times 10^{6}$ & $1.628\times 10^{5}$ & $4.034\times 10^{2}$ \\
    $d = 10^{12}$              & $1.794\times 10^{6}$ & $1.628\times 10^{5}$ & $4.034\times 10^{2}$ \\
    \midrule
    \multicolumn{4}{c}{\textbf{Empirical ratio} $\frac{w^E_{ij}(x_\alpha)}{w^E_{i+1,j}(x_\alpha)}$} \\
    \midrule
    $32 \times 32$            & $7.240\times 10^{5}$ & $6.977\times 10^{4}$ & $1.907\times 10^{2}$ \\
    $64 \times 64$            & $1.129\times 10^{6}$ & $1.057\times 10^{5}$ & $2.763\times 10^{2}$ \\
    $128 \times 128$          & $1.419\times 10^{6}$ & $1.309\times 10^{5}$ & $3.335\times 10^{2}$ \\
    $256 \times 256$          & $1.595\times 10^{6}$ & $1.459\times 10^{5}$ & $3.667\times 10^{2}$ \\
    $d = 10^{12}$              & $1.794\times 10^{6}$ & $1.628\times 10^{5}$ & $4.034\times 10^{2}$ \\
    \bottomrule
  \end{tabular}
\end{table}

We observe that:

\begin{itemize}
  \item For $\alpha = 1/10$, the theoretical ratio is $e^{18(1-2\alpha)} \approx 1.794\times 10^{6}$. The empirical ratios across $32\times32$, $64\times64$, $128\times128$ and $256\times256$ image dimensions lie in the range $[7.2\times 10^{5},\,1.6\times 10^{6}]$, and move steadily closer to the theoretical value as $d$ increases. At $d = 10^{12}$, the empirical ratio is indeed identical to the theoretical prediction.
  \item For $\alpha = 1/6$, the theoretical ratio is $e^{18(1-2\alpha)} \approx 1.628\times 10^{5}$. The empirical ratios lie between $7.0\times 10^{4}$ and $1.5\times 10^{5}$ across all tested resolutions, again approaching the theoretical prediction as $d$ grows, and finally coinciding with it.
  \item For $\alpha = 1/3$, the theoretical ratio is $e^{6} \approx 4.034\times 10^{2}$. The empirical ratios range from about $1.9\times 10^{2}$ at $32\times32$ to $3.7\times 10^{2}$ at $256\times256$, and reach $4.034\times 10^{2}$ at $d = 10^{12}$, matching the theoretical value.
\end{itemize}

These numerical results confirm the $\sigma$-domination effect in Proposition~2.1 under the same shell-touch schedule used in our main analysis. Even when the ``optimal'' noise level $\sigma_{\mathrm{opt}}$ is only slightly closer to shell $i$ than to shell $i+1$, the corresponding weight ratio $w^E_{ij}(x)/w^E_{i+1,j}(x)$ is already on the order of $10^2$–$10^6$ in realistic image dimensions, and converges cleanly to the dimension-free theoretical limit as $d\to\infty$ (as illustrated by the $d=10^{12}$ experiment). This numerically justifies the approximation of keeping only the dominating noise level when analyzing the empirical score function, as done throughout Appendix~\ref{A.4}.

\subsubsection{The difference between the learned score function and empirical score function}
\label{B.2.3}

\paragraph{Generalization case on the CIFAR-10 dataset.} Concretely, we randomly select $10000$ images $\{\mu_j\}$ from the CIFAR-10 training set and add Gaussian perturbations
with noise levels $\sigma \in \{0.01, 0.05, 0.1\}$:
\[
x = \mu_j + \sigma \varepsilon, \qquad \varepsilon \sim \mathcal{N}(0, \mathbf{I}).
\]
For each perturbed sample $x$, the empirical score function of the training Gaussian mixture at noise level $\sigma$
still points toward its corresponding clean image $\mu_j$:
\[
\nabla_x \log p^\mathrm{emp}_\sigma(x)
\propto \mu_j - x,
\]
 We then compute the cosine similarity between the network-predicted score
$s_\theta(x,\sigma)$ (from our trained VE-SDE model) and this empirical score:
\[
\mathrm{cos\_sim}(x;\sigma)
=
\frac{\big\langle s_\theta(x,\sigma),\; \nabla_x \log p^\mathrm{emp}_\sigma(x)\big\rangle}
{\|s_\theta(x,\sigma)\|_2 \cdot \|\nabla_x \log p^\mathrm{emp}_\sigma(x)\|_2},
\]
and report the average over the $10000$ perturbed samples.

The resulting average cosine similarities are
\[
\mathbb{E}[\mathrm{cos\_sim}(x;0.01)] \approx 0.7383,\quad
\mathbb{E}[\mathrm{cos\_sim}(x;0.05)] \approx 0.9038,\quad
\mathbb{E}[\mathrm{cos\_sim}(x;0.10)] \approx 0.9266,
\]
corresponding to angular deviations of approximately $43^\circ$, $25^\circ$, and $22^\circ$, respectively.

These angles are clearly non-zero, indicating that the learned score does not exactly match the empirical score function around each training point. At the same time, the deviations are only moderately large ($22^\circ$--$43^\circ$), so the network outputs still remain aligned with the empirical directions.

\paragraph{Memorization case on the Cat-Caracal dataset.}
We additionally perform the same analysis on the Cat--Caracal dataset, where we empirically observe strong memorization.
Concretely, we randomly select $1000$ training images $\{\mu_j\}$ and add Gaussian perturbations with noise level
$\sigma = 0.01$:
\[
x = \mu_j + 0.01\,\varepsilon,\qquad \varepsilon \sim \mathcal{N}(0,\mathbf{I}).
\]
For each perturbed sample $x$, the empirical score of the training Gaussian mixture at noise level $\sigma$ still points
toward its corresponding clean image $\mu_j$:
\[
\nabla_x \log p^\mathrm{emp}_{0.01}(x)\;\propto\; \mu_j - x.
\]
Using the conditioning model on Cat--Caracal, we compute the cosine similarity between the network-predicted score
$s_\theta(x,0.01)$ and this empirical score,
\[
\mathrm{cos\_sim}(x;0.01)
=
\frac{\big\langle s_\theta(x,0.01),\; \nabla_x \log p^\mathrm{emp}_{0.01}(x)\big\rangle}
{\|s_\theta(x,0.01)\|_2 \cdot \|\nabla_x \log p^\mathrm{emp}_{0.01}(x)\|_2},
\]
and report the average over the $1000$ perturbed samples. On Cat--Caracal we obtain
\[
\mathbb{E}[\mathrm{cos\_sim}(x;0.01)] \approx 0.9942,
\]
corresponding to an average angular deviation of only about $6^\circ$. Compared to the CIFAR-10 case at the same noise level $\sigma = 0.01$, where the average similarity is about $0.7383$ ($\approx 43^\circ$), the noise conditioning NN on Cat-Caracal learns a score field that is much closer to the empirical score function around each training point.

Taken together with the CIFAR-10 results, these findings support our main claim about how diffusion models balance memorization and generalization through the smoothness of the learned score field. On the small Cat-Caracal dataset, the empirical score function is relatively simple: there are fewer training points and thus fewer regions where Gaussian shells from different centers overlap, so the empirical score field has limited complexity. In this regime, the conditioning network can effectively fit the empirical score: $s_\theta(x,\sigma)$ is almost collinear with $\mu_j - x$ at low noise, as reflected by the cosine similarity $\approx 0.9942$ (about $6^\circ$). In contrast, on CIFAR-10 the much larger number of training samples leads to substantial overlap between shells associated with different centers, making the empirical score field highly irregular and complex in these overlap regions. Correspondingly, the learned score exhibits noticeably lower cosine similarities (angles of $22^\circ$–$43^\circ$ across $\sigma \in \{0.01, 0.05, 0.1\}$), indicating that the model does \emph{not} match the empirical score function exactly, but instead learns a smoothed approximation that remains well aligned while deviating by a nontrivial angle. This smoothed score field is still sufficiently accurate to guide the dynamics toward meaningful images, and is better interpreted as being governed by local data manifolds formed by multiple nearby training samples rather than by pointwise contributions of single examples. The contrast between Cat-Caracal (where the network can closely track the relatively simple empirical score) and CIFAR-10 (where it learns a smoother approximation to a highly complex empirical score) thus provides direct empirical evidence for our explanation framework.

\clearpage

\subsubsection{Empirical validation of the Hessian-spectrum analysis}
\label{B.2.4}

We empirically validate the Hessian-based theory in the general multi-center setting under the optimal-shell approximation:
\[
H(x;T)
=
\nabla_x^2 \log \pMN(x,T)
=
\frac{1}{T}\,\mathrm{Cov}_{w^\ast(x;T)}\!\bigl(a_j(x)\bigr)
-
c_T(x)\mathbf I,
\]
with
\[
c_T(x)=\sum_{j=1}^M \frac{w_j^\ast(x;T)}{\sigma_j^{\ast 2}},
\qquad
\mu_{\max}(x;T)=\frac{\lambda_{\max}(x;T)}{T}-c_T(x),
\qquad
\mu_{\min}(x;T)=\frac{\lambda_{\min}(x;T)}{T}-c_T(x).
\]

\paragraph{Setup.}
We sample \(M=200\) CIFAR-10 training images as centers \(\{\mu_j\}_{j=1}^M\subset[0,1]^d\), \(d=3072\), and evaluate \(K=1000\) points near pairwise perpendicular-bisector regions:
\[
x=\frac{\mu_a+\mu_b}{2}+\epsilon,\quad
\epsilon\sim\mathcal N(0,(5\times10^{-3})^2\mathbf I),
\]
followed by clamping to \([0,1]^d\). This probes overlap regions where inter-center competition is strongest.

Temperatures are
\[
T\in\{1,5,10,50,100,200,300,400,500,600,700,800,900,1000,2000,5000,10^4,10^5,10^6\}.
\]
For each \((x,T)\), we compute \(\lambda_{\max}\), \(\lambda_{\min}\), and \(c_T\), then derive \(\mu_{\max},\mu_{\min}\). Since \(d>M\), \(\mathrm{Cov}_{w^\ast}(a_j)\) is rank-deficient with rank at most \(M-1\), so \(\lambda_{\min}=0\) numerically.

\paragraph{Two radii used in this section.}
To avoid notation ambiguity, we distinguish:
\[
\rho_H(x;T):=\max_k |\mu_k(x;T)|
\]
(Hessian/score-Jacobian spectral radius), while keeping the expansiveness factor definition unchanged:
\[
\rho(x;T):=\max_k |1+\eta \mu_k(x;T)|.
\]
In this section, \(\eta=10^{-3}\), and the reported ``\(\rho(x;T)\)'' corresponds to the one-step linearized expansiveness factor above.

\begin{table}[!htb]
\centering
\caption{Mean statistics over \(K=1000\) overlap-region points across temperatures, for $\eta=10^{-3}$.}
\label{tab:rho_table}
\setlength{\tabcolsep}{3.5pt}
\renewcommand{\arraystretch}{0.9}
\scriptsize
\resizebox{0.60\linewidth}{!}{%
\begin{tabular}{c|ccccc}
\toprule
\(T\) & \(\mathbb E[\lambda_{\max}]\) & \(\mathbb E[\lambda_{\max}/T]\) & \(\mathbb E[\lambda_{\min}/T]\) & \(\mathbb E[c_T]\) & \(\mathbb E[\rho(x;T)]\) \\
\midrule
1      & 26816.0223 & 26816.0223 & 0.0000 & 50.4309 & 27.7656 \\
5      & 67954.6794 & 13590.9359 & 0.0000 & 50.4159 & 14.5405 \\
10     & 76676.0359 & 7667.6036  & 0.0000 & 50.4043 & 8.6172 \\
50     & 84104.2276 & 1682.0846  & 0.0000 & 50.2042 & 2.6319 \\
100    & 85205.1677 & 852.0517   & 0.0000 & 49.5066 & 1.8025 \\
200    & 77766.2678 & 388.8313   & 0.0000 & 46.2252 & 1.3426 \\
300    & 62384.5206 & 207.9484   & 0.0000 & 40.9646 & 1.1670 \\
400    & 46225.8263 & 115.5646   & 0.0000 & 35.4227 & 1.0801 \\
500    & 33595.9590 & 67.1919    & 0.0000 & 30.8225 & 1.0364 \\
600    & 25034.3448 & 41.7239    & 0.0000 & 27.4226 & 1.0143 \\
700    & 19515.5761 & 27.8794    & 0.0000 & 25.0035 & 1.0029 \\
800    & 15952.2970 & 19.9404    & 0.0000 & 23.2722 & 0.9967 \\
900    & 13593.5033 & 15.1039    & 0.0000 & 22.0027 & 0.9931 \\
1000   & 11979.0346 & 11.9790    & 0.0000 & 21.0442 & 0.9909 \\
2000   & 7470.2625  & 3.7351     & 0.0000 & 17.3934 & 0.9863 \\
5000   & 6466.0326  & 1.2932     & 0.0000 & 15.6673 & 0.9856 \\
10000  & 6289.8363  & 0.6290     & 0.0000 & 15.1546 & 0.9855 \\
100000 & 6162.6676  & 0.0616     & 0.0000 & 14.7166 & 0.9853 \\
1000000& 6151.1046  & 0.0062     & 0.0000 & 14.6739 & 0.9853 \\
\bottomrule
\end{tabular}%
}
\end{table}

\begin{figure}[!htb]
\centering
\includegraphics[width=0.6\linewidth]{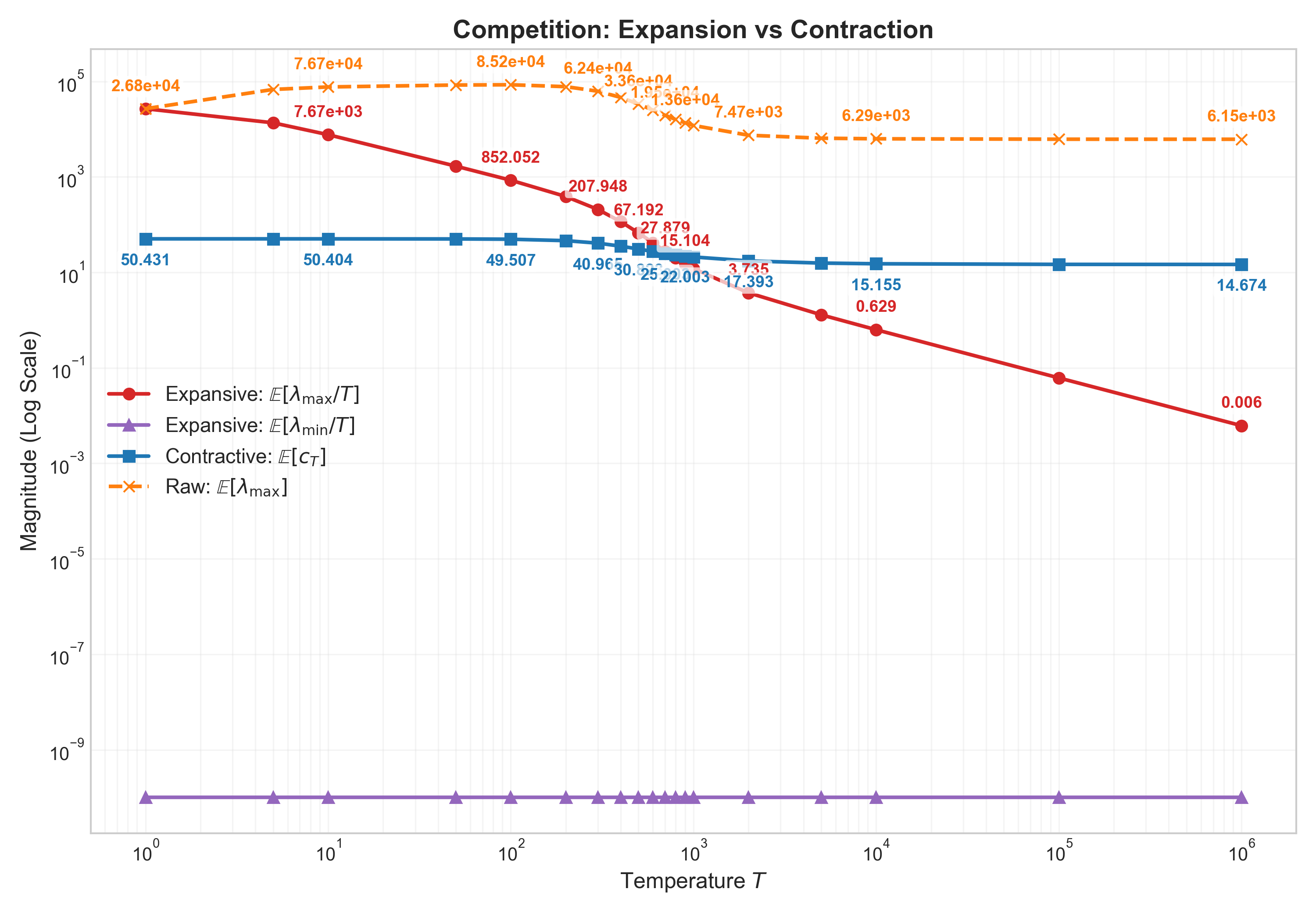}
\caption{Competition between expansion and contraction across temperature: \(\mathbb E[\lambda_{\max}/T]\), \(\mathbb E[\lambda_{\min}/T]\), \(\mathbb E[c_T]\), and raw \(\mathbb E[\lambda_{\max}]\).}
\label{fig:rho_fig1}
\end{figure}

\begin{figure}[!htb]
\centering
\includegraphics[width=0.6\linewidth]{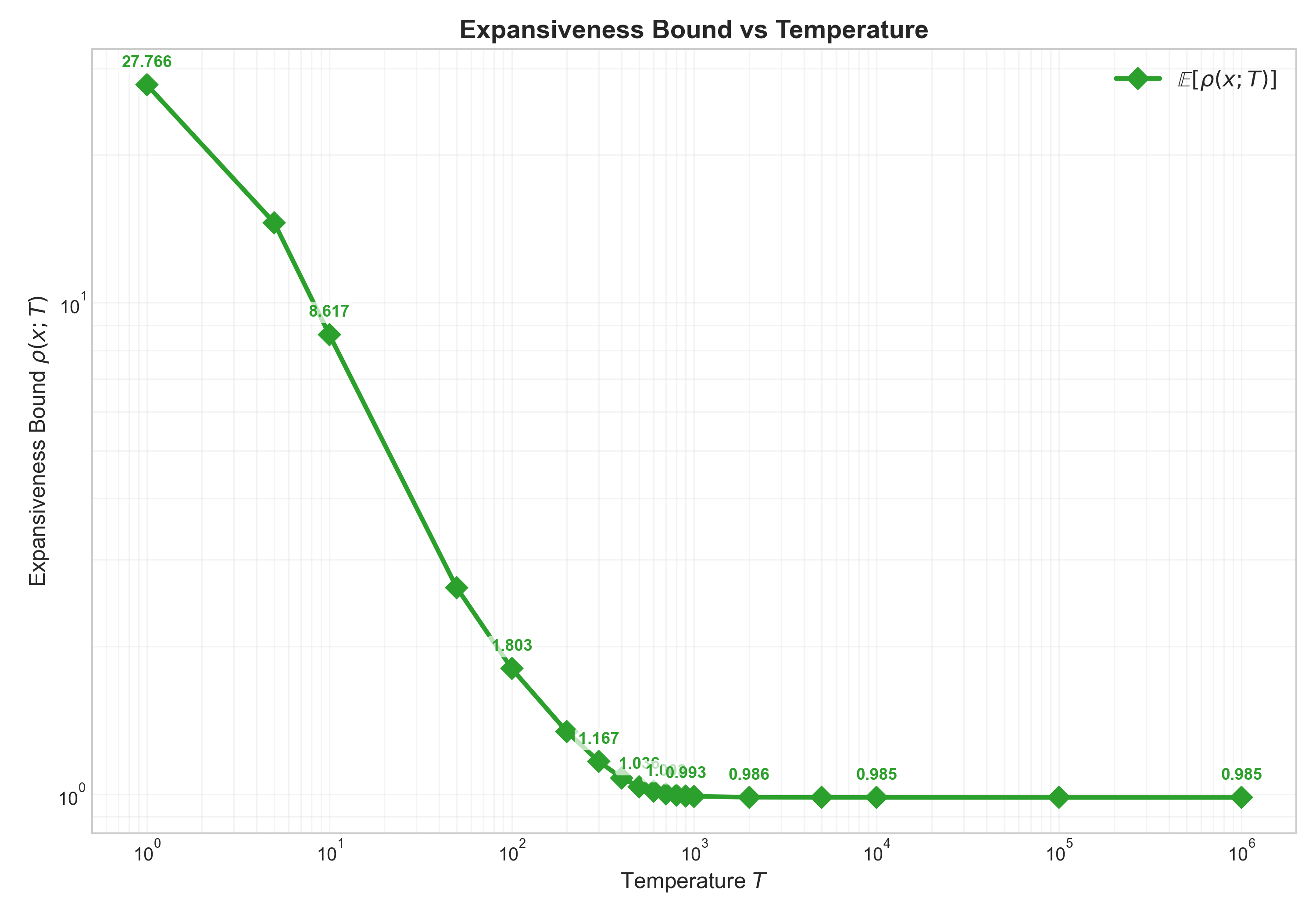}
\caption{Expansiveness bound \(\mathbb E[\rho(x;T)]\) versus temperature \(T\) (\(\eta=10^{-3}\)).}
\label{fig:rho_fig2}
\end{figure}

\begin{figure}[!htb]
\centering
\includegraphics[width=0.6\linewidth]{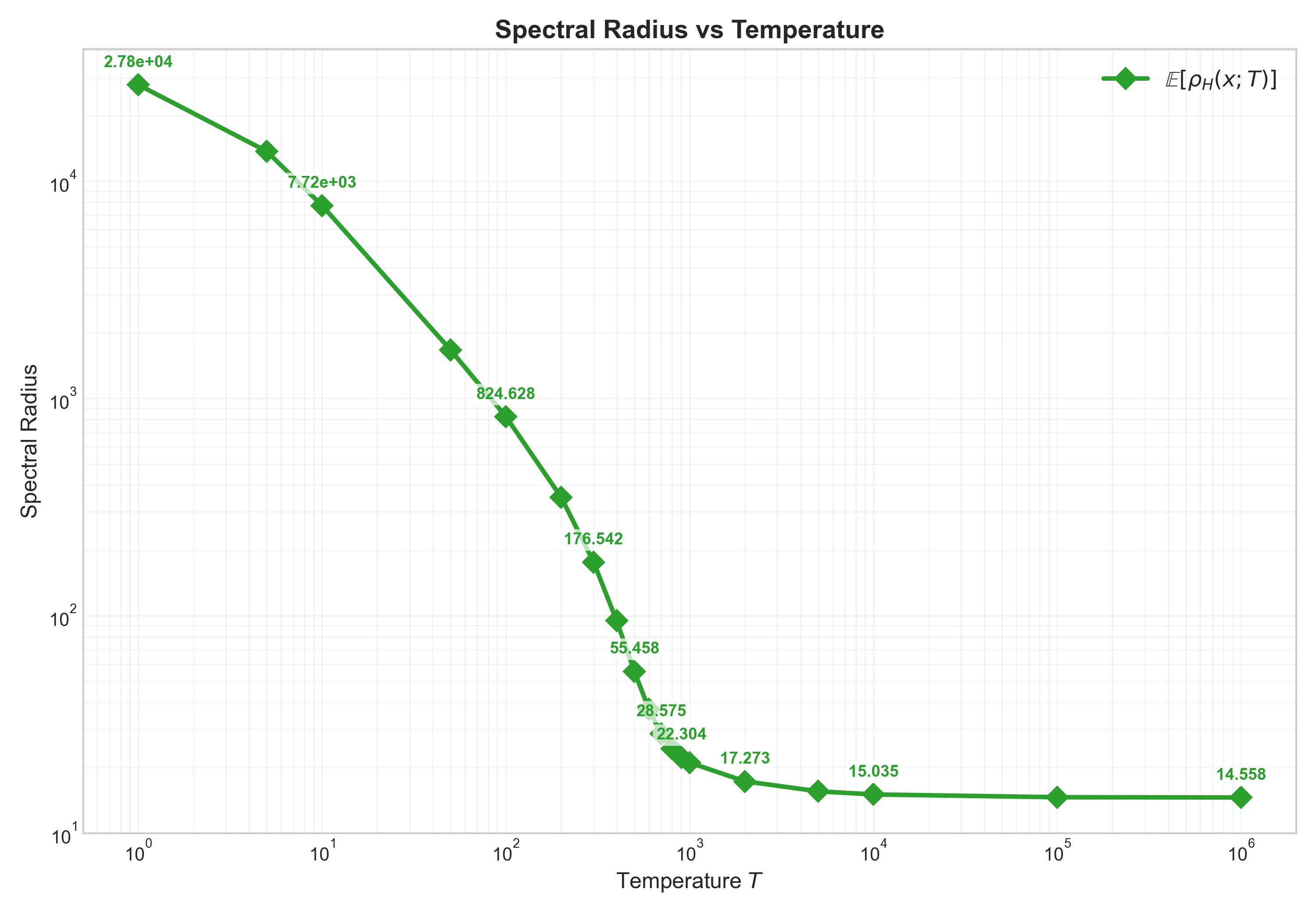}
\caption{Hessian spectral radius trend \(\mathbb E[\rho_H(x;T)]\) versus temperature \(T\).}
\label{fig:rho_fig3}
\end{figure}

\begin{figure}[!htb]
\centering
\includegraphics[width=0.6\linewidth]{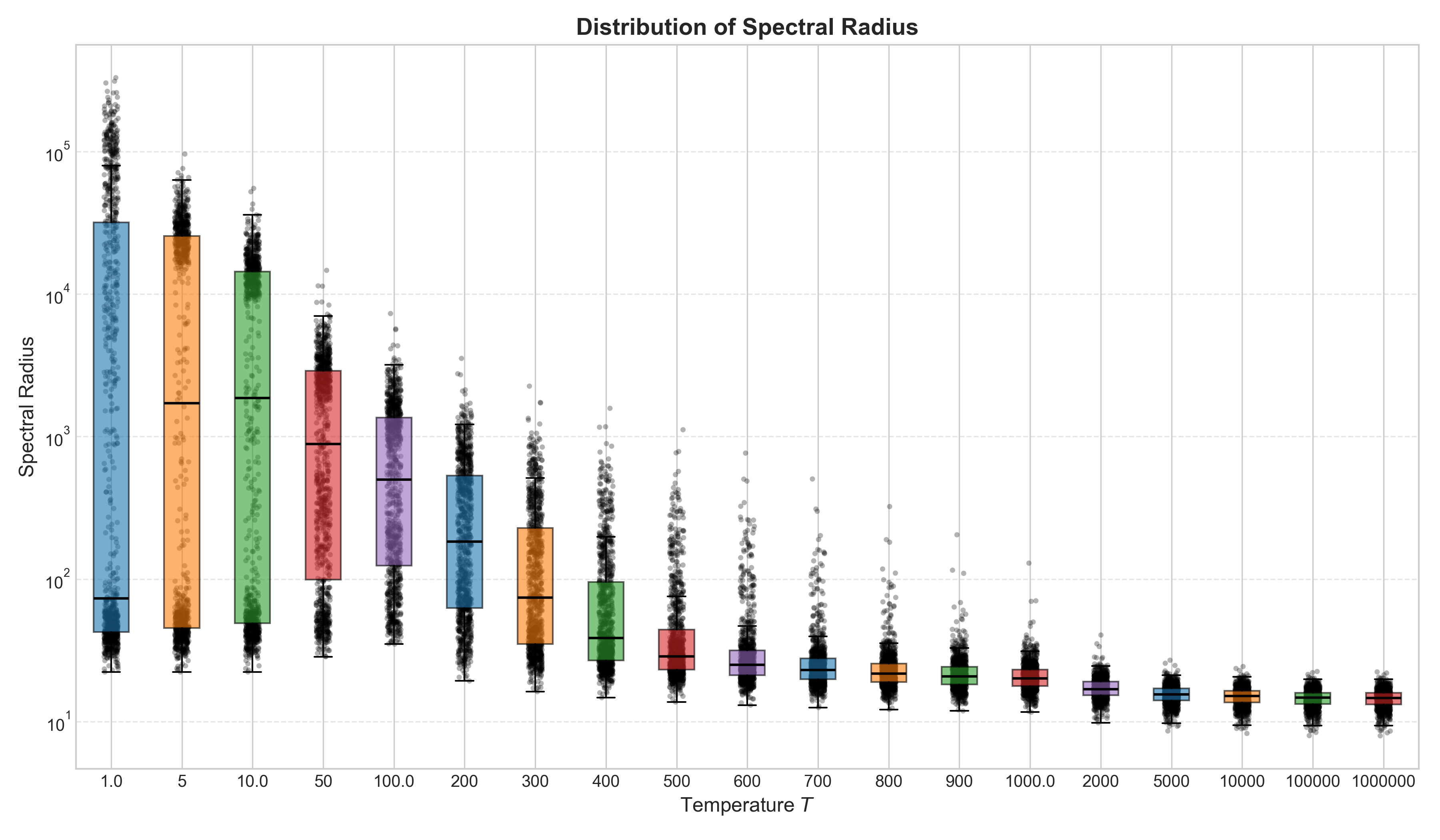}
\caption{Distribution of \(\rho_H(x;T)\) across sampled overlap points at different temperatures.}
\label{fig:rho_fig4}
\end{figure}

\paragraph{Results and consistency with theory.}
As summarized in Table~\ref{tab:rho_table} and Figs.~\ref{fig:rho_fig1}--\ref{fig:rho_fig4}, the empirical trends strongly match the theoretical predictions:

\begin{enumerate}
\item \textbf{The expansive term decays rapidly with temperature.}  
\(\mathbb E[\lambda_{\max}/T]\) decreases from \(2.6816\times10^4\) (\(T=1\)) to \(6.2\times10^{-3}\) (\(T=10^6\)); see Fig.~\ref{fig:rho_fig1}.  
Although raw \(\mathbb E[\lambda_{\max}]\) is not strictly monotone at low \(T\), the normalized expansive component shrinks sharply.

\item \textbf{The contractive baseline is non-increasing and remains positive.}  
\(\mathbb E[c_T]\) decreases from \(50.4309\) to \(14.6739\), consistent with \(\partial_T c_T(x)\le 0\), and remains bounded away from zero; see Table~\ref{tab:rho_table} and Fig.~\ref{fig:rho_fig1}.

\item \textbf{Sign flip of the dominant mode.}  
The crossover \(\lambda_{\max}/T \approx c_T\) occurs around \(T\in[700,800]\): at \(T=700\), \(\lambda_{\max}/T>c_T\), while at \(T=800\), \(\lambda_{\max}/T<c_T\); see Fig.~\ref{fig:rho_fig1}.  
Hence \(\mu_{\max}\) changes sign from positive to negative.

\item \textbf{Spectral contraction across the full distribution.}  
Both \(\mathbb E[\rho_H(x;T)]\) and its upper tail drop substantially with \(T\), indicating reduced curvature heterogeneity in overlap regions; see Figs.~\ref{fig:rho_fig3} and \ref{fig:rho_fig4}.

\item \textbf{Local expansiveness converges to \(1-\eta c_\infty\).}  
\(\mathbb E[\rho(x;T)]\) decreases from \(27.7656\) (\(T=1\)) to \(0.9853\) (\(T\ge 10^5\)); see Fig.~\ref{fig:rho_fig2}.  
This matches
\[
1-\eta c_\infty \approx 1-10^{-3}\times 14.6739 = 0.9853.
\]
\end{enumerate}

\paragraph{Conclusion.}
Increasing \(T\) suppresses the anisotropic expansive covariance term \(\frac{1}{T}\mathrm{Cov}\), while the isotropic negative baseline \(-c_T\mathbf I\) remains effective. Consequently, local dynamics move from strongly expansive at low \(T\) to near-nonexpansive/contractive at high \(T\), supporting improved stability and reduced memorization tendency.

\clearpage
\subsection{Extra Results on Cat-Caracal Dataset}
\subsubsection{Uncurated generated samples}
\begin{figure}[!htb]
    \centering
    \includegraphics[width=0.7\linewidth]{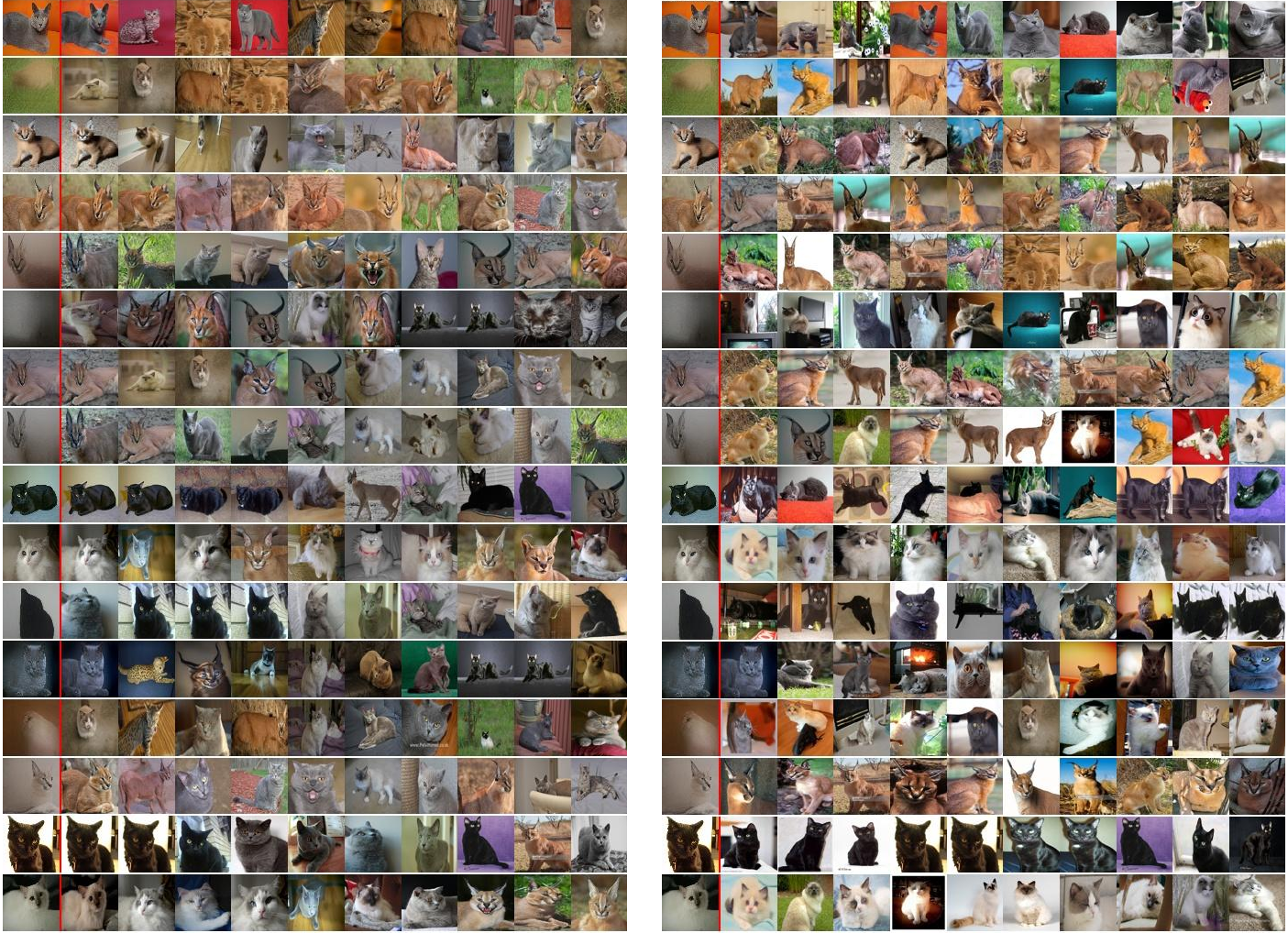}
    \caption{Uncurated samples of conditioning modeling and their closest training samples in pixel space (left) and feature space (right).}
    \label{fig4}
\end{figure}

\begin{figure}[!htb]
    \centering
    \includegraphics[width=0.7\linewidth]{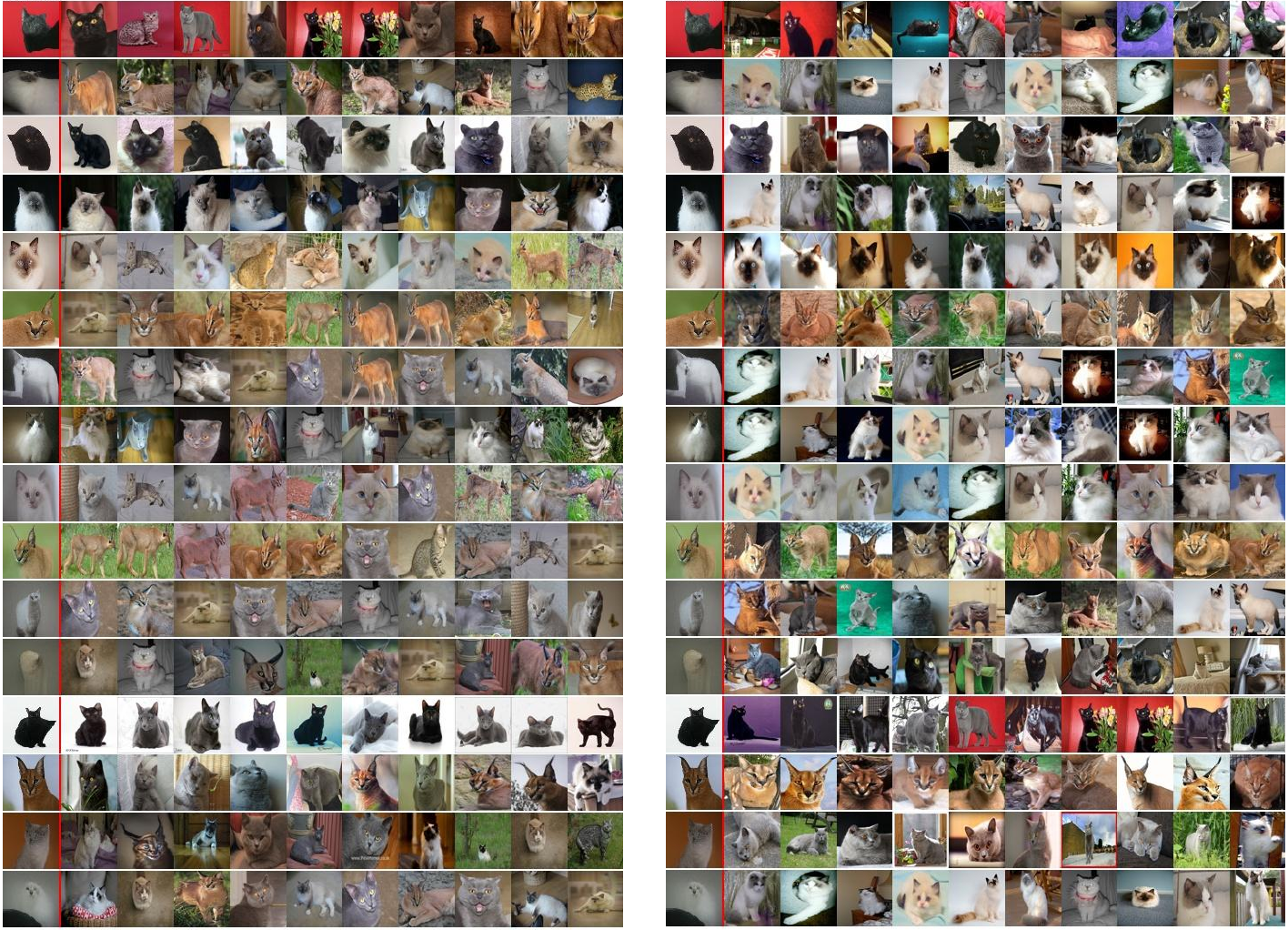}
    \caption{Uncurated samples of unconditioning modeling and their closest training samples in pixel space (left) and feature space (right).}
    \label{fig5}
\end{figure}

\begin{figure}[!htb]
    \centering
    \includegraphics[width=0.8\linewidth]{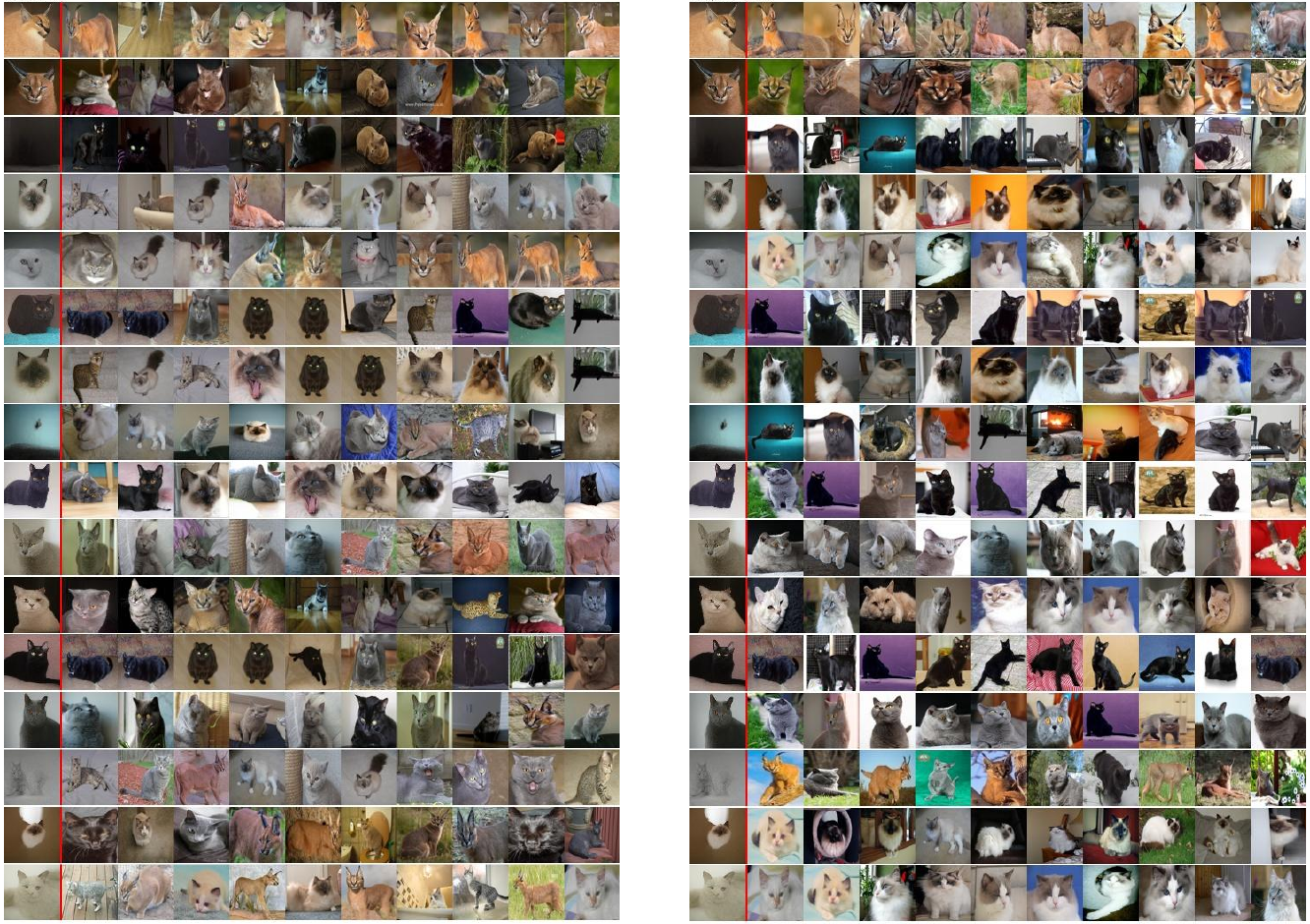}
    \caption{Uncurated samples of temperature-based ($T_i=7/\sigma_i$, K=10) modeling (feature space-based KNN) and their closest training samples in pixel space (left) and feature space (right).}
    \label{fig6}
\end{figure}

We provide additional uncurated samples on the Cat-Caracal dataset and their top-10 closest training samples in both pixel space and feature space (from a pre-trained Inception V3 network), measured by the $\ell_2$ distance. We can see serious memorization in the \cref{fig4} and certain generalization in \cref{fig5,fig6}. 

\subsubsection{Quantitative memorization fraction of different modelings}
\label{B.3.2}

To quantify how much different model variants memorize the Cat-Caracal training set, we perform a nearest-neighbor analysis in pixel space over synthesized samples.

\paragraph{Experimental setup.}
For each model variant, we generate 1000 images. Let \(x\) be a generated sample and \(\{y_i\}\) be all training images. We resize all images to \(64\times 64\), convert them to RGB tensors in \([0,1]\), and flatten them into vectors in \(\mathbb{R}^{3\cdot 64\cdot 64}\). In this pixel space, we use the Euclidean distance
\[
\operatorname{dist}(x,y_i) \;:=\; \|x - y_i\|_2
\]
to find the nearest training image \(y_{(1)}\) and the second nearest \(y_{(2)}\). Similarly to \cite{yoon2023diffusion}, for each generated sample we then define a \emph{memorization ratio}
\[
r(x) \;:=\; \frac{\operatorname{dist}(x, y_{(1)})}{\operatorname{dist}(x, y_{(2)})}.
\]
If a generated image is essentially a copy of one training example and clearly farther from all others, the nearest distance will be much smaller than the second nearest, so \(r(x)\) will be small. We mark a sample as a \emph{memorization case} if
\[
r(x) < \tfrac{2}{3},
\]
i.e., its nearest neighbor in pixel space is at least \(1.5\times\) closer than the second nearest one.

We conduct the experiments for conditioning, unconditioning, \((T_i{=}1/\sigma_i, K{=}30)\), \((T_i{=}1/\sigma_i, K{=}100)\), and \((T{=}7/\sigma_i, K{=}10)\). For each configuration we report: (i) the number of samples flagged as memorization (\#Mem); (ii) the mean ratio \(\mathbb{E}[r(x)]\); (iii) the minimum ratio \(\min_x r(x)\), indicating the strongest single-copy case observed.

\begin{center}
\begin{tabular}{lccc}
\toprule
Modeling & \#Mem (out of 1000) & Mean \(r(x)\) & Min \(r(x)\) \\
\midrule
Conditioning       & 128 & 0.8468 & 0.2276 \\
Unonditioning     & 92  & 0.8932 & 0.3569 \\
\(T_i{=}1/\sigma_i, K{=}30\) & 11  & 0.9424 & 0.5501 \\
\(T_i{=}1/\sigma_i, K{=}100\)& 4   & 0.9376 & 0.5342 \\
\(T_i{=}7/\sigma_i, K{=}10\) & 4   & 0.9465 & 0.6257 \\
\bottomrule
\end{tabular}
\end{center}

The standard conditioning diffusion shows clear memorization: 128 out of 1000 samples (12.8\%) have a nearest training image that is much closer than the second nearest one, with ratios as low as 0.23. In contrast, our proposed methods substantially reduce the memorization rate. The unconditioning already lowers the memorization count to 92 and increases both the mean and minimum ratios. All three temperature-based methods have fewer than 15 flagged samples out of 1000 and much larger minimum ratios (above 0.53), with mean ratios around \(0.94\), close to 1. This indicates that generated samples usually lie between multiple training images, rather than collapsing onto a single training example in pixel space.

\clearpage
\subsection{Extra Experiments on CIFAR-10}

\subsubsection{Collapse time of different empirical score functions}
\label{B.4.1}
As we mentioned in the introduction section, the collapse time means the sampling step at which a single center will dominate the whole score function weight, leading to sampling collapse and memorization. So in this section, we will provide the experimental results, sampling with different empirical score functions to verify our claim that conditioning will collapse before unconditioning, and increasing the temperature can further delay the collapse time.

\begin{figure}[!htb]
    \centering
    \includegraphics[width=0.65\linewidth]{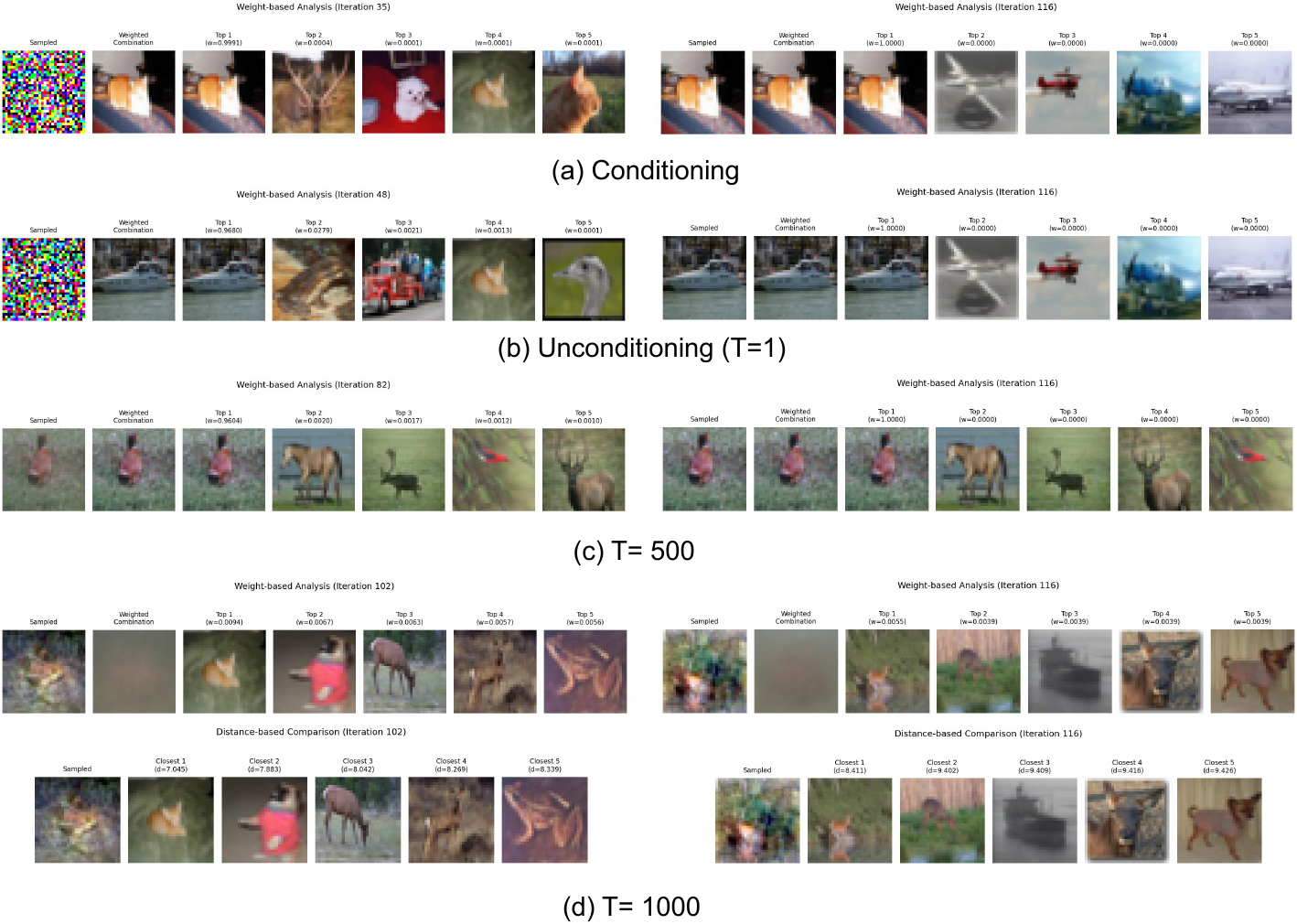}
    \caption{Illustration of sampling with different empirical score functions. The left part of (a, b, c) is the intermediate sample and their top-5 score function weight training samples at the ``collapse time" (the first time that the biggest score function weight is bigger than 0.95); The right part is the last sampling step. In (d), since the temperature is very high, there is no collapse time, so we select a meaningful intermediate sample and the generated sample at the last sampling step.}
    \label{fig7}
\end{figure}

\noindent
We conduct experiments using 10k images in the training set and calculate their explicit score function. Then, we use the ODE sampling, and the ``shell-touch" noise scheduling (116 noise levels of the log-uniform distribution between 0.01 and 50) from \cref{A.3}: 

\[
  \sigma_i\sqrt{d} + \frac{3\sigma_i}{\sqrt{2}}
  = \sigma_{i+1}\sqrt{d} - \frac{3\sigma_{i+1}}{\sqrt{2}}.
\]

Note that the ODE sampling here is based on the actual noise level so that it will not lead to sampling overshoot even in late sampling stages.

As shown in \cref{fig7}, conditioning modeling exhibits collapse at the 35th iteration when the sampling point remains highly noisy, while unconditioning and T=500 significantly delay this collapse. Notably, at T=500, the generated sample approaches the image manifold even before the score function weight becomes dominant. More surprisingly, when we further increase the temperature to T=1000 in \cref{fig7}(d), the score function weight never dominates during the sampling process. Instead, the sampling trajectory continues exploring the image manifold, producing diverse samples such as the frog at the 102nd iteration, without eventual collapse. However, this exploration exhibits inherent instability: the stopping criterion becomes unpredictable, and many intermediate or final samples may lack semantic meaning. This instability fundamentally motivates our choice of neural network-based score function learning, which provides stable and reliable score estimation throughout the sampling process. Beyond the settings reported here, we have also experimented with a wide range of temperature schedules (e.g., gradually increasing $T$ over time) and different numbers of training samples for constructing the empirical score. As the resulting behaviors are highly consistent yet overly numerous and redundant, we do not present them all here. Overall, we observe that increasing the temperature drives the sampling trajectory first into (and within) the convex hull of a small neighborhood of training samples, and then along directions approximately tangent to the data manifold, until it either collapses to a single example or converges to a stationary point, as illustrated in \cref{fig1}.

\medskip
\subsubsection{Qualitative comparison for different types of KNN}
\label{B.4.2}

We also provide the visualization pairs between pixel space-based KNN and feature space-based KNN as the following:
\begin{figure}[!htb]
    \centering
    \includegraphics[width=0.8\linewidth]{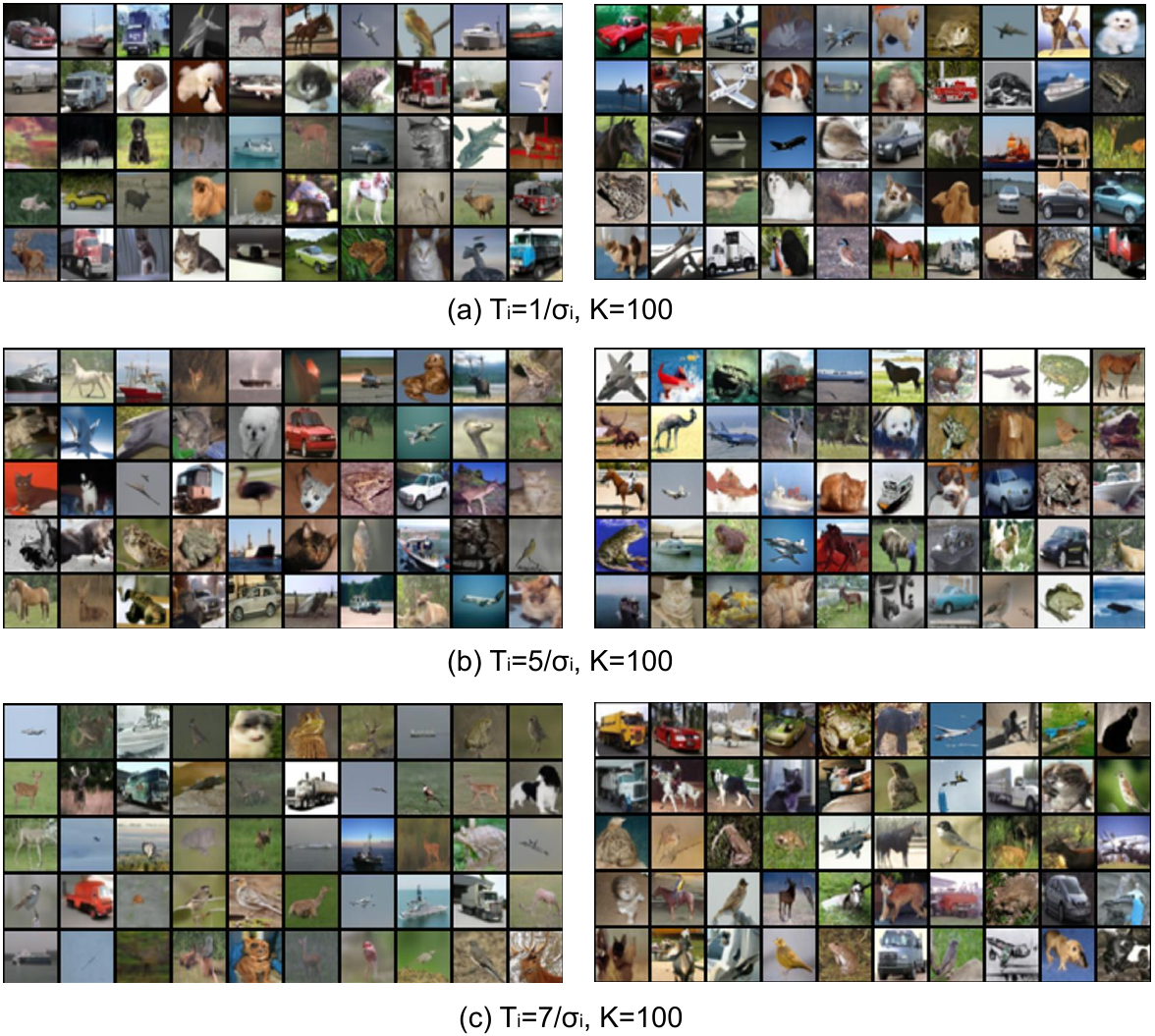}
    \caption{Comparison of generated samples using pixel space-based (left) and feature space-based (right) KNN across increasing temperatures. Higher temperatures lead to severe quality degradation in pixel space while feature space maintains high quality.}
    \label{fig8}
\end{figure}

\noindent
As illustrated in \cref{fig8}, the visual comparison clearly demonstrates the superiority of feature space-based KNN over pixel space-based approaches under varying temperature settings. When the temperature increases from $T_i=1/\sigma_i$ to $T_i=7/\sigma_i$, pixel space-based KNN modeling exhibits severe quality degradation with blurred, distorted, and semantically inconsistent samples due to over-smoothing that deviates from the image manifold. But feature space-based modeling maintains consistently high generation quality while demonstrating remarkable feature fusion and generalization capabilities, e.g., illustrated by the ``frog car", ``cat dog", ``blue horse", and so on, which will not appear in the conditioning modeling:

\begin{figure}[!htb]
    \centering
    \includegraphics[width=0.3\linewidth]{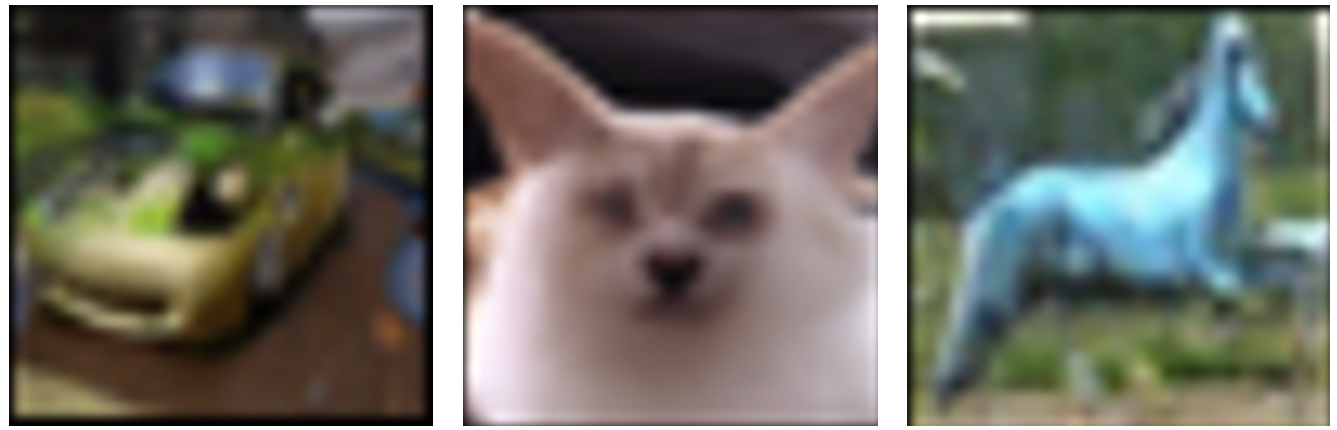}
    \caption{Examples from temperature-based modeling, showing creative feature fusion and generalization capabilities.}
    \label{fig9}
\end{figure}

\clearpage
\subsubsection{ODE sampling with the true noise level $\sigma_{n*}$}
\label{B.4.3}

\begin{figure}[!htb]
    \centering
    \includegraphics[width=0.85\linewidth]{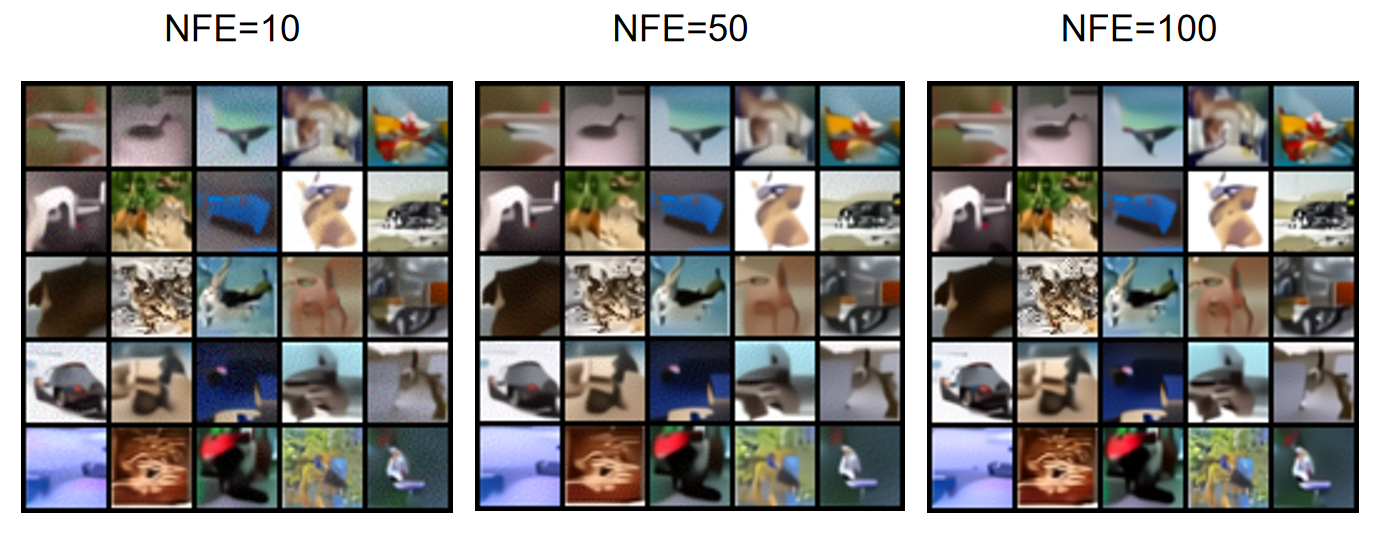}
    \caption{Under the same initial noise, the results from different sampling steps via ODE sampling with true noise level $\sigma_{n*}$.}
    \label{fig10}
\end{figure}

To support our claim that our unconditioning modeling enables larger sampling steps, we provide the generated samples via NCSNv2 \cite{song2020improved} as \cref{fig10} with the same initial noise for 10, 50, 100 NFE, respectively. The sampling iteration is:
$$
x_{n+1} = x_n + \alpha\sigma_{n*}^{2}s_\theta(x_n)
$$
\noindent
where the $\alpha$ is 0.6, 0.16, and 0.09, respectively. We can see the generated images are basically consistent, showing the potential to be learned by an ODE solver, namely training a Consistency Model \cite{song2023consistency}. As for reducing the cost of calculating the closest distance to determine the $\sigma_{n*}$, we can just calculate the $\sigma_{n*}$ at late sampling stages. At early sampling stages, we can still use the prefixed noise schedule $\sigma_n$.   

Note that for both the NN learned and empirical unconditioning score functions, we can fix a small step size and keep doing sampling as follows: 
$$
x_{n+1} = x_n + \alpha s_\theta(x_n).
$$

\noindent
We find that stopping when the $\sigma_{n*}$ is small can also lead to high quality samples --- a standard gradient ascent process.

\subsubsection{Other choices of temperature}
Besides the temperature in the reported experiments, we also tried other temperature values and other types.

\medskip
\noindent
\textbf{Other temperature values.}
We also tried $T_i=1/\sigma_i$ to $T_i=8/\sigma_i$, where we cannot see difference between $T_i=(1,5)/\sigma_i$ and $T_i = (2,3,4,6)/\sigma_i$ based on pixel space KNN. However, starting from the $T_i=7/\sigma_i$, we can see obvious smoothing in generated images. Especially, when we further increase the temperature or the K, the generation quality will significantly decrease shown as the \cref{fig11}

\begin{figure}[!htb]
    \centering
    \includegraphics[width=0.9\linewidth]{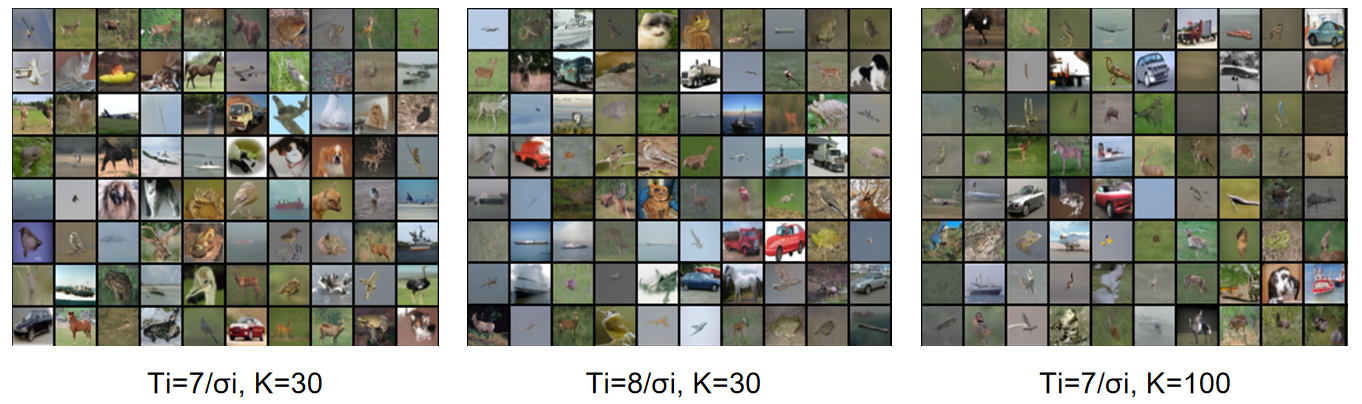}
    \caption{Illustration of keeping increasing temperature or K (pixel space).}
    \label{fig11}
\end{figure}

\textbf{Other temperature types.} Except setting $T \propto 1/\sigma_i$, we also tried other types like $T \propto 1/\sigma_i^2$ and $T=\sqrt{d}$ (inspired by the conclusion in the analysis of the $\mu$ domination that $\ln \frac{w_j^*}{w_l^*}\propto\sqrt{d}$, which means that setting $T=\sqrt{d}$ can cancel the influence of dimensionality). However, both of them will produce over-smoothed samples:

\begin{figure}[!htb]
    \centering
    \includegraphics[width=0.8\linewidth]{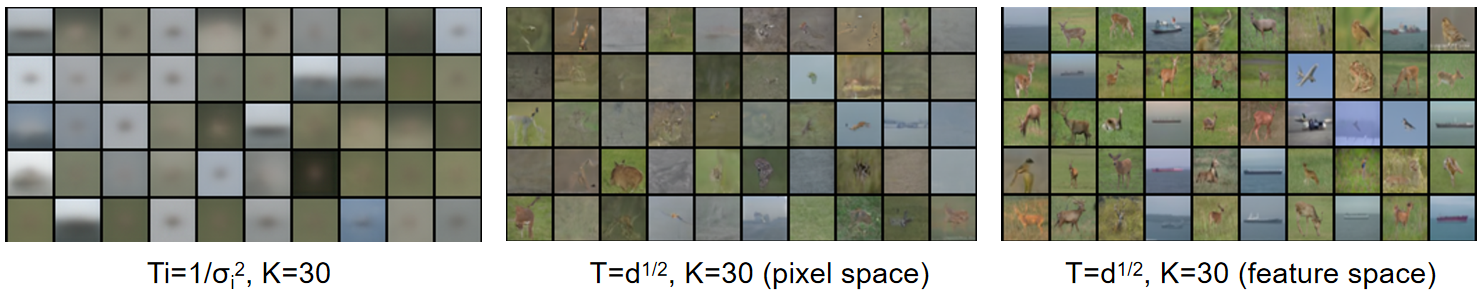}
    \caption{Other types of temperature.}
    \label{fig11}
\end{figure}

\noindent
\textbf{Other KNN types.} Except for setting a hard K for the empirical score function approximation, we also tried a natural approximation based on the neighbor region. Specifically, we first calculate the distance to the second nearest neighbor $d_{sc}$ in the training set as a reference, with a threshold $R=rd_{sc}$, then using all samples in this neighbor region for score function approximation, where $r$ is the coefficient to control the size of the neighbor region. This algorithm is easy to implement by first choosing a hard K and doing a mask depending on the threshold R for later empirical score function approximation. Although we cannot see an obvious difference between this choice and the hard K approximation, it is better from the theoretical perspective since this setting is related to its local manifold curvature and should be more robust.

\subsubsection{Uncurated generated samples and memorization pairs}

We also provide more uncurated samples:
\begin{figure}[!htb]
    \centering
    \includegraphics[width=0.7\linewidth]{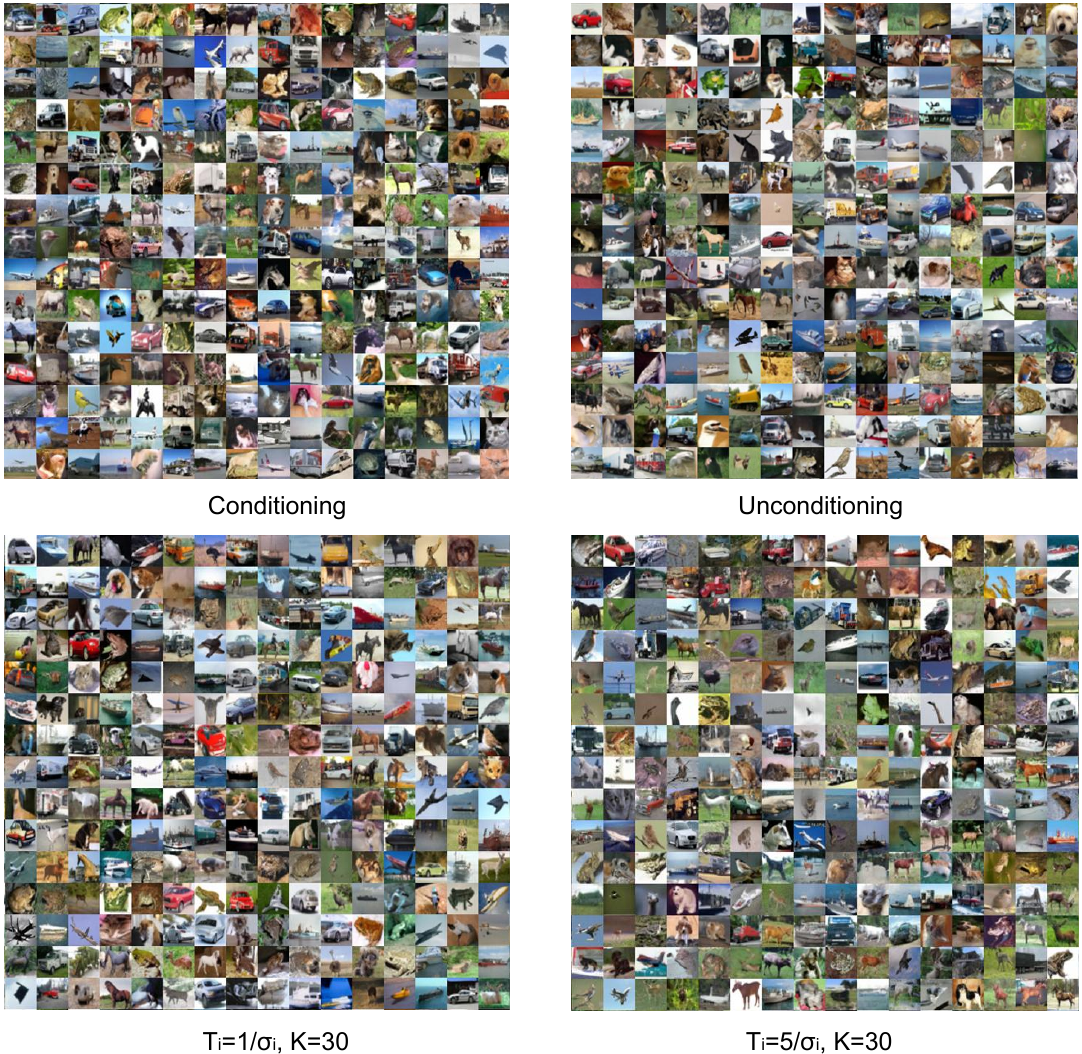}
    \caption{Uncurated results of different modeling on CIFAR-10.}
    \label{fig12}
\end{figure}

\paragraph{Extracting memorization pairs on CIFAR-10.} We first applied the ratio defined in \cref{B.3.2}, but the values for different modelings were indistinguishable. Then we tried other quantitative metrics such as top-$K$ entropy, which similarly failed to separate the methods, though we can clearly see better generalization with temperature-based methods. Therefore, to systematically study training data memorization on CIFAR-10, we therefore adopt a two-stage ``generate-and-filter'' pipeline following \cite{carlini2023extracting}, but with a smaller sampling budget and modified scoring rules. Concretely, we first generate 100k images using each modeling method, and then search for training images that are abnormally close to these samples in both pixel space and feature space.

\medskip
\noindent
\textbf{Stage 1: pixel-space nearest neighbors.}
Let $\{x_i\}_{i=1}^N$ denote the $N=50{,}000$ training images, and let $\{g_m\}_{m=1}^M$ denote the $M= 10^{5}$ generated images. We flatten each image to a vector in $\mathbb{R}^d$ with $d=3072$, and define the normalized $\ell_2$ distance
\[
d_{\text{pix}}(u,v)
\;=\;
\frac{\lVert u-v\rVert_2}{\sqrt{d}}.
\]
For every generated sample $g_m$, we compute its distances $\{d_{\text{pix}}(g_m,x_i)\}_{i=1}^N$ to all training images and retrieve the top-$k$ nearest neighbors (we use $k=50$). Denote the corresponding distances by $d_1 \le d_2 \le \dots \le d_k$. To detect whether the closest training image
is ``relatively too close'' compared to other candidates, we define the \emph{relative neighbor ratio}
\[
r_m
\;=\;
\frac{d_1}{\frac{1}{k-1}\sum_{j=2}^k d_j}.
\]
Intuitively, $r_m$ is small when the nearest training image is much closer than the remaining $k-1$ neighbors, suggesting a potentially memorized match rather than a generic sample from a dense region (e.g., many similar skies or grass textures). We compute $r_m$ for all generated samples and select as \emph{Stage~1 candidates} those with $r_m$ below a small left-tail percentile of the $\{r_m\}_{m=1}^M$ distribution. E.g., the $1\%$ percentile, which will include more potential memorization samples compared with the 0.5\% in \cite{carlini2023extracting}.

\medskip
\noindent
\textbf{Stage 2: feature-space consistency check.}
Pixel-level similarity alone is insufficient to reliably identify memorization on CIFAR-10: many non-memorized samples can also have small pixel distances to multiple training images. To obtain a stronger notion of ``abnormally close'', we enforce an additional constraint in a semantic feature space. We use the InceptionV3 (with 2048-dimensional pool3 features) as a feature extractor $f(\cdot)$.
First, we extract features $f(x_i)\in\mathbb{R}^{2048}$ for all training images and, for each $x_i$, compute its nearest-neighbor distance among the other training points:
\[
d^{\text{train}}_f(i)
\;=\;
\min_{j\neq i} \bigl\lVert f(x_i) - f(x_j)\bigr\rVert_2.
\]
This quantity characterizes how isolated $x_i$ is on the real data manifold. For each Stage~1 candidate $g_m$, we take its pixel-space nearest training neighbor
$x_{\text{nn}(m)}$ (the image achieving $d_1$ above) and compute the feature-space distance
\[
d_f(g_m, x_{\text{nn}(m)})
\;=\;
\bigl\lVert f(g_m) - f(x_{\text{nn}(m)})\bigr\rVert_2.
\]
We then retain $(g_m, x_{\text{nn}(m)})$ as an \emph{automatic memorization candidate} only if
\[
d_f(g_m, x_{\text{nn}(m)}) \;<\; \beta \cdot d^{\text{train}}_f\bigl(\text{nn}(m)\bigr),
\]
with a fixed margin parameter $\beta$ (we use $\beta=2.0$).

\medskip
\noindent
\textbf{Manual verification.}
The above two-stage procedure yields a set of high-scoring generated–training pairs, but some of them are only semantically similar rather than true memorization (e.g., different instances of the same class under similar backgrounds). We therefore add a final manual screening step: for each candidate, we show a side-by-side grid of \((g_m, x_{\text{nn}(m)})\), and a human annotator checks whether the two images are nearly identical in fine details. We apply this to four models: conditioning, unconditioning, \((T_i=1/\sigma_i, K=10)\), and \((T_i=1/\sigma_i, K=30)\), obtaining 514, 447, 530, and 426 memorization pairs, respectively. As illustrated in the following figures, three types of images
appear repeatedly:

\begin{figure}[!htb]
    \centering
    \includegraphics[width=0.15\linewidth]{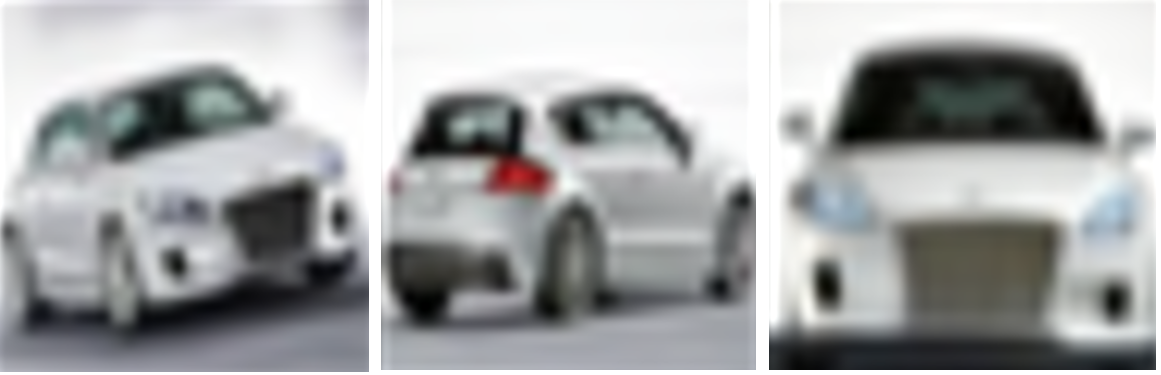}
\end{figure}

\noindent
As expected, we found that they also appeared repeatedly in the training set, especially in the first two images, which appeared more than 50 times. This means that our method may naturally fail since our methods are based on local manifold smoothing. We also count the number of memorization pairs without these three images, which are 324, 264, 238, and 172, respectively, demonstrating the significant ability of our method to alleviate memorization. However, memorizing the repeated training samples is an inevitable drawback in diffusion modeling theory, so we strongly recommend that both standard diffusion models and our unconditioning modeling be deduplicated in pixel space before training. In the following, we show all the memorization pairs with these four modeling methods as mentioned above.

\begin{figure}[!htb]
    \centering
    \includegraphics[width=0.85\linewidth]{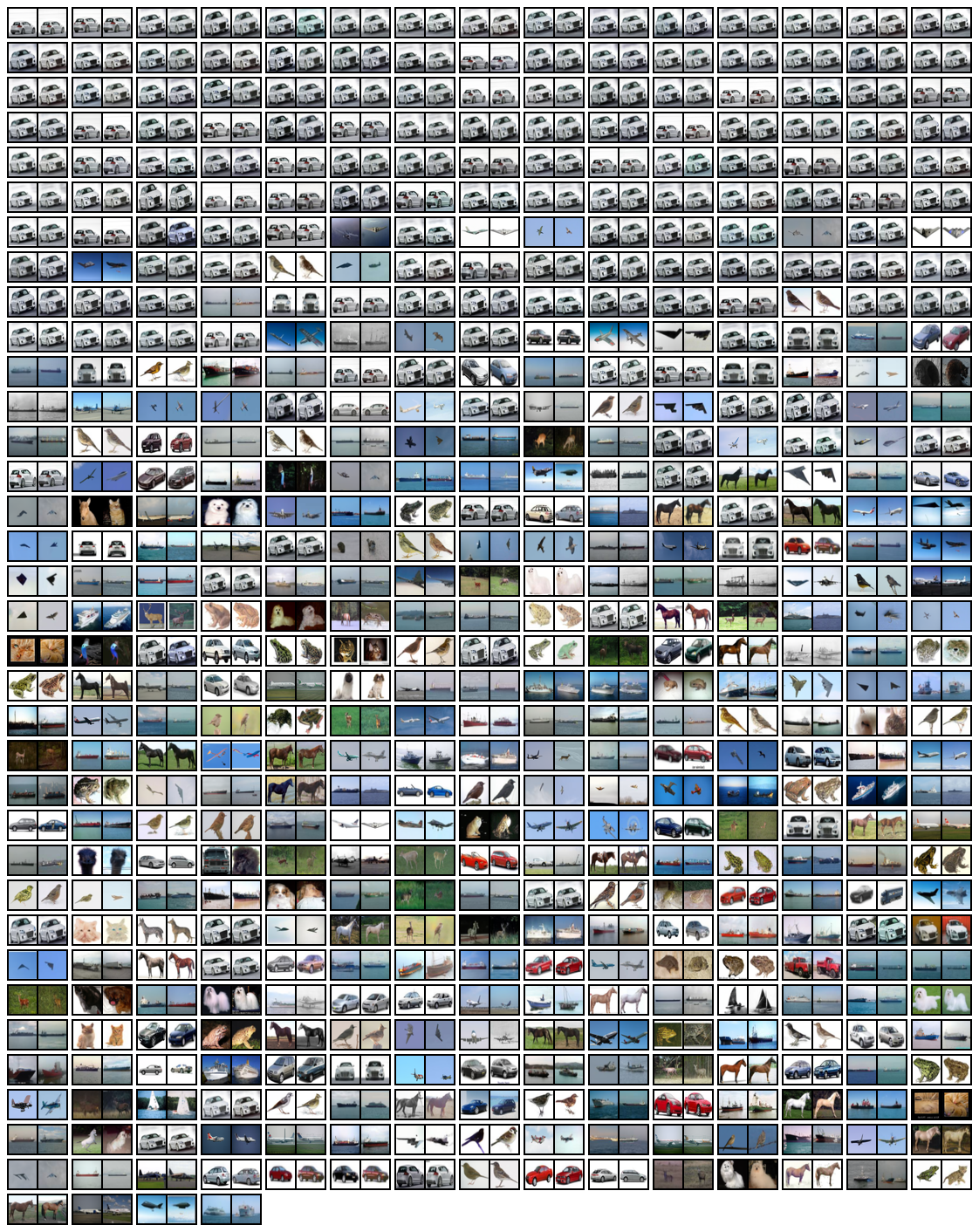}
    \caption{The memorization pairs of the conditioning modeling. (before/after removing 3 cars: 512/324)}
    \label{memo_con}
\end{figure}

\begin{figure}[!htb]
    \centering
    \includegraphics[width=0.9\linewidth]{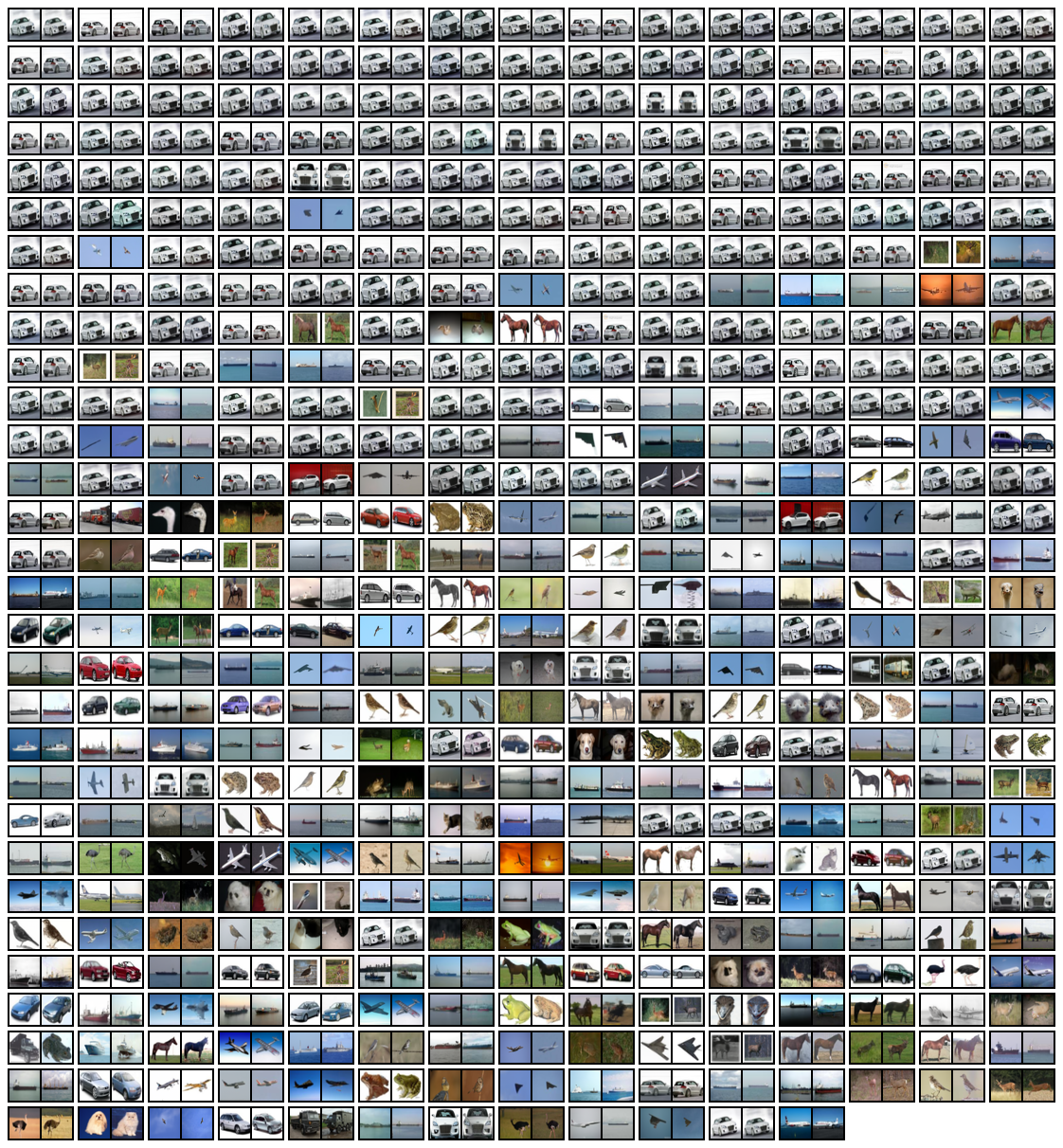}
    \caption{The memorization pairs of the unconditioning modeling. (before/after removing 3 cars: 447/264)}
    \label{memo_uncon}
\end{figure}

\begin{figure}
    \centering
    \includegraphics[width=0.9\linewidth]{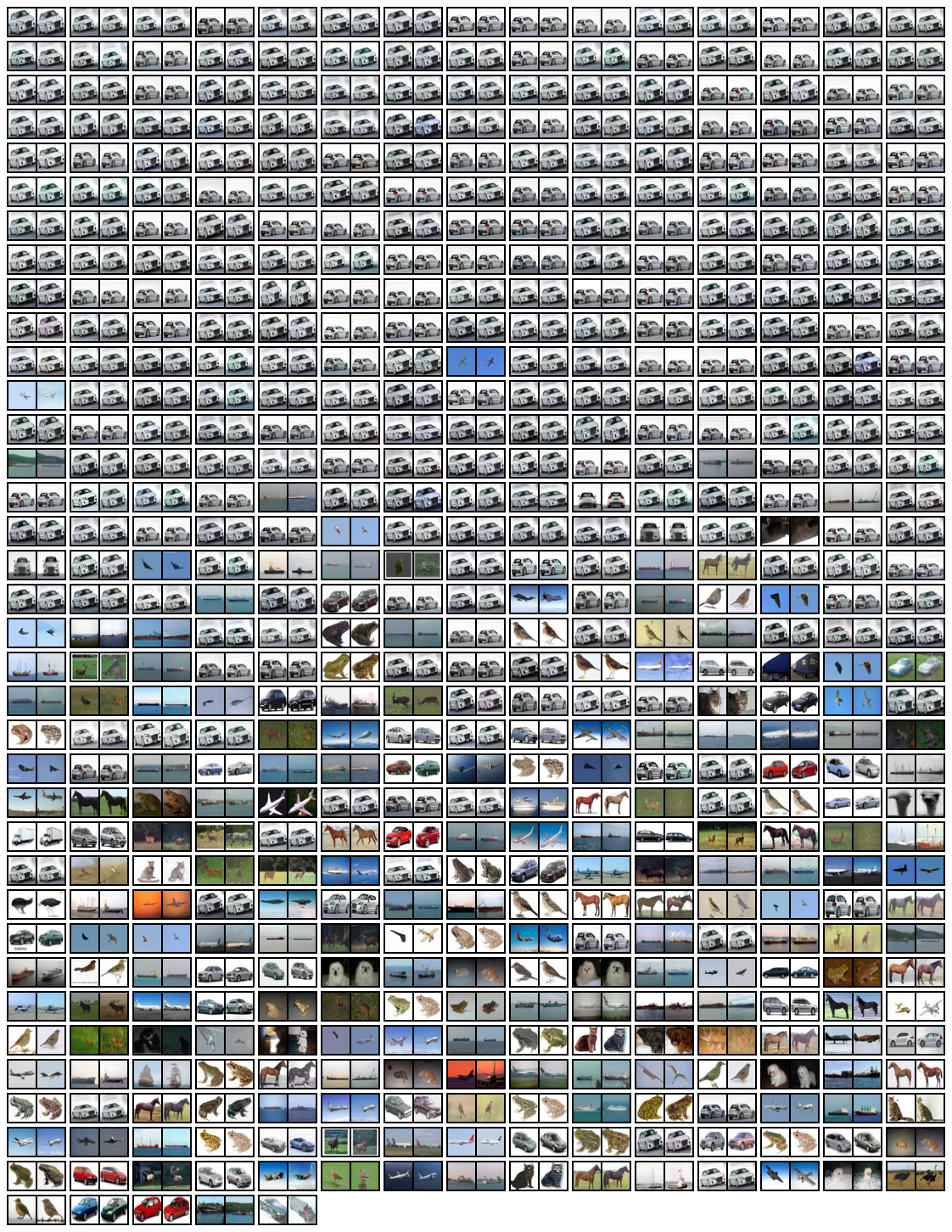}
    \caption{The memorization pairs of the unconditioning modeling with $T_i=1/\sigma_i, K=10$. (before/after removing 3 cars: 530/238)}
    \label{memo_uncon_k=10}
\end{figure}

\begin{figure}
    \centering
    \includegraphics[width=0.9\linewidth]{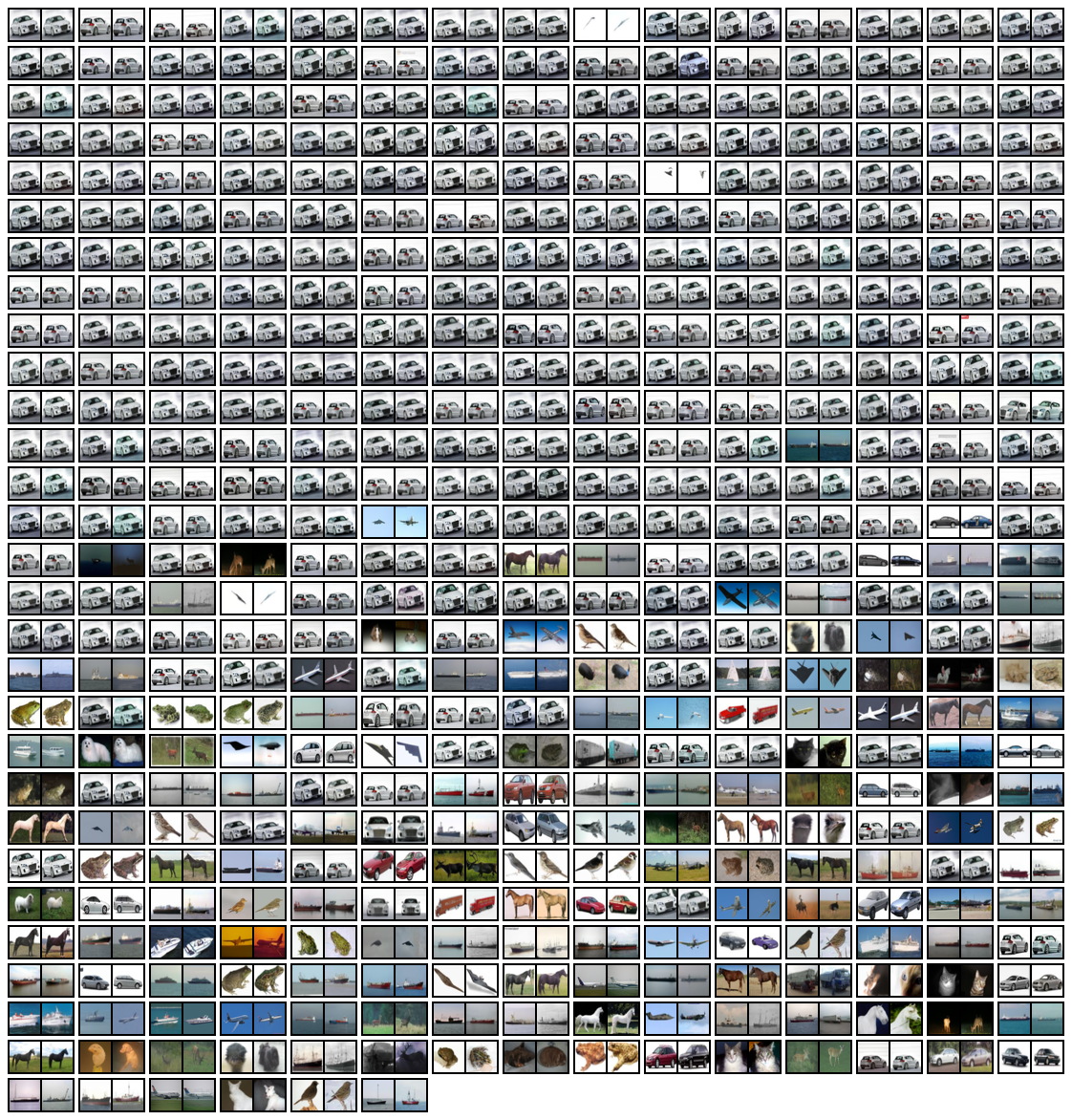}
    \caption{The memorization pairs of the unconditioning modeling with $T_i=1/\sigma_i, K=30$. (before/after removing 3 cars: 426/172)}
    \label{memo_uncon_k=30}
\end{figure}

\clearpage
\subsection{Extra Experiments on CelebA}

\subsubsection{Ablation studies of different temperature and K}

We also conduct ablations on the CelebA $64\times64$ with ODE sampling. We conduct the experiments for conditioning, unconditioning, $(T_i=5/\sigma_i, K=30),(T_i=5/\sigma_i, K=100),(T_i=10/\sigma_i, K=30),(T_i=10/\sigma_i, K=100).$

\begin{table}[!htbp]
\centering
\caption{Performance of different temperature and K setting.}
\label{tab2}
\begin{tabular}{c|c|c}
\hline
\multirow{2}{*}{\makecell{\textbf{Method}\\ \textbf{(ODE (PC) 1K NFE)}}} & 
\multirow{2}{*}{\makecell{\textbf{FID}$(G, \mathcal{T}_{\text{train}})$}} & 
\multirow{2}{*}{\makecell{\textbf{FID}$(G, \mathcal{T}_{\text{test}})$}} \\
& & \\
\hline
Conditioning  &  7.65  & 7.87  \\
Unconditioning &  7.71 & 7.88 \\
$T_i=5/\sigma_i$, K=30 & 9.36 $\to$ 8.67 & 9.50 $\to$ 8.80  \\
$T_i=5/\sigma_i$ K=100 &  36.81 $\to$  10.26  &  36.25  $\to$  10.41  \\
$T_i=10/\sigma_i$ K=30 & 12.07 $\to$ 10.62 &  11.21  $\to$ 10.05   \\
$T_i=10/\sigma_i$ K=100 &  63.11  $\to$  10.96   &  62.56   $\to$ 10.83  \\
\hline
\end{tabular}
\end{table}

The CelebA results demonstrate different parameter sensitivity compared to CIFAR-10. While CIFAR-10 shows severe degradation at high temperatures, CelebA exhibits better temperature robustness with more gradual performance decline. However, it is more sensitive to the K parameter (in the pixel space): increasing K from 30 to 100 causes significant FID deterioration (e.g., from 12.07 to 63.11 at $T_i=10/\sigma_i$). This suggests that excessive K values cause the model to learn an ``average face" by incorporating too many diverse neighbors, leading to blurred and generic facial features that degrade generation quality. As expected, we find that implementing the feature space-based KNN can also effectively reduce the oversmoothing in the CelebA as well as in the CIFAR-10.

\subsubsection{Uncurated generated samples and their closest training samples}
We also provide more uncurated samples on CelebA:

\begin{figure}[!htb]
    \centering
    \includegraphics[width=1.0\linewidth]{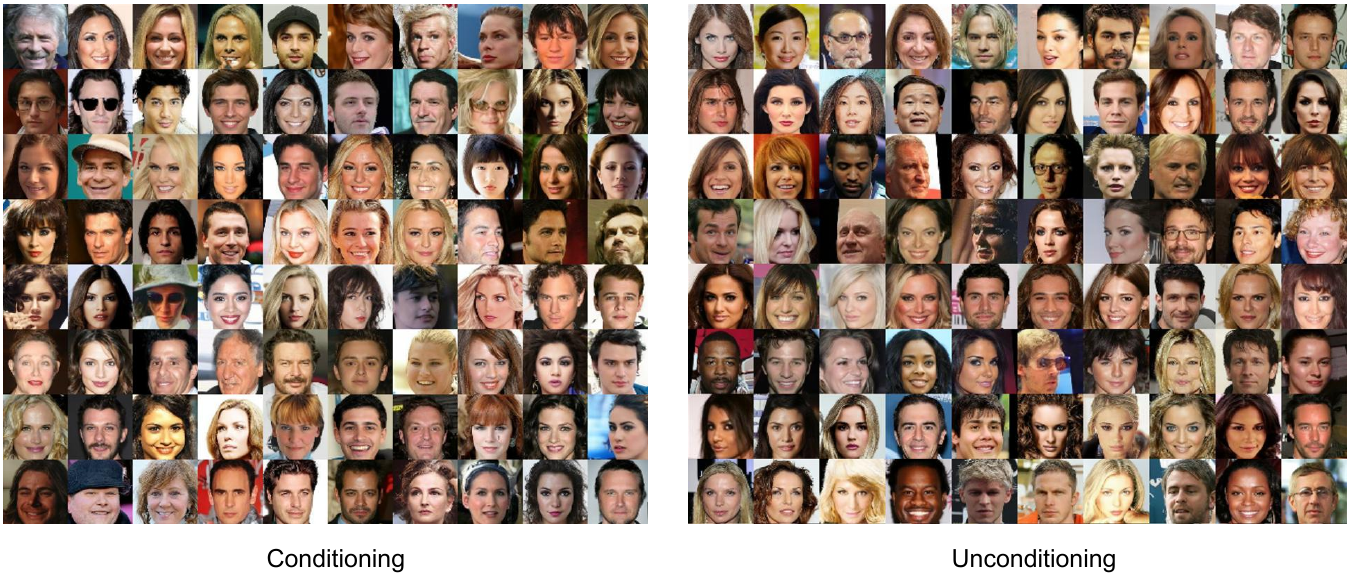}
    \caption{Uncurated results of conditioning and unconditioning modeling on CelebA.}
    \label{fig14}
\end{figure}

\begin{figure}[!htb]
    \centering
    \includegraphics[width=1.0\linewidth]{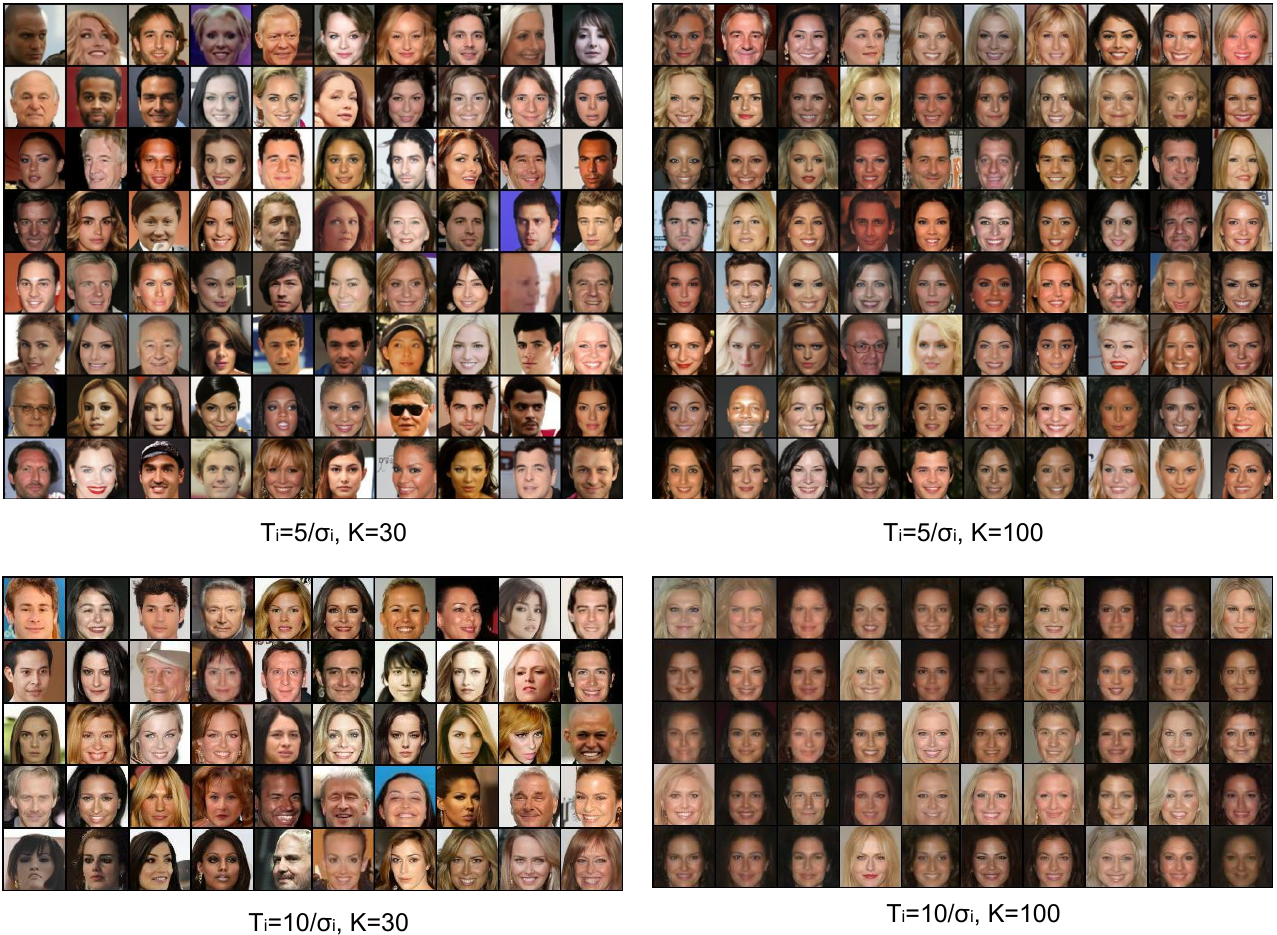}
    \caption{Uncurated results of different temperature and K (pixel space) on CelebA.}
    \label{fig15}
\end{figure}

\begin{figure}[!htb]
    \centering
    \includegraphics[width=1.0\linewidth]{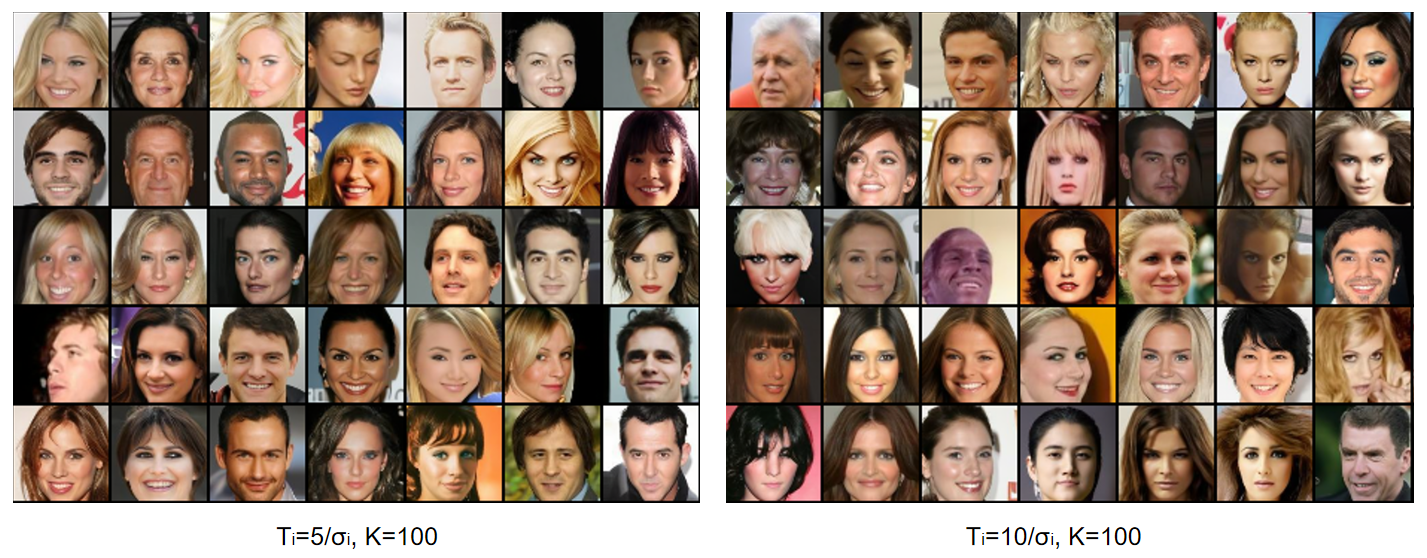}
    \caption{Uncurated results via feature space based KNN on CelebA.}
    \label{celebafeature}
\end{figure}

\begin{figure}
    \centering
    \includegraphics[width=1.0\linewidth]{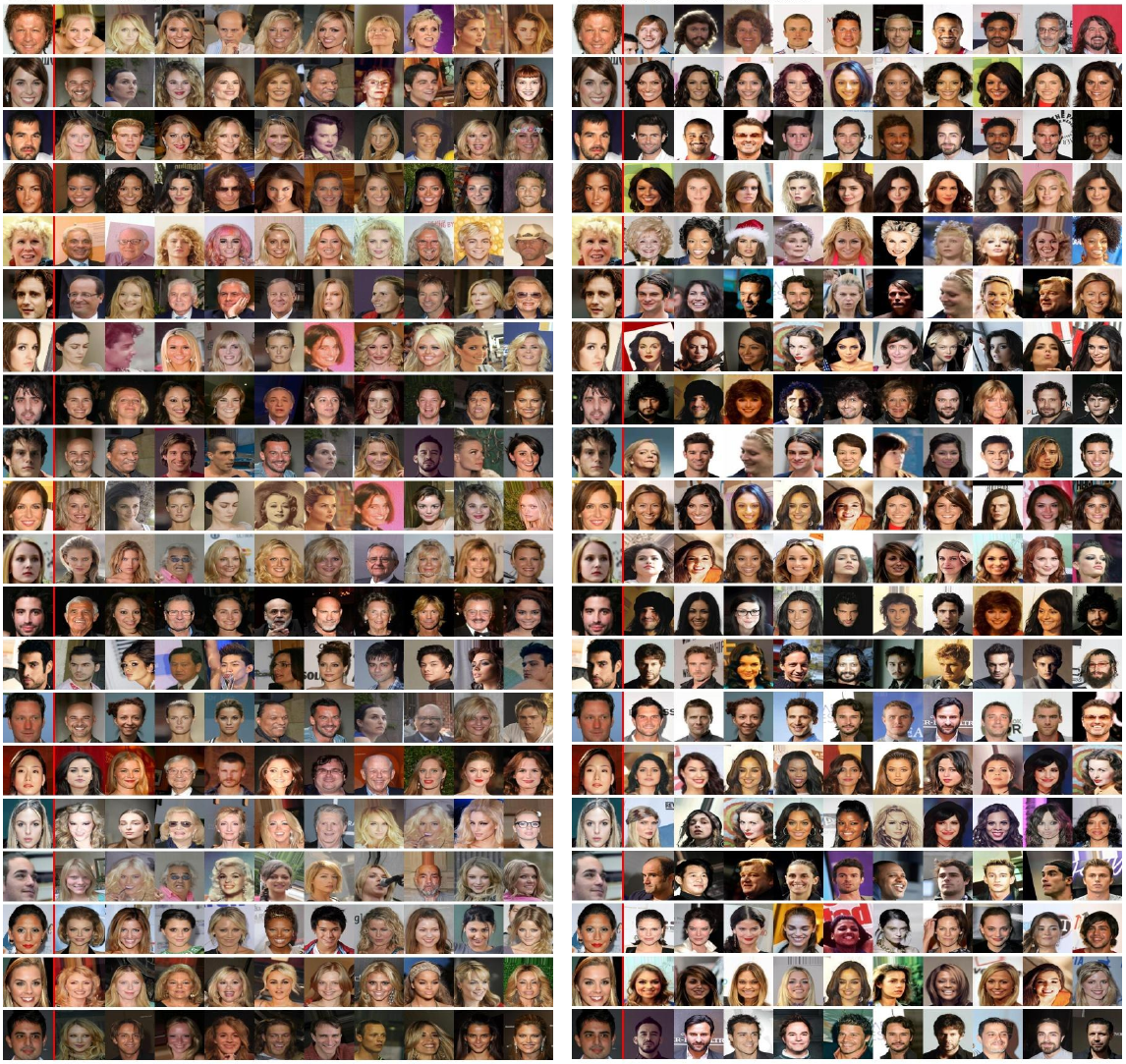}
    \caption{Uncurated samples of unconditioning modeling and their closest training samples in pixel space (left) and feature space (right).}
    \label{fig16}
\end{figure}

\cref{fig15} shows that when we let K=100, the NN indeed tends to learn a smoothed ``average face'', which aligns with our prediction. But when we use the feature space based KNN like on cifar10, we can get more reasonable results under the same K and T settings \cref{celebafeature}. Meanwhile, we provide the top-10 nearest training samples for unconditioning modeling in both pixel space and feature space in \cref{fig16}.

\clearpage
\subsection{Supplementary Experiments on ImageNet and CelebA-HQ}
We further conduct supplementary experiments on ImageNet $64\times64$ and CelebA-HQ $256\times256$ to demonstrate that our methods can be extended to large-scale and/or high-resolution datasets.

We do not report quantitative FID scores in this section. Since both our model and the baseline of \citet{song2020score} are class-\emph{unconditional}, FID on a highly diverse dataset such as ImageNet is known to be dominated by the presence or absence of class-conditioning \cite{dhariwal2021diffusion}, making a direct FID comparison uninformative. We therefore focus on a direct visual comparison between conditioning and unconditioning models.

While these qualitative results are consistent with our theoretical findings, we view incorporating class-conditioning and scaling to even higher resolutions (e.g., ImageNet $512\times512$) as important directions for future work.

\subsubsection{Uncurated Generated Samples on ImageNet $64\times64$}
\begin{figure}[!htb]
    \centering
    \includegraphics[width=1.0\linewidth]{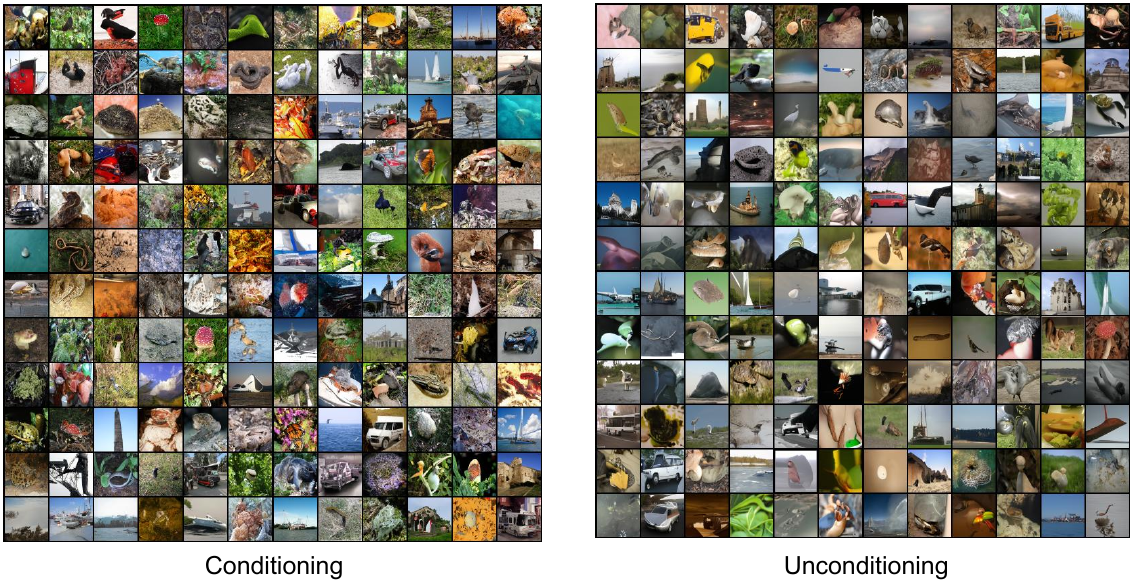}
    \caption{Uncurated samples comparing conditioning (left) and unconditioning (right) modelings on ImageNet $64\times64$.}
    \label{fig17}
\end{figure}

As shown in Figure~\ref{fig17}, samples from the unconditioned model appear smoother and better capture the global structure of objects compared to those from the conditioning model. The samples from unconditioned model exhibit fewer fine-grained details but preserve the core structural features, which may reduce the likelihood of memorizing training examples. This observation aligns with our theoretical framework, which suggests that unconditioning encourages the model to learn more generalizable representations.

\clearpage

\subsubsection{Uncurated Generated Samples on CelebA-HQ $256\times256$}
\begin{figure}[!htb]
    \centering
    \includegraphics[width=0.95\linewidth]{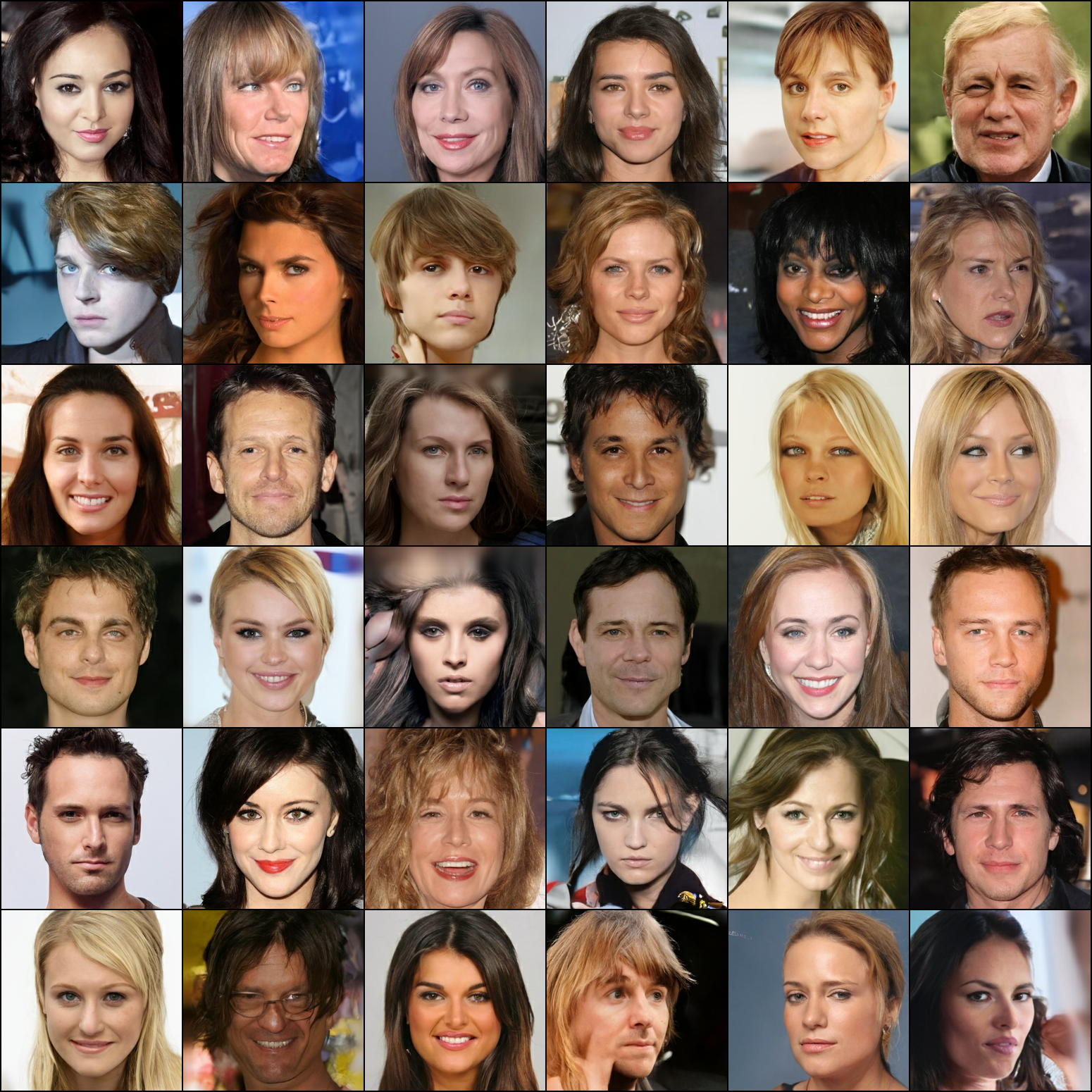}
    \caption{Uncurated samples from the unconditioning modeling on CelebA-HQ $256\times256$.}
    \label{fig18}
\end{figure}

\begin{figure}[!htb]
    \centering
    \includegraphics[width=1.0\linewidth]{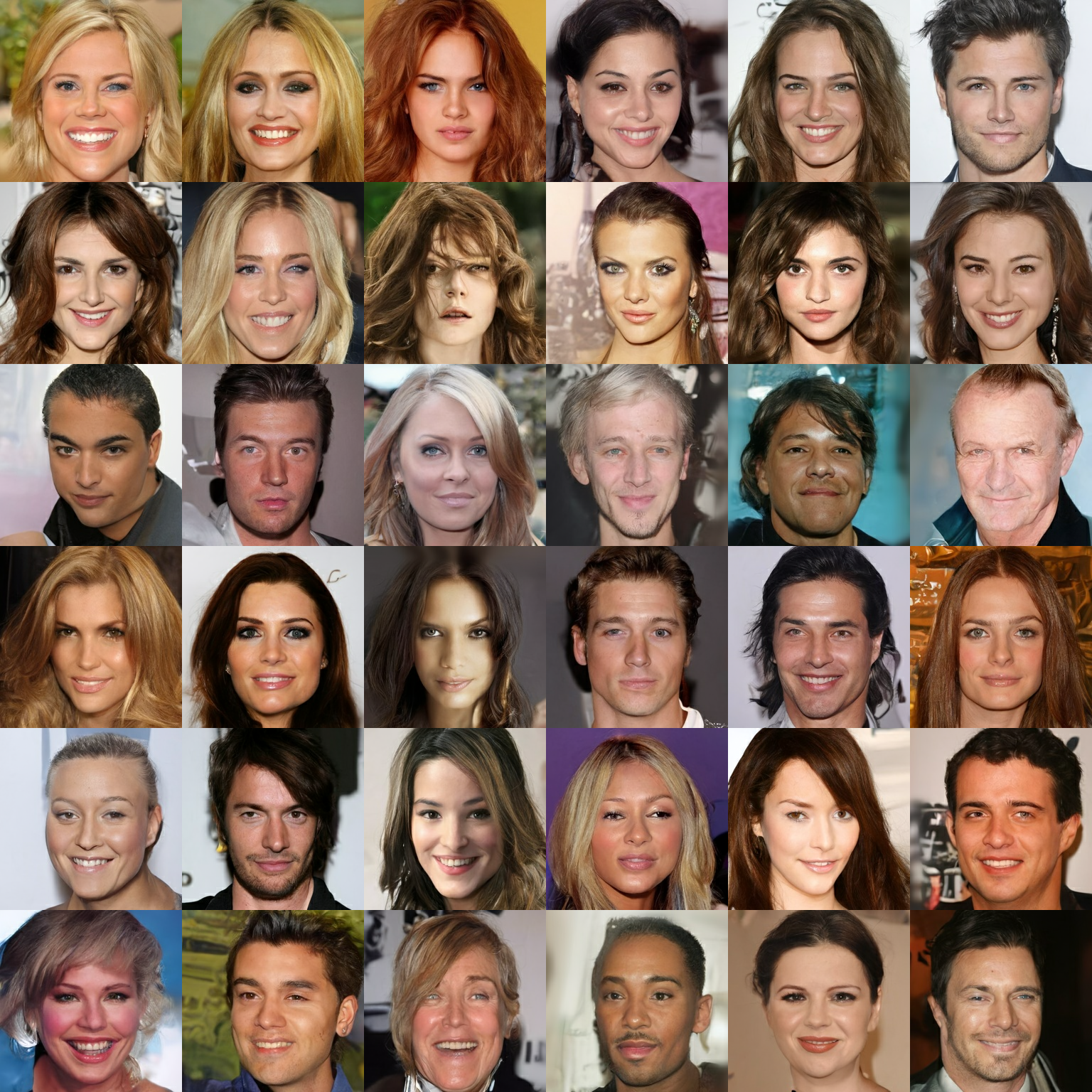}
    \caption{Uncurated samples from the temperature-based model ($T_i = 5/\sigma_i, K = 30$) on CelebA-HQ $256\times256$.}
    \label{fig19}
\end{figure}

We also provide uncurated samples from our unconditioned model and from the temperature-based model (\(T_i = 5 / \sigma_i, K = 30\)) in Figures~\ref{fig18} and~\ref{fig19}, respectively. In both figures, we observe interesting samples that appear to mix characteristics typically associated with different genders, such as predominantly male faces with more feminized hairstyles.

\clearpage

\section{Discussion}

In this section, we summarize the main insights of our framework, relate them to existing diffusion practice, and outline promising directions for future work. A recurring theme is that our analysis recasts diffusion behavior in terms of simple optimization and high-dimensional geometry, avoiding heavy PDE machinery or complex probabilistic inference and thus lowering the conceptual barrier for understanding diffusion models.

\paragraph{Neural networks as implicit temperature-smoothed scores.}
Our analysis suggests that the score field learned by neural networks is not simply a noisy estimator of the empirical score, but behaves like an \emph{implicitly temperature-smoothed} version of the empirical softmax weights. The expansiveness analysis in \cref{sharp_overlap} and the Jacobian spectrum measurements (\cref{A.5,B.2.4}) make this precise: in overlap regions of multiple Gaussian shells, the empirical score is extremely sharp and highly expansive (large positive Jacobian eigenvalues), whereas the learned score has much smaller expansiveness, comparable to explicitly applying a large temperature $T$ (or a larger $K$ in KNN approximations) to the empirical weights.

Our memorization studies connect this picture to concrete behavior on different datasets. On the small Cat--Caracal dataset, where shell overlaps are relatively sparse and simple, the network can almost perfectly track the empirical score field: the cosine similarity to the empirical mixture score at low noise is $\approx 0.99$ (\cref{B.2.3}), and a pixel-space nearest-neighbor ratio clearly flags a substantial fraction of generated samples as copies (\cref{B.3.2}). In this regime, the network can fit the ``spiky'' empirical score, so implicit smoothing plays a limited role and memorization is easy to detect.

On CIFAR-10, the much larger number of training points induces dense overlaps and a highly irregular empirical score. Here the cosine similarities are noticeably lower (angles of about $20^\circ$–$40^\circ$ even at small noise; \cref{B.2.3}), indicating that the network no longer reproduces empirical scores pointwise, but instead learns a smoother field aligned with local manifolds formed by multiple nearby samples. In this regime, our nearest-neighbor memorization ratio—highly discriminative on Cat--Caracal—largely fails to separate different modelings on CIFAR-10, precisely because generated samples lie within dense local manifolds where ``nearest vs.\ second-nearest'' distances are no longer a reliable proxy for memorization.

Overall, these results support the view that neural networks act as implicit temperature smoothers of the empirical score: on small or low-complexity datasets they can closely fit the empirical objective and thus memorize, while on larger or more complex datasets their inductive bias forces them to smooth the score weights, suppressing extreme local expansiveness and reducing exact replication. Our explicit Temperature Smoothing and Noise Unconditioning mechanisms make this effect \emph{controllable and observable}: they reduce memorization fractions on Cat--Caracal and lower sampling expansiveness (\cref{exp_ratio,B.3.2}), while still enabling meaningful generalization on CIFAR-10 and beyond. Conceptually, this ``implicit temperature'' view pinpoints \emph{what} is being smoothed: not an abstract learned function, but the empirical softmax weights over overlapping Gaussian shells. This yields a concrete geometric picture that ties the same smoothing mechanism to both memorization on small datasets and generalization on large ones.

At the same time, our findings raise an open methodological question: how should one \emph{quantitatively} characterize the memorization–generalization trade-off at realistic scale? The nearest-neighbor ratio is informative on Cat--Caracal but quickly loses discriminative power on CIFAR-10, exactly when samples move from single-point collapse to residing in dense local manifolds. Developing metrics that remain sensitive in this regime—for example by combining pixel and feature spaces or explicitly modeling local density—is an important direction for future work.

\paragraph{Optimization-based sampling via noise unconditioning.}
A key contribution of this work is the optimization-based reformulation of sampling enabled by noise unconditioning. By removing the explicit noise input and optimizing the unified loss in Equation~(11), the model learns the score of a \emph{fixed} Gaussian mixture $\pMN$, independent of diffusion time. Sampling can then be interpreted as continuous-time gradient ascent on $\log \pMN(x)$, rather than as a purely time-dependent reverse SDE or ODE.

This perspective offers both conceptual and practical benefits. Conceptually, it demystifies the sampling process: once noise is unconditioned and the model learns the score of a fixed Gaussian mixture $\pMN$, our sampler can be viewed as a continuous-time gradient flow on the static objective $\log \pMN(x)$ whose modes coincide with training samples. This gradient-flow picture relies only on elementary optimization and high-dimensional Gaussian geometry, offering a lower-barrier and more visual alternative to traditional SDE/PDE-based explanations of diffusion behavior. Practically, this viewpoint suggests treating sampling as an optimization problem, motivating the use of adaptive step sizes, principled stopping rules, and constrained gradient flows (e.g., via projection) to better exploit the geometry of the learned score field. More broadly, it invites future work to systematically import tools from modern optimization—such as line-search, trust-region, and variational methods—into the design and analysis of diffusion samplers.

However, this optimization view also highlights a deeper question. Our analysis characterizes \emph{what} the learned, smoothed score field looks like, but not fully \emph{why} standard neural networks trained with simple MSE losses should converge to this particular smoothed solution among all possibilities. In particular, it remains unclear why the network prefers to organize data in a semantic \emph{feature} space rather than in raw pixel space, and how architectural choices and optimization dynamics jointly induce this preference. The gradient-flow interpretation offers a useful lens on these questions by framing learning and sampling as motion in a fixed energy landscape, but a complete understanding of the resulting implicit bias is still an open problem.

\paragraph{Extensions to latent diffusion and consistency models.}
Because our framework is formulated at the level of score fields, it is largely agnostic to architecture and suggests several extensions. First, applying Noise Unconditioning and Temperature Smoothing to \emph{latent diffusion models} is especially natural. Latent spaces typically have lower curvature and better-structured manifolds than pixel space, so our smoothing mechanisms could operate more stably—allowing larger $T$ or broader neighborhoods—without drifting off-manifold, as suggested by our feature-space KNN experiments. In this sense, our geometric view of ``local shells'' and manifolds carries over almost verbatim from pixel space to learned latent spaces, potentially making the theory even simpler there. Second, consistency models and ODE-based distillation methods could directly exploit the optimization viewpoint: if sampling is seen as gradient ascent on $\log \pMN$, a consistency network could be trained to map coarse noise directly to near-stationary points of this objective, potentially yielding fast samplers with improved generalization. Our preliminary results in \cref{B.4.3} already indicate that our model can be trained as a consistency model like the standard diffusion models.

\paragraph{Toward adaptive smoothing and temperature design.}
Our hand-crafted temperature schedules (e.g., $T_i \propto 1/\sigma_i$) already yield substantial memorization reduction with modest FID degradation, but the Jacobian analysis in Appendix~A.5 shows that temperature interacts in a nontrivial way with both the covariance and isotropic contraction terms of the score Jacobian. This points to the need for more principled, \emph{adaptive} schemes. Future work could explore temperature schedules $T(x,\sigma)$ that depend on local geometric statistics—such as estimated manifold curvature, local density, or semantic features—and learn these jointly with the score network. Regularizing such schedules by local expansiveness constraints or memorization-aware objectives (e.g., penalizing low nearest-neighbor ratios on held-out training batches) is a natural next step toward controllable generalization.

Finally, we hope that the optimization- and geometry-based perspective developed here, together with our quantitative memorization analysis across small and large datasets, provides a simple and unified foundation for studying diffusion models, and helps further bridge the gap between intuitive geometric understanding and the observed memorization behavior of large-scale generative models.


\end{document}